%% file: main.tex
\documentclass[10pt,twocolumn,letterpaper]{article}

\usepackage[pagenumbers]{cvpr} % To force page numbers, e.g. for an arXiv version

\definecolor{cvprblue}{rgb}{0.21,0.49,0.74}
\usepackage[pagebackref,breaklinks,colorlinks,allcolors=cvprblue]{hyperref}

\usepackage{booktabs}
\usepackage{multirow}
\usepackage{caption}
\usepackage{graphicx}
\usepackage{caption}
\usepackage[dvipsnames]{xcolor} % To access some named colors used with \highLight
\usepackage{amssymb}
\usepackage{arydshln}
\usepackage{pifont}
\usepackage{subcaption}
\usepackage{bold-extra}
\usepackage{url}
\usepackage{amsmath,amsfonts,bm}
\usepackage{array} % For defining column widths
\usepackage{adjustbox}
\usepackage{xcolor}
\usepackage{verbatim}
\usepackage{tabularx}

\usepackage{tikz}
\input{math_commands.tex}

\definecolor{mydarkblue}{rgb}{0,0.08,1}
\definecolor{mydarkgreen}{rgb}{0.02,0.6,0.02}
\definecolor{myred}{rgb}{1.0,0.0,0.0}
\definecolor{myred2}{rgb}{0.7,0.1,0.1}
\definecolor{mydarkblue2}{rgb}{0.05,0.1,0.7}
\definecolor{mydarkblue2}{rgb}{111,0,255}
\definecolor{mypurple2}{rgb}{111,0,111}

\def\Approach{ShapeWords\xspace}

\title{ShapeWords: Guiding Text-to-Image Synthesis with 3D Shape-Aware Prompts}

\author{Dmitry Petrov$^1$~~~Pradyumn Goyal$^1$~~~Divyansh Shivashok$^1$~~~Yuanming Tao$^1$
	\vspace{0.1cm}\\
	Melinos Averkiou$^{2,3}$~~~Evangelos Kalogerakis$^{1,2,4}$
	\vspace{0.2cm}\\
	$^1$UMass Amherst~~~$^2$CYENS CoE~~~$^3$University of Cyprus~~~$^4$TU Crete
	\vspace{-0.2cm}\\
}

\begin{document}

\twocolumn[{%
\renewcommand\twocolumn[1][]{#1}%
\maketitle
\begin{center}
    \centering
    \captionsetup{type=figure}
    \includegraphics[width=\textwidth]{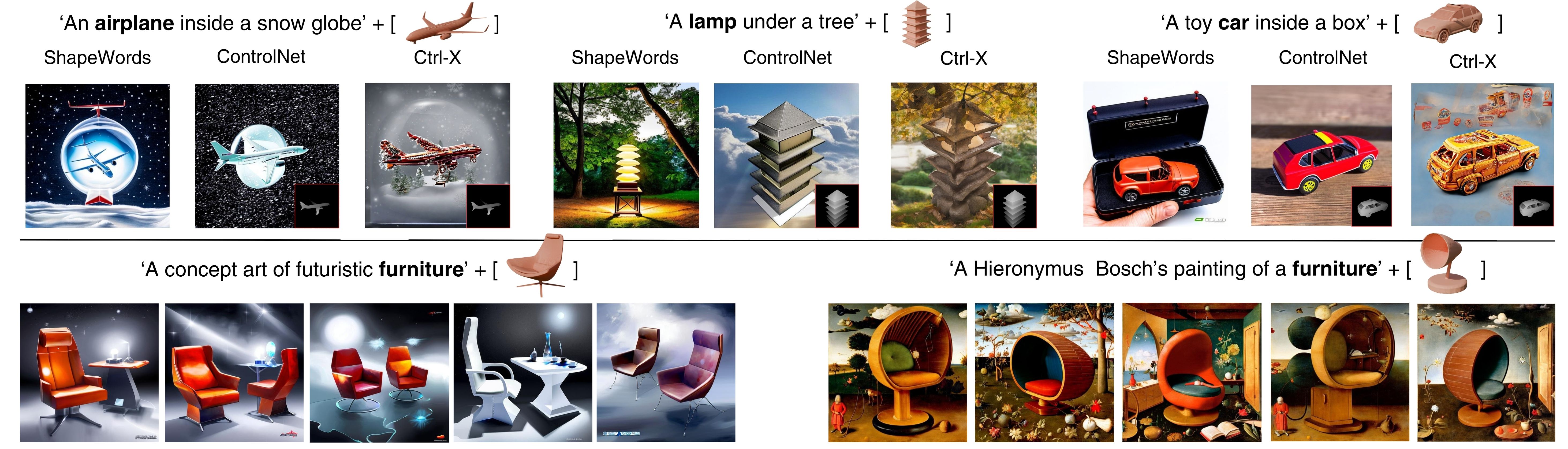}
    \captionof{figure}{ShapeWords enables 3D shape-aware text-to-image generation via mapping of shape geometries into CLIP  space. Given an input 3D shape and text prompts describing desired appearance and context, our method generates images that maintain both shape fidelity and text compliance. Unlike existing methods that use view-dependent guidance like depth maps, ShapeWords generalizes to compositional settings \textbf{(top row)} and allows for exploration of target geometries with stylistic deviations aligned with the prompt \textbf{(bottom row)}.}
    \label{fig:teaser}
\end{center}%
}]

\input{sec/0_abstract}

\input{sec/1_intro}
\input{sec/2_relwork}

\input{sec/3_method}

\input{sec/4_experiments}

\input{sec/5_discussion}

{
    \small
    \bibliographystyle{ieeenat_fullname}
    \bibliography{main}
}

 \input{sec/X_suppl}

\end{document}

%% file: math_commands.tex
\def\eqref#1{equation~\ref{#1}}
\def\1{\mathbf{1}}

\def\vtheta{{\mathbf{\theta}}}
\def\vepsilon{{\mathbf{\epsilon}}}

\def\vp{{\mathbf{p}}}

\def\vt{{\mathbf{t}}}

\def\vz{{\mathbf{z}}}

\def\mB{{\mathbf{B}}}

\def\mI{{\mathbf{I}}}

\def\mK{{\mathbf{K}}}

\def\mQ{{\mathbf{Q}}}

\def\mT{{\mathbf{T}}}

\def\mV{{\mathbf{V}}}

\def\mX{{\mathbf{X}}}

\DeclareMathAlphabet{\mathsfit}{\encodingdefault}{\sfdefault}{m}{sl}
\SetMathAlphabet{\mathsfit}{bold}{\encodingdefault}{\sfdefault}{bx}{n}

\def\emS{{S}}

\newcommand{\R}{\mathbb{R}}

\newcommand{\deltaT}{\delta \mathbf{T}}

%% file: sec/0_abstract.tex
%\begin{figure}[ht]
%\vspace{-1mm}
%\begin{center}
% \includegraphics[width=\textwidth]{example-image-a}
%\end{center} 
%\vspace{-3mm}
% \caption{We present \Approach.
% \label{fig:teaser} 
% }  
%\end{figure}

\begin{abstract}
We introduce \Approach\footnote{Project page (with code): \href{https://lodurality.github.io/shapewords/}{lodurality.github.io/shapewords}}, an approach for synthesizing images based on 3D shape guidance and text prompts.
\Approach incorporates target 3D shape information within specialized tokens embedded together with the input text, effectively blending 3D shape awareness with textual context to guide the image synthesis process. Unlike conventional shape guidance methods that rely on depth maps restricted to fixed viewpoints and often overlook full 3D structure or textual context, \Approach generates diverse yet consistent images that reflect both the target shape’s geometry and the textual description. Experimental results show that \Approach produces images that are more text-compliant, aesthetically plausible, while also maintaining 3D shape awareness.
%In case of coarse target shape structure specifications (e.g., primitives for shape parts, such as cuboids and spheres), our method generates plausible stylized variations instead of reproducing coarse shape appearance in the synthesized images. Our experiments demonstrate  qualitatively and quantitative image generation results that are  more aesthetically plausible and diverse, while also accurately capturing the target shape structure and rest of textual descriptions.
\end{abstract}

%% file: sec/1_intro.tex
\section{Introduction}
\label{sec:intro}

Recent advances in generative image models based on diffusion \cite{ddpm,ddim,dalle1,dalle2,dalle3,imagen,rombach2022stable} have made it possible to generate impressive imagery from input text prompts. A challenge in text-to-image models has been to provide users with fine-grained control over shapes or forms in the synthesized images, which can be difficult to convey through text descriptions alone. To address this, conditioning methods have been proposed, such as  ControlNet~\cite{zhang2023controlnet} and IP-adapter~\cite{ye2023ip-adapter}, 
that aim to capture the desired shape or form more explicitly through the use of edge or depth maps as input conditions.

Despite these advancements, current text- and image-conditioned synthesis approaches still face a number of challenges. First, they often struggle to balance both textual and visual conditions, when text describes a particular context that should be combined with the target shape to guide an image synthesis (Figure \ref{fig:teaser}, top row). Second, commonly used visual conditions such as edge or depth maps are limited to a single viewpoint, resulting in a loss of valuable 3D shape information when users seek image variations of an underlying shape from different poses. Third, even when these models accurately reflect the target shape in specific views, users may want to explore shape variations -- yet current models often lack flexible controls for such exploration.

To overcome these challenges, we propose \Approach, a method designed to generate images that faithfully adhere to both the text prompt and a target 3D shape geometry, while at the same time allows users to easily explore both pose and shape variations. 
At the heart of \Approach is the embedding of the 3D shape into a shared space with textual descriptions, specifically OpenCLIP~\cite{cherti2023OpenCLIP} space, enabling the target shape to be  fully integrated with the prompt. Unlike view-specific conditioning (e.g., through depth, edge, or normal maps), our approach captures the entire 3D shape in the embedding, facilitating diverse image synthesis across challenging prompts, yet remaining geometrically consistent with the target 3D shape (Figure \ref{fig:teaser}, top row).
Furthermore, \Approach\ allows users to control the degree of deviation from the target shape -- a feature that is particularly useful when the target structure is represented by coarse geometric primitives. This capability enables the generation of diverse, stylized images that retain plausible geometric variations of the specified target structure (Figure \ref{fig:teaser}, bottom row).

Our experiments evaluate \Approach\ on consistency with both text and target shape, as well as aesthetic quality. It achieves significant improvements against ControlNet-based variants, as measured by several metrics capturing target shape compliance, text prompt compliance, image plausibility, and also as validated by a perceptual user study.

Overall, our technical contributions are as follows: 
\begin{itemize} 
\item We introduce 3D shape tokens -- specialized tokens that enable text-to-image models to generate plausible images adhering to both 3D geometry and textual conditions. 
\item We also enable user control over the shape guidance, allowing users to explore images that depict variations of target shapes across varying poses and appearances.
\end{itemize}

%-------------------------------------------------------------------------

%% file: sec/2_relwork.tex
\section{Prior work}
\label{sec:priorwork}
\label{sec:relatedwork}

\paragraph{Conditional Diffusion.}
Denoising diffusion models~\cite{ddpm, ddim} have revolutionized image synthesis by generating high-quality, diverse, and plausible image content. To control the generation process, the most common condition is text. Popular diffusion approaches, such as Stable Diffusion \cite{rombach2022stable}, DALLE~\cite{dalle1,dalle2,dalle3}, Imagen~\cite{imagen} have used large-scale text-image datasets, including pretrained LLMs~\cite{gpt3,t5, bert} for text-to-image synthesis. However, relying solely on text prompts cannot fully take
advantage of the knowledge learned by diffusion models, especially when flexible and accurate control is needed in terms of form or layout. To this end, 
ControlNet~\cite{zhang2023controlnet} pioneered explicit conditioning through visual signals like depth or edge maps, enabling spatial control but limited to single viewpoints.
T2I-Adapter~\cite{mou2023t2i} proposed a lighter-weight alternative through specialized adapters of visual control signals.
UniControl~\cite{qin2023unicontrol} 
consolidated a wide array of controllable
condition-to-image tasks within a single framework. 
IP-Adapter~\cite{ye2023ip-adapter} proposed a  decoupled cross-attention strategy for text features and image features from input conditions for more accurate controllable generation. 
In general, these methods rely on view-dependent control signals (e.g., depth, edge maps), lack explicit 3D geometry awareness, and may fail to generate images adhering to both the text prompt and target 3D shapes. In contrast, \Approach enables view-independent shape control, while adhering well both to target shape geometry and text prompts.

\paragraph{Structure guidance.} 
Several approaches have explored novel view synthesis aimed at maintaining some level of 3D awareness for 
view consistency ~\cite{xiang20233daware, liu2023zero123, gu2022stylenerf, liu2023syncdreamer, melaskyriazi2023realfusion}. However, these methods do not disentangle 3D structure from appearance and do not provide geometric control to users, meaning they generate images of a shape with consistent appearance across different views without user-driven control over geometry.
A number of approaches have been proposed to incorporate structure guidance from depth maps or coarse 3D primitives disentangled from appearance. FreeControl~\cite{mo024freecontrol} explores structure guidance   in diffusion feature subspaces
extracted from depth maps and other visual conditions. Ctrl-X~\cite{lin2024ctrlx} investigated more efficient, disentangled and zero-shot control of structure and appearance. LooseControl~\cite{bhat2024loosecontrol} introduced more flexible conditioning schemes through 3D scene boundary control, 3D box control and attribute editing. Diffusion Handles~\cite{pandey2024diffusionhandles} enabled localized control of 3D object parts in diffusion models by introducing deformation handles to edit images while maintaining consistent perspective and structure across views.
In our approach, we take a different route by embedding 3D shapes into tokens within text prompts, providing more explicit geometry guidance in image synthesis and combining it with additional specification of separate context, style, and appearance constraints through text.

\input{figures/method}
\paragraph{Guidance from learnable tokens and concept learning.}
Several methods have explored embedding concepts into textual tokens for personalized image synthesis using textual inversion~\cite{gal2023textual}. The required optimization for textual inversion is often very slow, thus various methods have explored more efficient fine-tuning strategies~\cite{kumari2023custom}, feed-forward architectures to predict textual tokens~\cite{shi2024instantbooth,li2024photomaker}, or using hypernets~\cite{ruiz2024hyperdreambooth}. Recently, more efficient methods of visual concept learning were introduced \cite{safaee2024clic, avrahami2023break, gal2023encoder, ding2024clip} allowing for more efficient learning and transfer of visual concepts. A more spiritually similar approach to ours is that of ``continuous 3D words''\cite{cheng2024learning}, which embeds 3D-aware attributes, such as time-of-day lighting, bird wing orientation, dolly zoom effects, and object pose, into learnable tokens. Viewpoint textual inversion\cite{viewneti} learns 3D view tokens that can be used to control the viewpoint for image synthesis. However, unlike these methods, our approach learns to embed 3D shapes directly into tokens, enabling image generation that is guided by both target 3D shape geometry and text. To our knowledge, text-to-image synthesis through learned 3D shape words has not been explored before.

\begin{comment}
Several methods have explored embedding new concepts into text-to-image models. Textual Inversion~\cite{gal2023textual} introduced learnable tokens for style and appearance, while Custom Diffusion~\cite{kumari2023custom} and HyperDreambooth~\cite{ruiz2024hyperdreambooth} proposed efficient parameter tuning approaches. PhotoMaker~\cite{li2024photomaker} and InstantBooth~\cite{shi2024instantbooth} further improved adaptation efficiency. 
However, these methods primarily target appearance and style transfer rather than geometric structure. Our work extends concept embedding to the 3D domain, learning shape-aware tokens that preserve both geometric and contextual information while enabling efficient inference.
\end{comment}

%% file: figures/method.tex
\begin{figure*}[th!]
    \centering
    \includegraphics[width=1.0\textwidth]{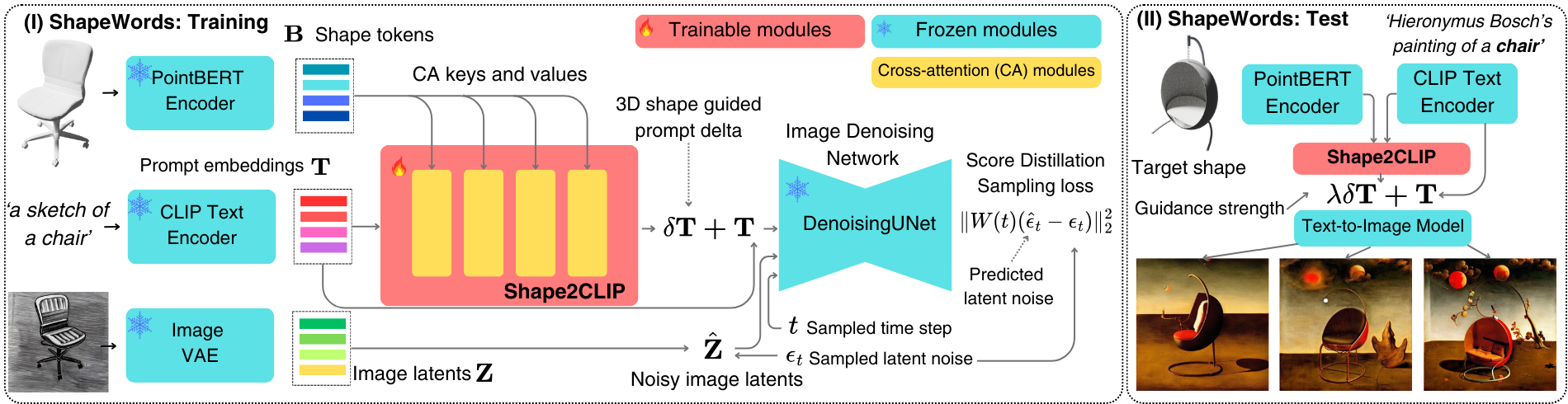}
    %\vspace{-7mm}
    \caption{\textbf{\Approach{} Training and Inference.} During training \textbf{(\romannumeral 1)}, \Approach{} takes as input triplets of a shape, a prompt, and an image. The shape \( S \) and prompt \( T \) are encoded using the shape encoder PointBert and the text encoder OpenCLIP, respectively. The resulting embeddings are passed through a cross-attention-based Shape2CLIP module, which produces a prompt residual \( \delta T \) to guide the source prompt toward the target geometry. This modified prompt is then passed as input to a Text-to-Image Denoising UNet along with sampled noisy image latents and time step \( t \). The Shape2CLIP module is optimized via Score Distillation Sampling. During inference \textbf{(\romannumeral 2)}, the CLIP embeddings of the input prompt and the target embeddings of the test shape \( S_{\text{test}} \) are passed through the Shape2CLIP module. Optionally, the desired strength of the shape guidance is controlled by the user parameter \( \lambda \).}     \label{fig:architecture}    
    %\vspace{-3mm}
\end{figure*}

%% file: sec/3_method.tex
\section{Method}

\paragraph{Overview.}
Given a text prompt $\vp$ and a target 3D shape~$\emS$, \Approach generates an image that reflects both the textual description and the desired shape. Users indicate the desired shape directly within the prompt using a special token, such as ``a red [SHAPE-ID] on a beach'', where ``SHAPE-ID'' corresponds to a shape, e.g., one imported from a 3D shape database.

The pipeline (illustrated in Figure \ref{fig:architecture}) proceeds as follows at test time. First, a shape representation is extracted from the chosen 3D shape using a pre-trained transformer, Point-BERT~\cite{yu2022pointbert}. The text prompt is mapped into CLIP space through an OpenCLIP encoder~\cite{cherti2023OpenCLIP}, where the token ``SHAPE-ID'' is replaced by a category name for the shape (e.g., ``chair''). Our method then applies a new module, named Shape2CLIP, trained to modify the prompt's word embeddings, including the shape identifier token, so that the resulting prompt embedding integrates the desired 3D shape geometry in the prompt  while preserving its original textual context. This shape-enhanced embedding is passed to a Stable Diffusion model~\cite{rombach2022stable}, along with a user-defined parameter that controls the degree of 3D shape influence, to synthesize images consistent with both the textual and 3D shape cues.

In the following sections, we discuss preprocessing of the input shapes and text prompts to extract  Point-Bert~\cite{yu2022pointbert} and OpenCLIP~\cite{cherti2023OpenCLIP} representations respectively   (Section \ref{sec:preliminaries}), then we discuss the Shape2CLIP module (Section
\ref{sec:shape2clip}), our  inference pipeline (Section \ref{sec:inference}), and finally its training procedure  (Section \ref{sec:training}).

\subsection{Preprocessing}
\label{sec:preliminaries}

\paragraph{Point-BERT.} 
To represent the input 3D shape, we use a pre-trained Point-BERT~\cite{yu2022pointbert}, a transformer-based model for point clouds. The input shape is first converted into a point cloud of $1024$ points using farthest point sampling, then passed through the Point-BERT architecture, which partitions the cloud into 64 patches. These patches are encoded by PointNet~\cite{qi2017pointnet} and a transformer based encoder as a set of tokens representing the geometry of the shape. The resulting shape embedding, $\mB \in \R^{65 \times 384}$, consists of 65 tokens representing both patches and a class token in a 384-dimensional space. Trained via masked modeling for self-supervised learning, Point-BERT effectively captures structure-aware features demonstrated in various shape processing applications, such as part segmentation.

\paragraph{OpenCLIP space.} As our base generative model, we use Stable Diffusion 2.1 whose text encoder is OpenCLIP ViT-H/14~\cite{cherti2023OpenCLIP}. 
Thus, we process the input textual prompt $\vp$, through this OpenCLIP encoder to generate a sequence of $77$ token embeddings capturing the words in the prompt and their context in the prompt:
\begin{equation}
\mathbf{T} = [ \vt_{0}, \vt_{1}, ..., \vt^{eos }_{j},..., \vt^{pad}_{76} ]
\end{equation}
where $\vt_j$ represents the encoded embedding of the $j$-th token in the prompt; $\vt^{eos}_j$ represents end-of-sequence (EOS) token which captures the context of the whole prompt; and $\vt^{pad}_j$ represent padding token embeddings that pad sequence after EOS tokens to some predefined length (e.g. 77).
\subsection{Shape2CLIP module}
\label{sec:shape2clip}

Given a text prompt embedding $\mT$ and shape representation~$\mB$, our Shape2CLIP module generates a modified prompt embedding $\mT'$ that combines information from both the text prompt and 3D shape. Specifically, we update the embeddings of two tokens. First, we update the shape identifier token(s), which was originally set to the category name of the shape (e.g. 'chair'), such that its embedding  instead reflects the specific 3D shape. We also found it important to update the EOS token since it captures the context of the whole prompt, including the desired shape to be incorporated. Formally, the module implements a learned function:
\begin{equation}
\mT'[s, e] = \mT[s, e] +\ \deltaT(\mB, \mT; \vtheta)
\end{equation}
where $s$ and $e$ are the position of the shape and EOS tokens, and  $\deltaT$ is a residual mapping with learned parameters $\vtheta$. We found this residual approach to be more effective than a direct feedforward network, as it is less prone to overfitting with limited training data. We found that updating only these specific two tokens preserves the prompt's original context while integrating geometric information. 

The residual function $\deltaT$ is constructed using cross-attention blocks~\cite{vaswani2017attention}. In the first block, the shape representation $\mB$ is treated as keys, while the original prompt embedding $\mT$ serves as queries:
\begin{equation}
\mK = \mB \cdot W_k, \mQ = \mT \cdot W_q, \mV = \mB \cdot W_v
\end{equation}
and new shape-aware features for the prompt $\mX$ are obtained as:
\begin{equation}
\mX = CrossAttention(\mQ, \mK, \mV) + \mT
\end{equation}
followed by a MLP-based residual block~\cite{vaswani2017attention}. In subsequent blocks, the updated prompt features act as queries, with Point-BERT representations serving as keys to further refine the prompt embedding. Our implementation uses a total of six cross-attention blocks. The output of last block is passed through a linear layer to yield the final modified embedding $\mT'$. %This layer is initialized with small weights, so $\deltaT$ values are close to zero in the beginning of the training.

\subsection{Guided Diffusion}
\label{sec:diffusion}
\label{sec:inference}

The modified embedding $\mT'$ can be directly used as input to a diffusion model to generate images, formulated as $\mI = D(\vz, \mT')$, where $\vz$  in the latent space and $D$ represents Stable Diffusion 2.1~\cite{rombach2022stable} in our implementation. Notably, our method does not require any additional shape conditions (e.g., depth, normal, or edge maps).

\paragraph{Shape Guidance.} Our method enables users to control the influence of the 3D shape, allowing for image variations that deviate from the target shape. This flexibility is valuable when the exact shape a user has in mind does not exist in any available 3D shape databases, deeming more necessary to explore variations of existing shapes.  The shape influence is modulated by a parameter $\lambda \in [0,1]$ using a linear interpolation scheme to adjust the embedding as follows:
\begin{equation}
\mT'[s, e] = \mT[s, e] + \lambda \cdot \deltaT(\mB, \mT; \vtheta)
\end{equation}
As shown in our results, varying $\lambda$ from $0$ to $1$ gradually shifts from disregarding the shape influence to incorporating it in the generated images. Intermediate values yield plausible images, while higher $\lambda$ values produce shapes in the generated images that increasingly match the desired target shape. 

\subsection{Training}
\label{sec:training}

Our training procedure aims to learn the parameters $\vtheta$ of the Shape2CLIP module, keeping the rest of the pipeline components fixed (i.e., PointBERT encoder, text encoder, image encoder/decoder, and denoising network).

\paragraph{Training dataset.}
To train the Shape2CLIP module, we constructed a dataset of shape-prompt-image triplets based on ShapeNet~\cite{chang2015shapenet} reference shapes. Images were generated using ControlNet~\cite{zhang2023controlnet} conditioned on ShapeNet's depth maps, as provided by the ULIP authors~\cite{xue2023ulip}, where each shape is rendered from 30 viewpoints, rotated in 12-degree increments around the vertical axis with fixed elevation (see~\cite{xue2023ulip} for details).

For each depth image, we applied a randomly selected
prompt from a  set of $13716$ prompts for additional ControlNet conditioning, created from a base set of $100$ prompts, then augmented with variants produced by ChatGPT
\cite{chatgpt}.
To achieve diversity and structural agnosticism in prompts, this set was created by combining $127$ artistic mediums (e.g., ``painting,'' ``watercolor,'' ``sketch'') with $108$ style adjectives (e.g., ``colorful,'' ``pixelated,'' ``fantasy''), deliberately avoiding references to specific 3D structures to help the model learn generalizable mappings instead of overfitting to particular geometries or appearance combinations. To reduce bias and enhance background diversity, we used the Stable Diffusion's inpainting model \cite{rombach2022stable} to modify the backgrounds while preserving the foreground objects. The above procedure resulted in a diverse dataset of $1.58$M prompt-image pairs (30 per each ShapeNet shape), generated from all depth images in the training split from 3DILG~\cite{zhang2022dilg} in ShapeNet.
As demonstrated in Section \ref{sec:results}, although our method is trained on data generated by ControlNet, ShapeWords exhibits strong generalization capabilities. Notably, ShapeWords achieves significantly better performance on compositional prompts, a setting that ControlNet struggles to handle effectively. \emph{The dataset will be publicly released along with our source code upon acceptance}.

\paragraph{SDS-based training.} 
We train our Shape2CLIP model using the Score Distillation Sampling (SDS) loss~\cite{poole2023dreamfusion}. Specifically, for sampled noise $\vepsilon_{t,i} \sim \mathcal{N}(0,\, \mathbf{I})$ for a training image $i$ at step $t$, we use  the pretrained Stable Diffusion model to predict the noise $\hat{\vepsilon}_{t,i}$, and backpropagate the loss:
\begin{equation}
\mathcal{L}_{\text{SDS}}(\vtheta) = W(t)\, \left\| \hat{\vepsilon}_{i,k} - {\vepsilon}_{i,k} \right\|_2^2,
\end{equation}
where $W(t)$ is a time-dependent weighting function proposed in DreamTime~\cite{huang2024dreamtime} for enhancing the stability of the SDS-based training. We provide details in supplementary.

%% file: sec/4_experiments.tex
\section{Evaluation}
\label{sec:results}
\label{sec:evaluation}
We now discuss our evaluation and experiments to test the effectiveness of \Approach compared to alternatives.

\paragraph{Evaluation goals \& metrics.} The evaluation aims to assess three main aspects:

\noindent (a) \emph{Prompt Adherence}: We evaluate how well the images generated by each competing method align with the given prompt. We use the standard score of \textbf{CLIP similarity} \cite{radford2021learning}.

\noindent  (b) \emph {Shape Adherence}: We assess how well the generated images from each method adhere to a reference 3D shape. To this end, we compare the shapes in the generated images with the reference 3D shapes based on their silhouette. 
To extract silhouttes from reference 3D shapes, we render them from a target pose, extract the silhouette, then we compare it with the silhouette of the shape in the generated images conditioned on the target pose. The silhouette from generated images are extracted through the foreground detection model \cite{zheng2024birefnet}.  The similarity between silhouettes is measured using standard geometric similarity metrics, specifically Intersection over Union (\textbf{S-IOU}) and Chamfer Distance (\textbf{S-CD}), averaged per shape across six uniform views sampled with 60 degree increments, starting at 0.

\noindent  (c) \emph{Image Plausibility}: We evaluate the aesthetic quality of the generated images using the aesthetics score (\textbf{Aes.}) \cite{Schuhmann}. We also include the commonly used image quality generation metrics of \textbf{FID}  \cite{heusel2017gans} and \textbf{KID} \cite{binkowski2018demystifying}. Given reported issues with Inception-based features \cite{kynkaanniemi2022role,parmar2022aliased}, we compute FID and KID on CLIP features as recommended in \cite{betzalel2022study}.

\paragraph{Test Datasets.} We have designed two datasets for evaluation to test different properties of our model:

\noindent (a)
\emph{Simple Prompts Dataset}: This dataset contains ``simple prompts'' with text structured as "a photo of a [SHAPE-ID]," where the [SHAPE-ID] token corresponds to a 3D shape from the ShapeNet test splits from all $55$ categories ($2592$ test shapes). This dataset is particularly useful for evaluating shape adherence through geometric similarity of extracted silhouettes (using S-CD and S-IOU metrics), as the generated images are photo realistic and feature a single object, so the foreground detection methods perform reliably here. 

\noindent (b) \emph{Compositional Prompts Dataset}: This dataset serves as our main evaluation set since it is more challenging, containing ``compositional prompts'' that involve a target 3D shape alongside additional objects or humans interacting with it. It contains five prompts not present in training data that are designed to test compositional properties of studied models: ``a [SHAPE-ID] under a tree'', ``a craftsman working on a [SHAPE-ID]'', or ``a toy [SHAPE-ID] in a box'', ``a [SHAPE-ID] in a snow globe'', ``an artist painting [SHAPE-ID]''. Here, the [SHAPE-ID] tokens correspond to 3D shapes from the ShapeNet test split we use in our evaluation (2592 shapes). We report all evaluation metrics for this dataset except for silhouette similarity metrics, as the foreground in these scenarios is more complex, often containing multiple interacting objects, leading to noisier silhouette extractions. For this dataset, we conduct a user study to assess both shape and prompt adherence from a perceptual standpoint.

\input{figures/geom_eval_by_k}

\input{figures/comparison_by_k_compact}

\input{figures/quant_visual_eval_table}

\paragraph{Baselines.}
Our experiments involve the following methods:

\input{figures/comp_eval_qualitative}

\noindent (a) \emph{ControlNet}: ControlNet receives two inputs: (i) a depth image rendered from a target 3D shape in a specified pose (we use inverted depth images provided  by the ULIP project~\cite{xue2023ulip}), and (ii) a text prompt from our test datasets where [SHAPE-ID] is replaced with the ShapeNet category label at the finest hierarchical level of ShapeNet. ControlNet's control strength is set to 1 (default). This setup makes ControlNet a strong baseline for shape adherence as it generates images whose object silhouettes closely match those of the input depth image. However, with compositional prompts, we observed that ControlNet tends to prioritize depth image content over textual context, leading to weaker prompt adherence. 

\noindent (b) \emph{CNet-Stop@K}: Here, we use ControlNet for a fixed percentage, $K\%$ of the generation steps, with conditioning on both the depth image and text prompt with category labels as in ControlNet. After this partial conditioning, we capture the latent representation from ControlNet at the $K\%$-intermediate step and pass it to Stable Diffusion. We then continue denoising for the remaining steps with only the text prompt, excluding the depth conditioning. This approach increasingly preserves the shape and target pose for larger $K$, while better adhering to the text prompt context for lower $K$. Given this trade-off, we test for various $K\%$ levels, including $K=20, 40, 60, 80$ ($K=100$ corresponds to the original ControlNet). 

\noindent (c) \emph{ShapeWords}: Our method, ShapeWords, uses text prompts with [SHAPE-ID] tokens as provided in our test datasets. For cases requiring pose control, we apply a similar procedure to CNet-Stop@K: ControlNet is applied for a fixed percentage, $K\%$, of the generation steps, conditioned on the depth image in the target pose and text prompt with a category label in place of [SHAPE-ID]. At the $K\%$-intermediate step, we capture the intermediate latent representation and pass it to Stable Diffusion, where denoising continues conditioned on the Shape2CLIP embedding generated for the text prompt based on the specific [SHAPE-ID] token. This preserves the target pose while allowing for shape and prompt adherence provided by our method. We call this variant of our method as ``ShapeWords@K''.

\paragraph{Numerical evaluation.} We first examine results on the ``simple prompts dataset''. As outlined earlier, this dataset allows us to evaluate shape adherence across all methods using S-CD and S-IOU metrics, which measure geometric similarity between the silhouettes of rendered reference shapes and generated shapes. Since the silhouette comparison is pose-dependent, we compare variants of "ShapeWords@K" with the corresponding "CNet-Stop@K" variants for matching values of $K$.

Figure \ref{fig:geom_eval_quant} shows plots of S-CD and S-IOU, with the horizontal axis representing different $K$ values ($20\%$, $40\%$, $60\%$, $80\%$), and the y-axis reporting the metrics. We also include results for the original ControlNet (i.e., $K=100$), which serves as a strong baseline for shape adherence.

\begin{figure}[t!]
  \centering
\includegraphics[width=1.0\linewidth]{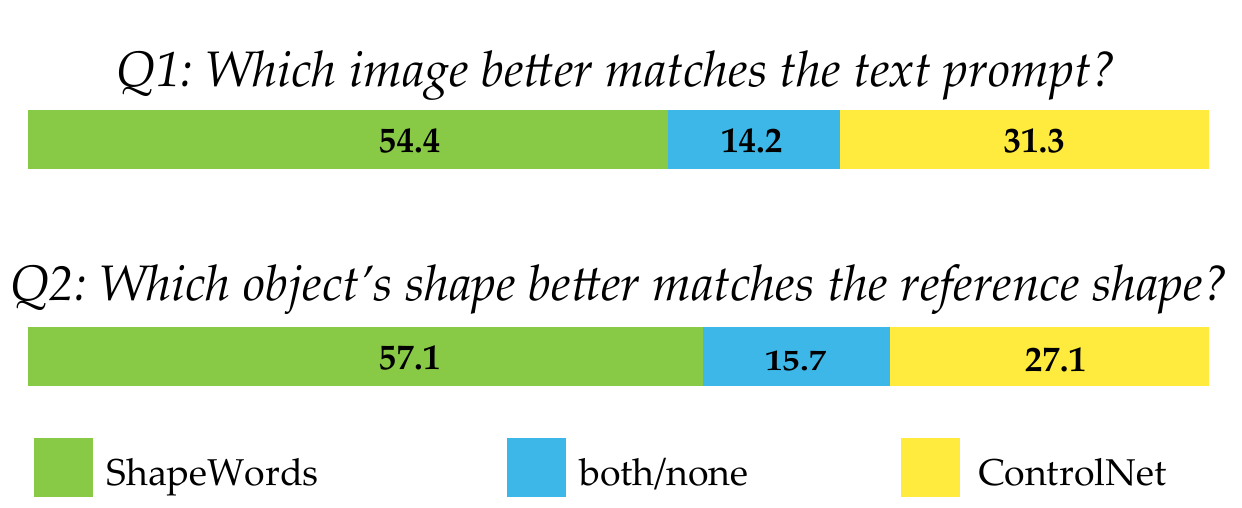}
\vspace{-6mm}  \caption{\textbf{User study results.} We asked 250 participants to compare how well outputs of competing models follow text prompt and target geometry.  Our method was selected as the favorite for both conditions.}
\vspace{-4mm}  \label{tab:ours_vs_theirs_userstudy}
\end{figure}

Our method, ShapeWords@K, consistently outperforms all corresponding ``CNet-Stop@K'' variants, demonstrating superior adherence to the target shape across varying $K$ levels. Notably, at $K=40\%$, ShapeWords@K achieves shape adherence comparable to the original ControlNet. This suggests that ShapeWords can achieve strong shape adherence without explicit depth conditioning. 

A reasonable question from the above comparison is: why not always use the original ControlNet instead of our method? The answer lies in the results shown in Table \ref{table:comp_quant_eval} for the more complex ``compositional prompt dataset.'' In this challenging setting, our method outperforms ControlNet across all measures -- FID, KID, aesthetics, and CLIP scores\footnote{we note that for CLIP score, we compare generated images with the text prompt where [SHAPE-ID] is replaced with the shape's category label.}
-- demonstrating superior prompt adherence and aesthetic plausibility. We note that in this dataset, we do not evaluate silhouette similarity due to the increased foreground complexity and existence of multiple objects. Furthermore, pose control is not enforced: ControlNet generates images using a depth image from a random pose (sampled from 30 viewpoints rotated in 12-degree increments around the vertical axis with fixed elevation, as used in training), while our method generates random poses by design. We also report comparisons with ControlNet-Stop@60 and ControlNet-Stop@80, which, in the previous experiment, demonstrated shape adherence levels similar to our method. However, even with comparable shape adherence, both ControlNet-Stop variants consistently underperform relative to ShapeWords in terms of prompt adherence and overall aesthetic quality.

\input{figures/prompt_shape_comparison}

\paragraph{Perceptual Evaluation.}
As a final evaluation, we conducted two user studies on Amazon MTurk to examine prompt and shape adherence from a perceptual standpoint. In the first study, each page of the questionnaire presented participants with a randomly ordered pair of generated images -- one from ShapeWords and one from ControlNet -- based on a randomly selected prompt from our compositional test dataset and a target shape from our test split. For this comparison, we used the ControlNet variant (CNetStop@K) with the best CLIP score based on our previous experiment. We asked participants: \emph{``Which generated image best matches the text in the given prompt?''}. Participants could pick either image,  specify ``none'' or ''both'' images matched the text well. 
 In the second study, participants were shown randomly ordered image pairs from ShapeWords and the best ControlNet variant according to CLIP score, along with a rendering of the target test shape and a randomly selected prompt from our compositional test dataset. Here, we asked, ``Which object in the generated images is more similar shape-wise to the reference shape in the rendering?''. Participants could again choose either image, or indicate "none" or "both" if both objects matched the reference shape. For both studies, we had $250$ participants each compared $10$ unique image pairs (totaling 2500 comparisons). To ensure, reliability of answers we repeated each question in a random order. 

Figure \ref{tab:ours_vs_theirs_userstudy}  shows the percentage of votes for each question type. Our results indicate that ShapeWords' outputs were preferred by a significantly larger proportion of participants for both prompt adherence and shape resemblance, underscoring the perceptual advantages of our approach.

\paragraph{Qualitative evaluation.}
Our qualitative evaluation (Figure \ref{fig:comp_qual_eval}) also suggests that ControlNet's depth conditioning tends to prioritize shape adherence at the expense of prompt adherence and aesthetic plausibility in the generated images especially for complex prompts. We also include a qualitative comparison with the recent method of Ctrl-X \cite{lin2024ctrlx} in the same figure, where we observed similar, or even worse behavior in terms of shape adherence compared to ControlNet. 
In contrast, our method strikes a much better balance, achieving a more effective trade-off between maintaining shape fidelity and adhering to the input prompt while producing visually plausible results.

We have also observed that our method is able to generalize across various textual contexts and styles as shown in Figure \ref{fig:shapes_and_styles}. We note that our method accurately combines geometric guidance with stylistic prompt cues, like thin object proportions and overly detailed backgrounds of Hieronymus Bosch's paintings. 

\paragraph{Guidance strength.} We demonstrate control of the guidance strength (parameter $\lambda$) in Figure \ref{fig:strengh_ablation},
%. For the fixed input noise for our base diffusion model, we fix prompt and shape and we vary 
where we vary $\lambda$ from $0.0$ to $1.0$. Increase of guidance strength increases the adherence of the resulting image to the target shape. We note that intermediate values of geometry guidance via ShapeWords do also result in plausible shapes. 

Overall, ShapeWords presents an argument for \textit{soft} structural guidance for text-to-image synthesis. In contrast to \textit{hard} structural guidance like depth images, canny edges or renders, it allows users to explore variations of target geometries guided by the learned structural priors. We also illustrate this useful property in Figure \ref{fig:concept_art}, where ShapeWords produces concept art of furniture items that bear both diverse and distinct resemblance to target geometries. Note that this guidance is purely geometric: even though target shapes come from `chair' and `lamp' categories, our model produces tables, chairs, couches, and bookshelves.

\paragraph{Additional evaluation.}
Our supplement includes ablation studies examining various design choices, such as predicting the text prompt embedding directly versus predicting a residual, and updating versus not updating the EOS token.
\input{figures/guidance_strength}
\input{figures/concept_art}

%% file: figures/geom_eval_by_k.tex
\begin{figure}[t]
\centering
\begin{subfigure}[b]{0.48\linewidth}
    \centering
    % Placeholder image and caption for the left image
    \includegraphics[width=\linewidth]{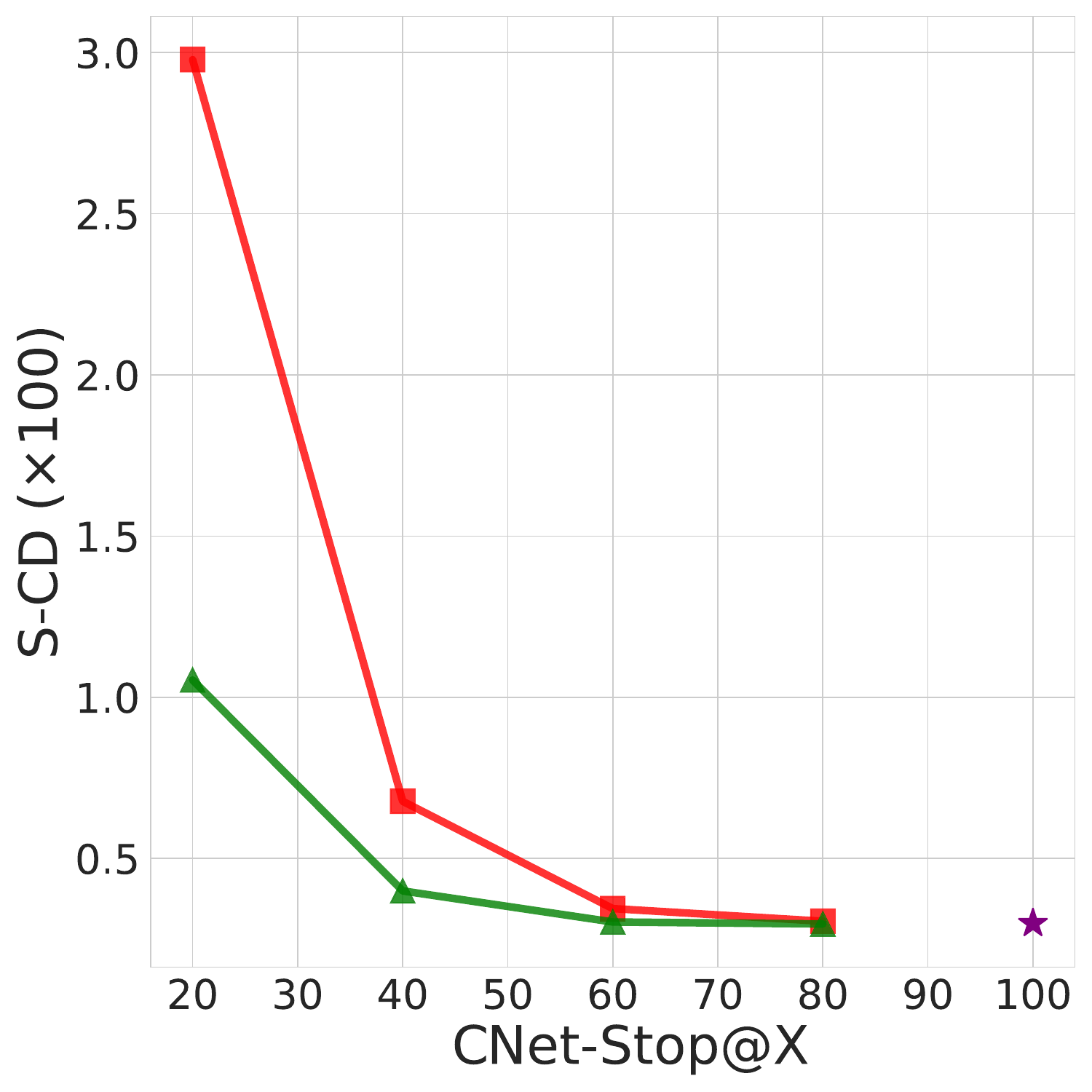}
    %\caption{Silhoutte Chamfer Distance per X}
    \label{fig:left_image}
\end{subfigure}
\hfill
\begin{subfigure}[b]{0.48\linewidth}
    \centering
    % Placeholder image and caption for the right image
    \includegraphics[width=\linewidth]{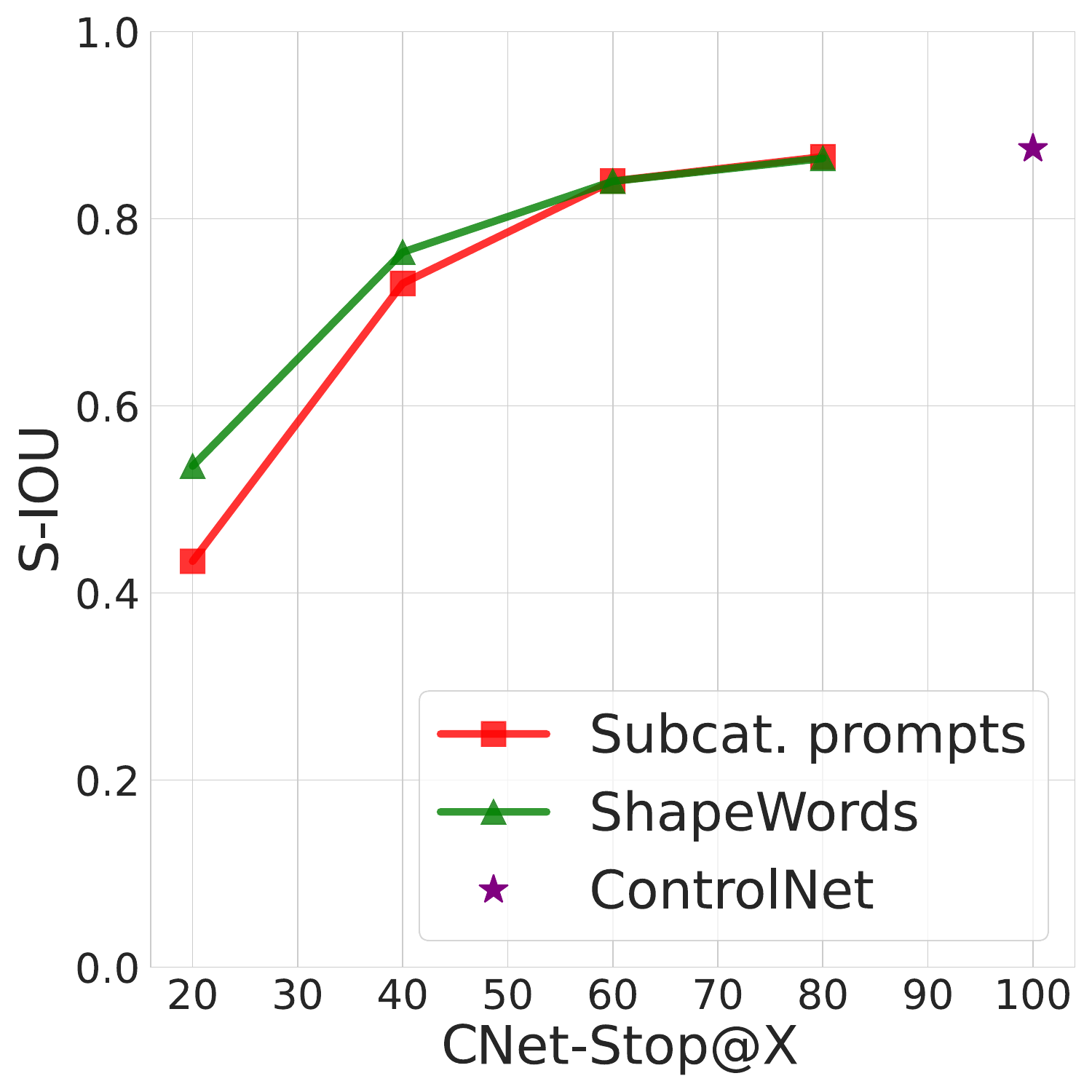}
    %\caption{Placeholder caption for right image}
    \label{fig:right_image}
\end{subfigure}
\vspace{-6mm}
\caption{\textbf{Quantitative evaluation of shape adherence}. ShapeWords@K consistently outperforms the corresponding ``CNet-Stop@K'' variants with text guidance. 
%Notably, at $K=60\%$, ShapeWords@K achieves shape adherence comparable to the original ControlNet. 
``Subcat. prompts'' refers to subcategory prompts (e.g. `a photo of an office chair'), which are the finest annotation level in ShapeNet.}  
\label{fig:geom_eval_quant}
\end{figure}

%% file: figures/comparison_by_k_compact.tex
\newcommand{\imgwidth}{0.11\textwidth}

\begin{figure}[htbp]
\centering
\setlength{\tabcolsep}{1pt} % Adjust horizontal spacing
\renewcommand{\arraystretch}{1} % Adjust vertical spacing
\begin{tabular}{cccc}
%& \multicolumn{2}{c}{\small{ControlNet-Stop@20}} \\
\small{Input} & \small{Category} & \small{Subcategory} & \small{ShapeWords@20} \\
\includegraphics[width=\imgwidth]{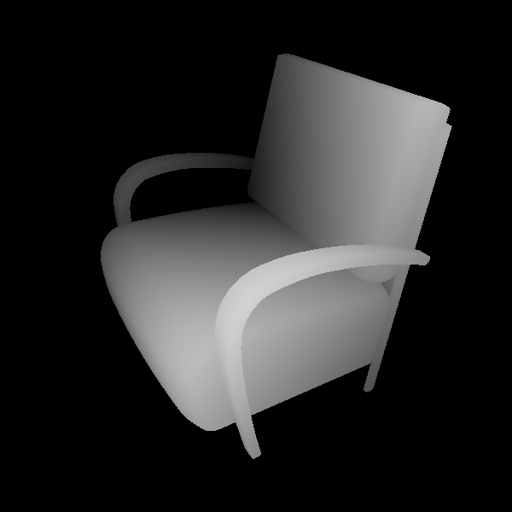} &
\includegraphics[width=\imgwidth]{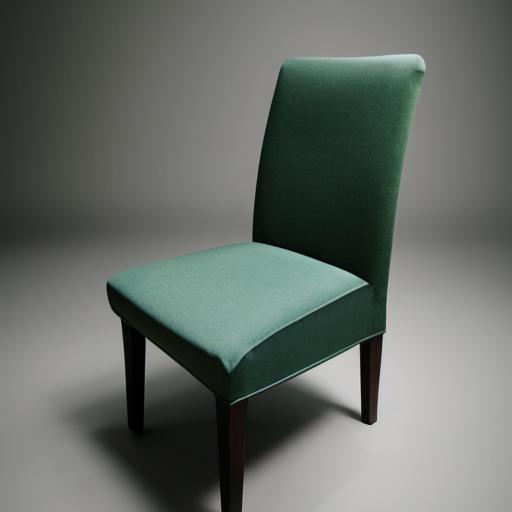} &
\includegraphics[width=\imgwidth]{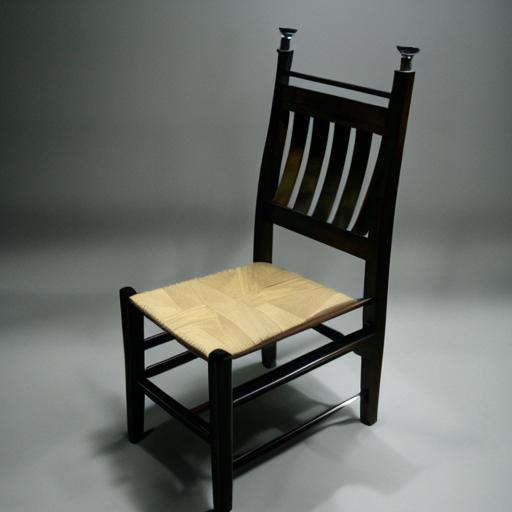} &
\includegraphics[width=\imgwidth]{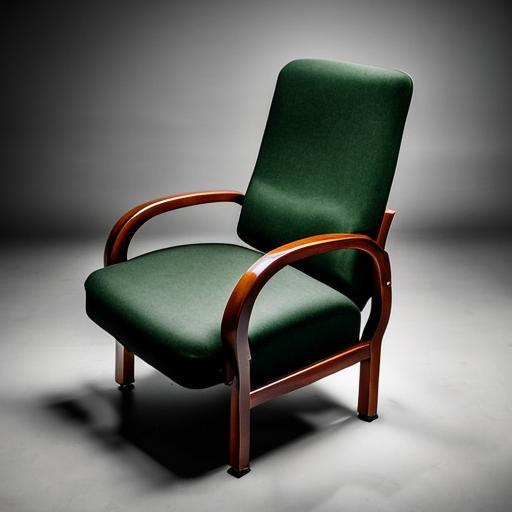} \\

%---------------------
%\includegraphics[width=\imgwidth]{figures/img/comparison_by_k/03593526_c81fa64321c8d49fb9be450193b1790b/060/input_depth.png} &
%\includegraphics[width=\imgwidth]{figures/img/comparison_by_k/03593526_c81fa64321c8d49fb9be450193b1790b/060/cat_k_0.2.jpg} &
%\includegraphics[width=\imgwidth]{figures/img/comparison_by_k/03593526_c81fa64321c8d49fb9be450193b1790b/060/subcat_k_0.2.jpg} &
%\includegraphics[width=\imgwidth]{figures/img/comparison_by_k/03593526_c81fa64321c8d49fb9be450193b1790b/060/ours_k_0.2.jpg} \\
%---------------------
\includegraphics[width=\imgwidth]{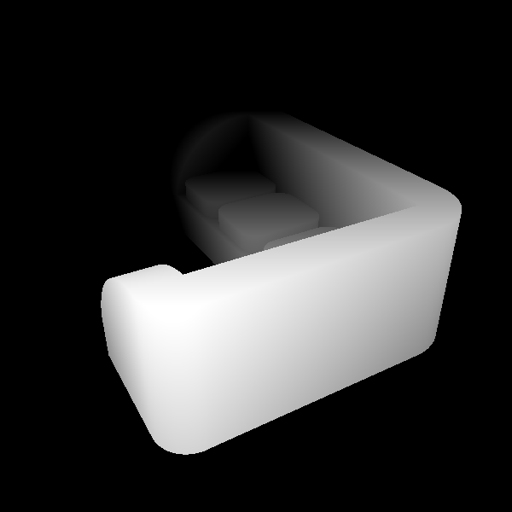} &
\includegraphics[width=\imgwidth]{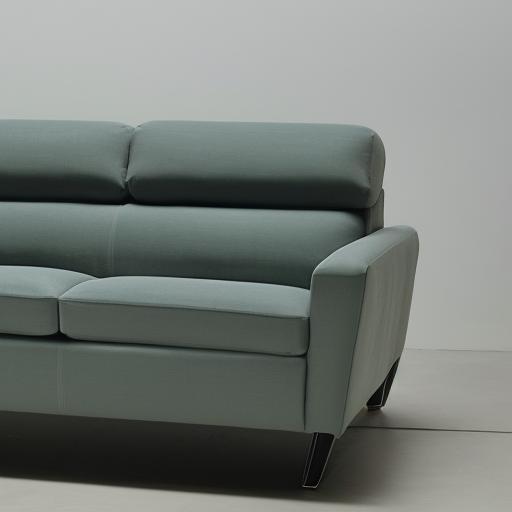} &
\includegraphics[width=\imgwidth]{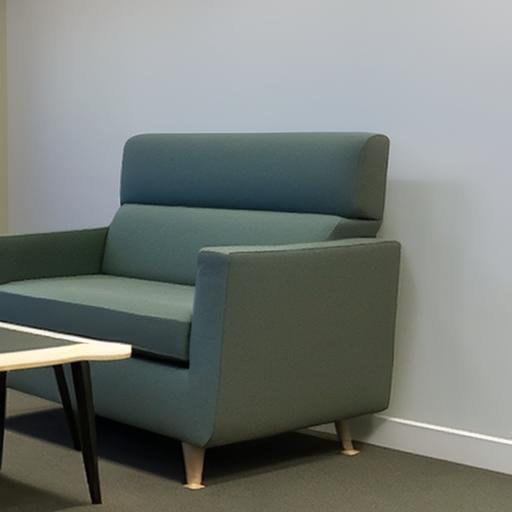} &
\includegraphics[width=\imgwidth]{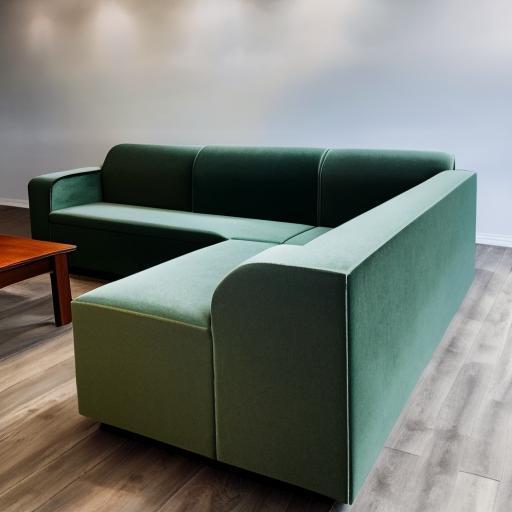} \\
%---------------------
\includegraphics[width=\imgwidth]{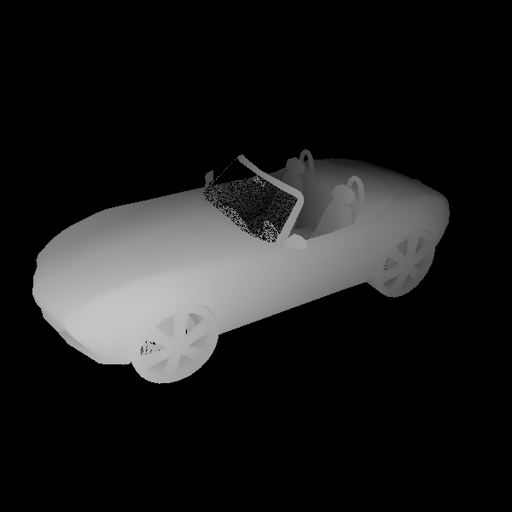} &
\includegraphics[width=\imgwidth]{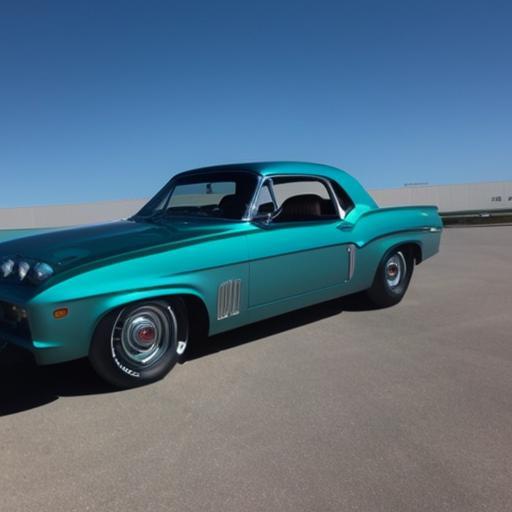} &
\includegraphics[width=\imgwidth]{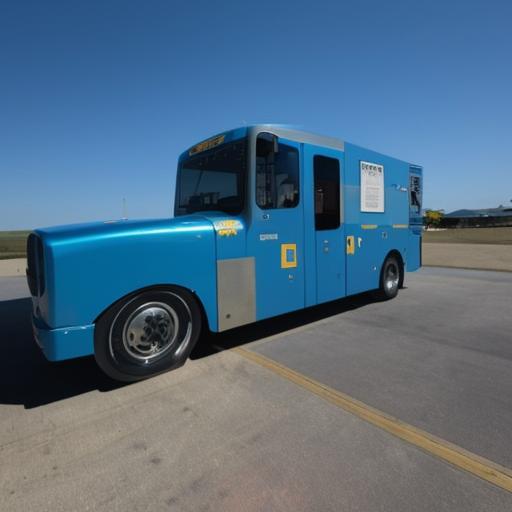} &
\includegraphics[width=\imgwidth]{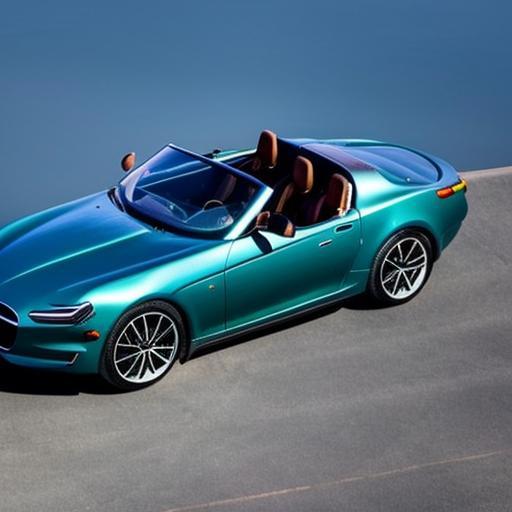} 
\\
%---------------------
%\includegraphics[width=\imgwidth]{figures/img/comparison_by_k/03636649_5b12386df80fe8b0664b3b9b23ddfcbc/120/input_depth.png} &
%\includegraphics[width=\imgwidth]{figures/img/comparison_by_k/03636649_5b12386df80fe8b0664b3b9b23ddfcbc/120/cat_k_0.2.jpg} &
%\includegraphics[width=\imgwidth]{figures/img/comparison_by_k/03636649_5b12386df80fe8b0664b3b9b23ddfcbc/120/subcat_k_0.2.jpg} &
%\includegraphics[width=\imgwidth]{figures/img/comparison_by_k/03636649_5b12386df80fe8b0664b3b9b23ddfcbc/120/ours_k_0.2.jpg} 
%\\

\end{tabular}
\vspace{-3mm}
\caption{\textbf{Shape adherence examples.} Compared to text-based guidance, ShapeWords@20 produces shapes that are significantly more consistent with target geometry, compared to the CNet-Stop@20 conditioned either on category or subcategory prompts. }
\end{figure}

%% file: figures/quant_visual_eval_table.tex
\begin{table}[h]
\centering
\begin{tabular}{l|cccc}
\toprule
\textbf{Model} & \textbf{FID $\downarrow$} & \textbf{KID $\downarrow$} & \textbf{Aes. $\uparrow$} & \textbf{CLIP $\uparrow$} \\
\midrule
ControlNet & 97.0& 10.40& 5.24& 26.9\\
\midrule
CNet-Stop@60     & 90.5& 10.25&  5.20& 28.3\\
CNet-Stop@80     & 92.4& 9.72&  5.17& 27.5\\
ShapeWords           & \textbf{73.8} & \textbf{8.58} &  \textbf{5.45}& \textbf{31.5}\\
\bottomrule
\end{tabular}
\vspace{-3mm}
\caption{\textbf{Evaluation results on compositional prompts.} Our method consistently outperforms over constrained ControlNet variants in the challenging compositional setting. KID metric and CLIP score are multiplied by 100}
\label{table:comp_quant_eval}
\end{table}

%% file: figures/comp_eval_qualitative.tex
\newlength{\imagewidth}
\setlength{\imagewidth}{0.115\textwidth} % Adjust this value as needed

% Define a new column type C based on \imagewidth
\newcolumntype{C}{>{\centering\arraybackslash}m{\imagewidth}}

% Set the color of table lines to grey
%\arrayrulecolor{gray}

\begin{figure*}[ht]
\centering
\setlength{\tabcolsep}{1pt} % Reduce horizontal padding between columns
\begin{tabular}{C*{7}{C}}
%\toprule
% First Row: Empty cell and Shape labels with images
\textbf{\small{Prompt}} & % Empty cell (upper-left corner)
\parbox[c]{\imagewidth}{\centering \textbf{\small{Depth}}} &
\parbox[c]{\imagewidth}{\centering \textbf{\small{CNet-Stop@30}}} &
\parbox[c]{\imagewidth}{\centering \textbf{\small{ControlNet}}} &
\parbox[c]{\imagewidth}{\centering \textbf{\small{CTRL-X@30}}} &
\parbox[c]{\imagewidth}{\centering \textbf{\small{CTRL-X@60}}} &
\parbox[c]{\imagewidth}{\centering \textbf{\small{ShapeWords}}} &
\parbox[c]{\imagewidth}{\centering \textbf{\small{GT Shape}}} \\
%\midrule
% Subsequent Rows: Prompts and generated images
`A \textbf{car} in a snow globe'  &
\includegraphics[width=\imagewidth]{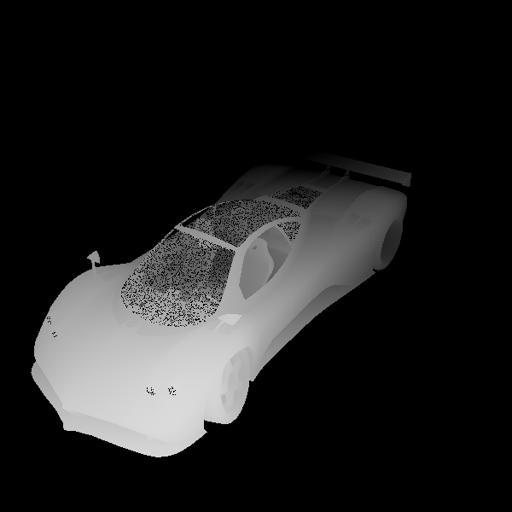} & % Shape 1 image
\includegraphics[width=\imagewidth]{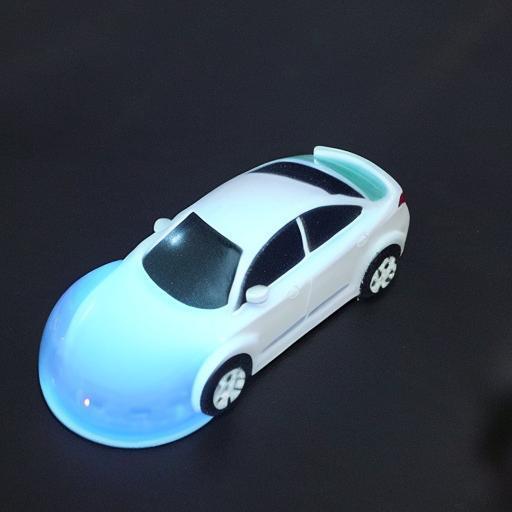} & % Shape 2 image
\includegraphics[width=\imagewidth]{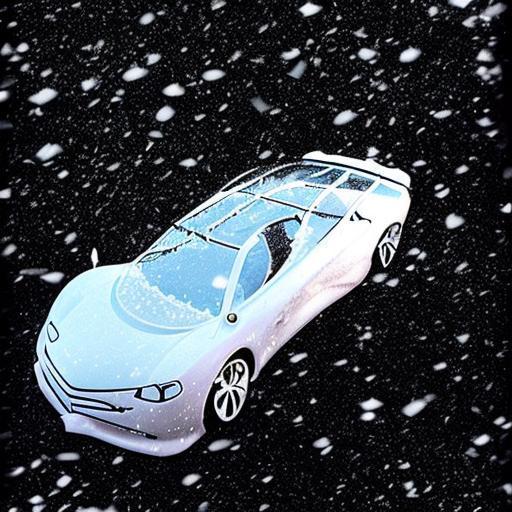} & % Shape 3 image
\includegraphics[width=\imagewidth]{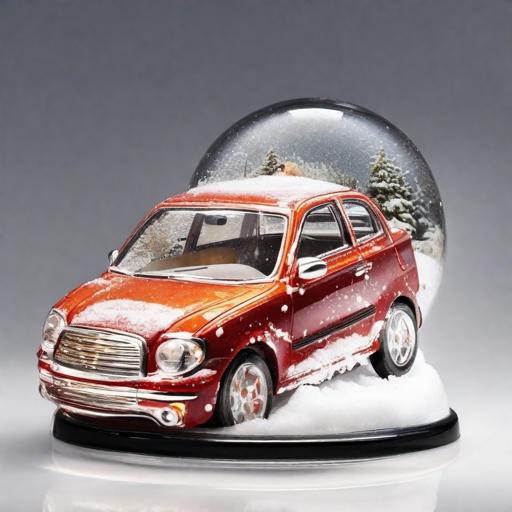} & 
\includegraphics[width=\imagewidth]{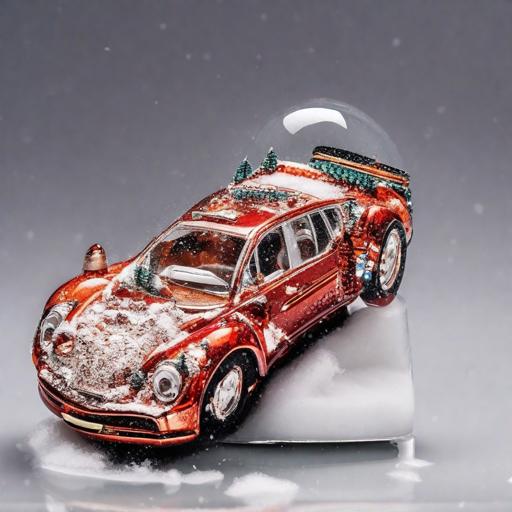} & 
% Shape 4 image
\includegraphics[width=\imagewidth]{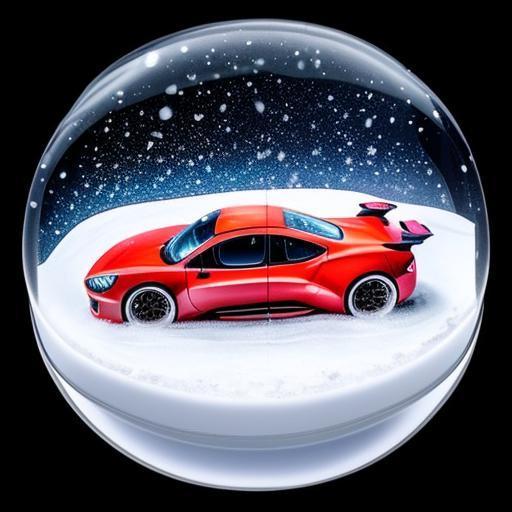} & % Shape 5 image
\includegraphics[width=\imagewidth]{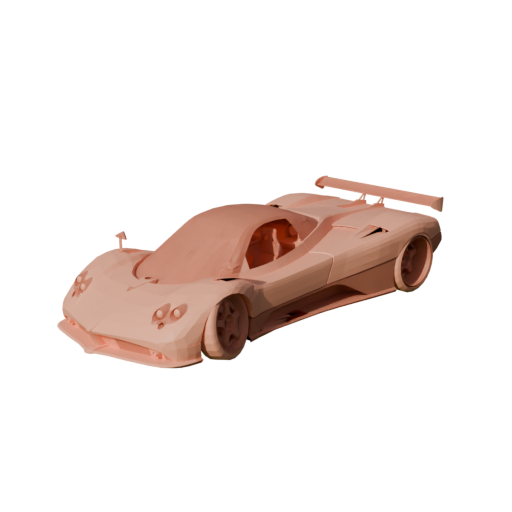}
\\ % Shape 6 image

% Subsequent Rows: Prompts and generated images
`A  \textbf{chair} under a tree'  &
\includegraphics[width=\imagewidth]{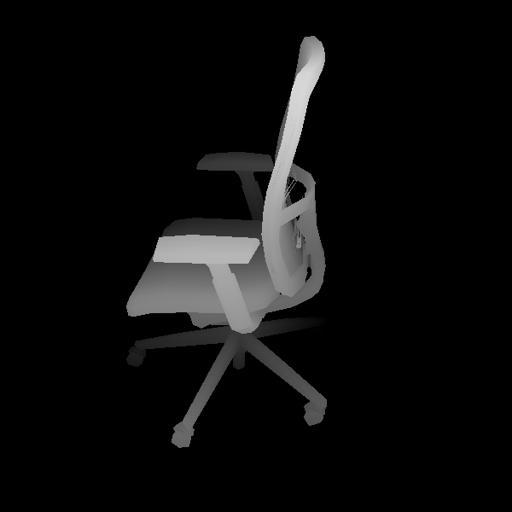} & % Shape 1 image
\includegraphics[width=\imagewidth]{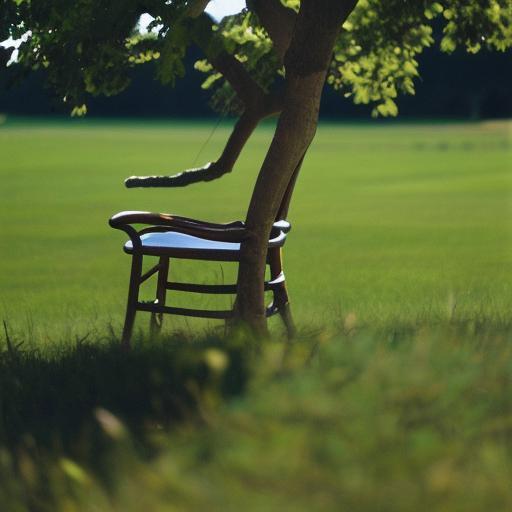} & % Shape 2 image
\includegraphics[width=\imagewidth]{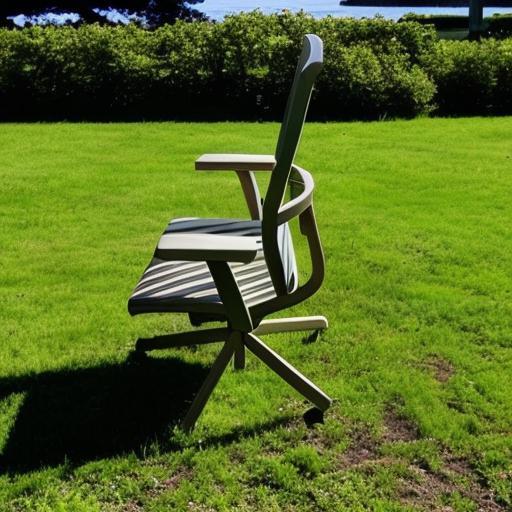} & % Shape 3 image
\includegraphics[width=\imagewidth]{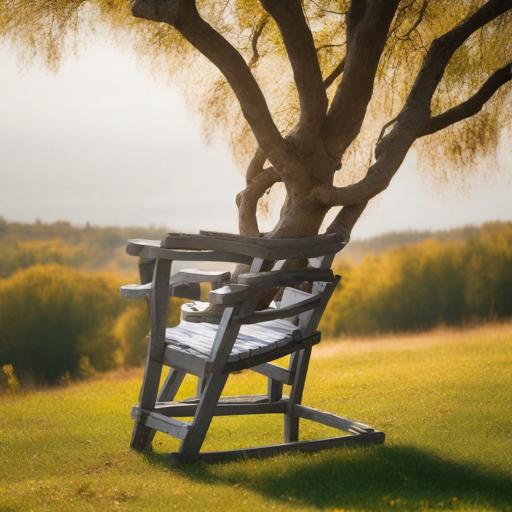} & % Shape 4 image
\includegraphics[width=\imagewidth]{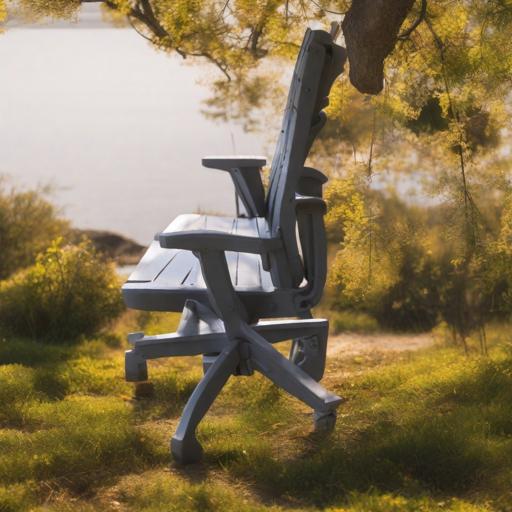} & 
\includegraphics[width=\imagewidth]{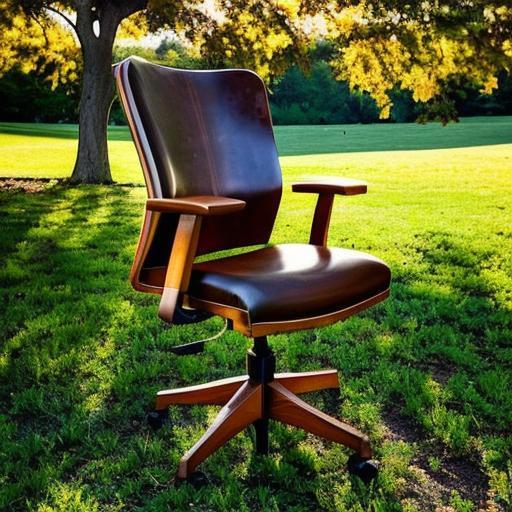} & % Shape 5 image
\includegraphics[width=\imagewidth]{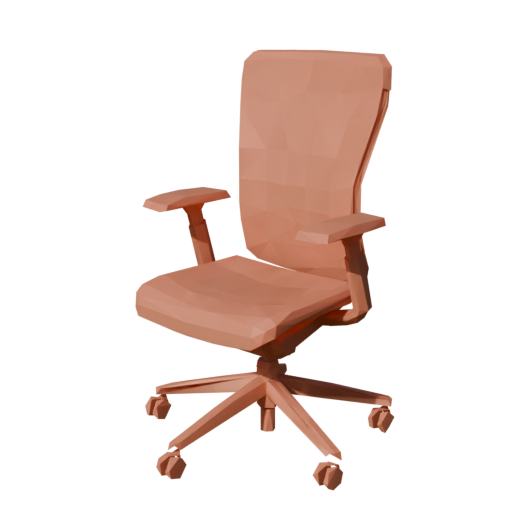}
\\ % Shape 6 image

% Subsequent Rows: Prompts and generated images
`A craftsman working on a \textbf{lamp}'  &
\includegraphics[width=\imagewidth]{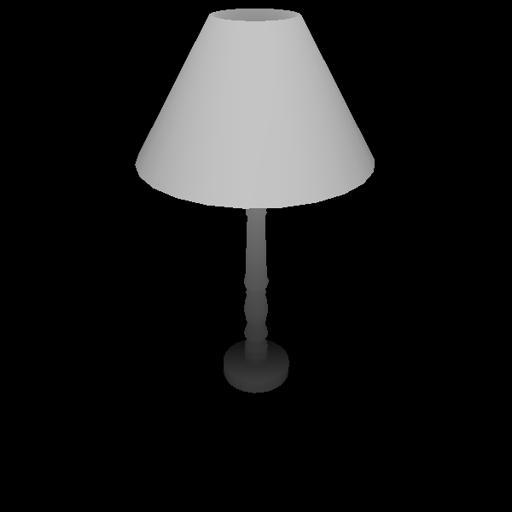} & % Shape 1 image
\includegraphics[width=\imagewidth]{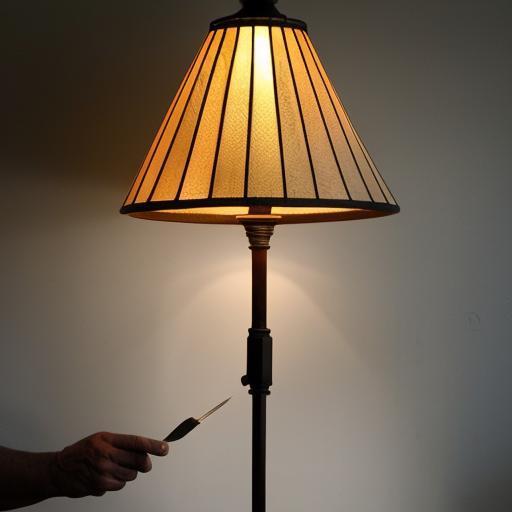} & % Shape 2 image
\includegraphics[width=\imagewidth]{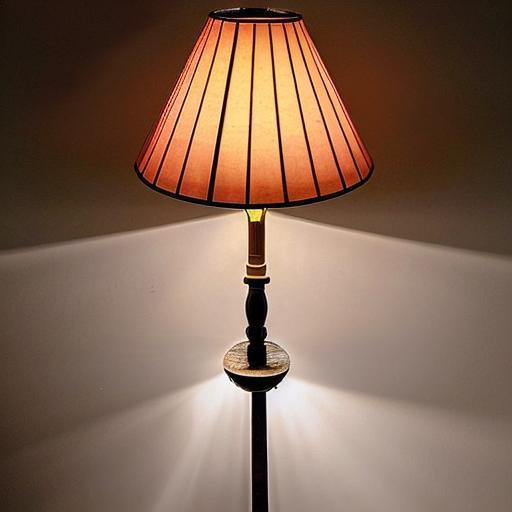} & % Shape 3 image
\includegraphics[width=\imagewidth]{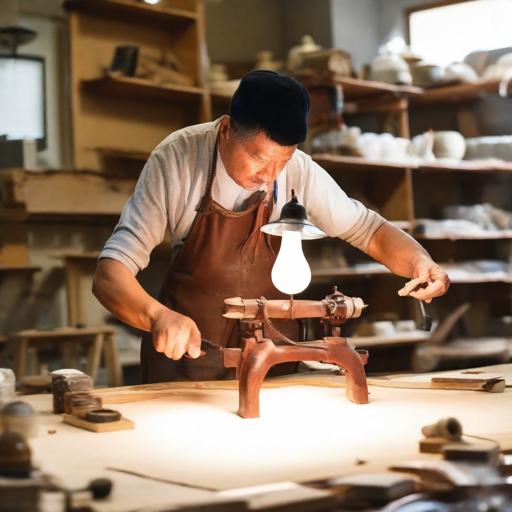} &
\includegraphics[width=\imagewidth]{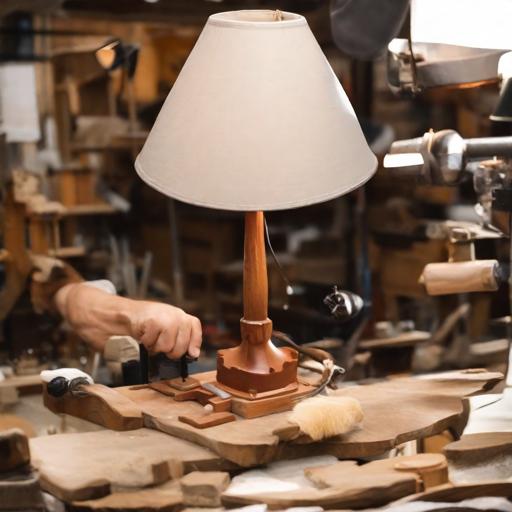} &% Shape 4 image
\includegraphics[width=\imagewidth]{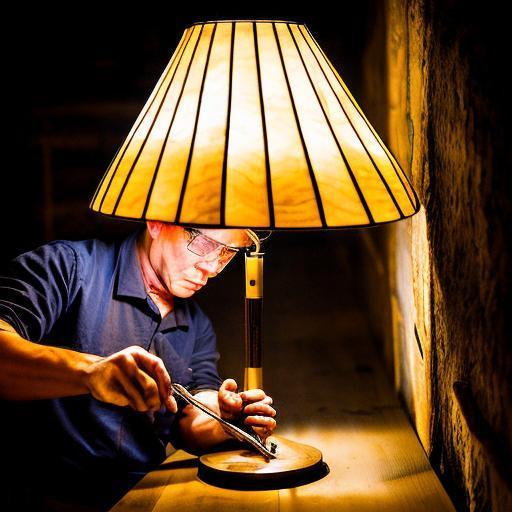} & % Shape 5 image
\includegraphics[width=\imagewidth]{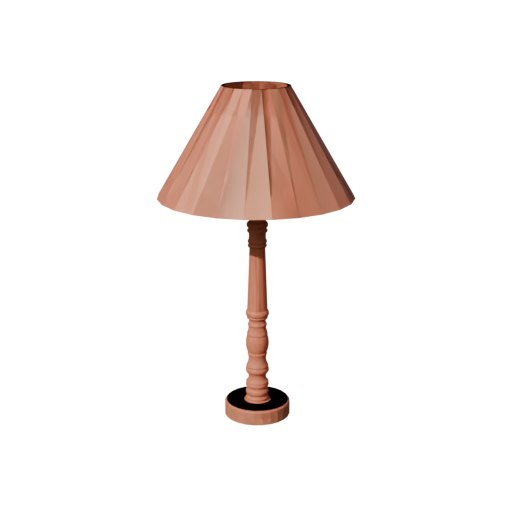}
\\ % Shape 6 image

% Subsequent Rows: Prompts and generated images
`An artist painting a \textbf{table}'  &
\includegraphics[width=\imagewidth]{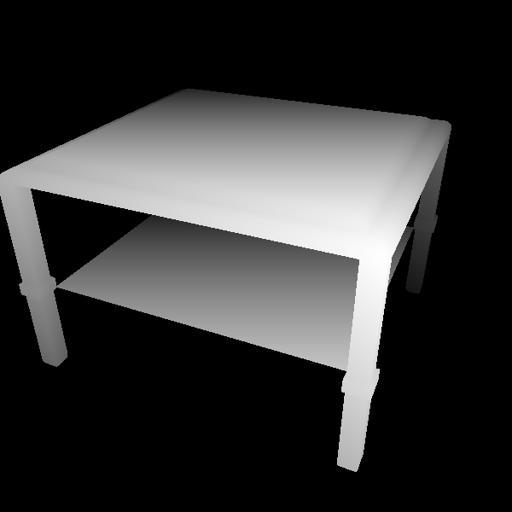} & % Shape 1 image
\includegraphics[width=\imagewidth]{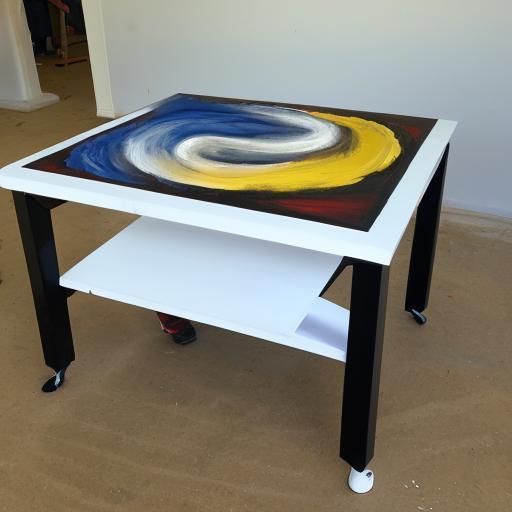} & % Shape 2 image
\includegraphics[width=\imagewidth]{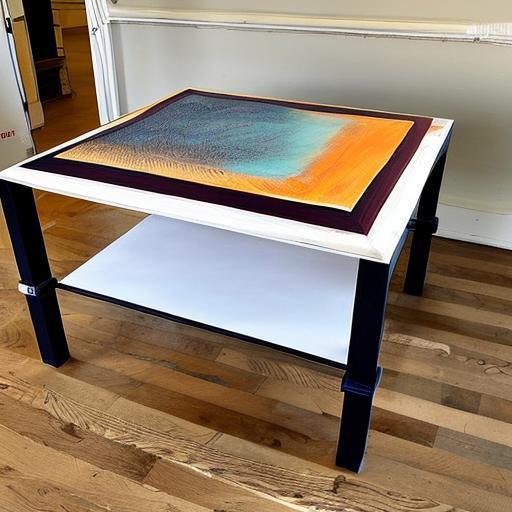} & % Shape 3 image
\includegraphics[width=\imagewidth]{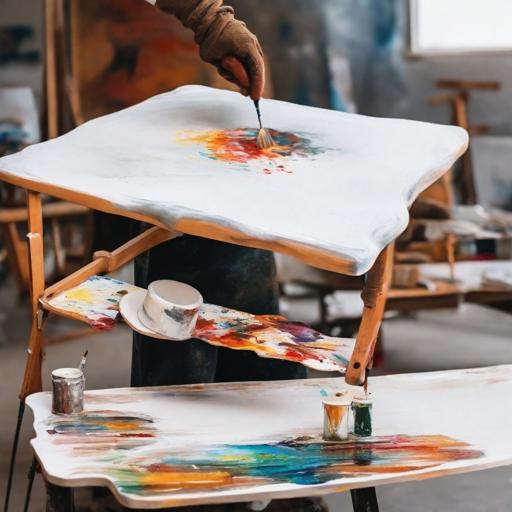} & % Shape 4 image
\includegraphics[width=\imagewidth]{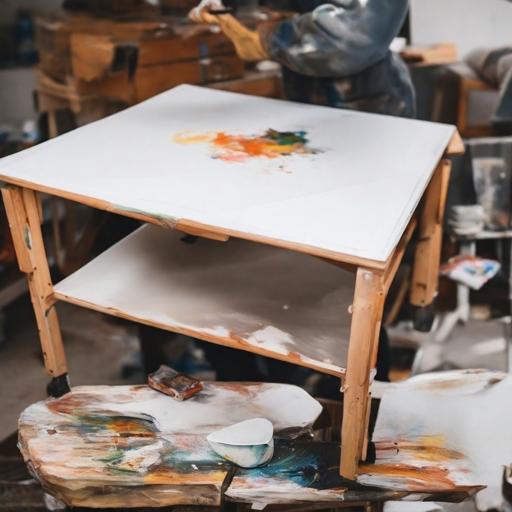} & % Shape 4 image

\includegraphics[width=\imagewidth]{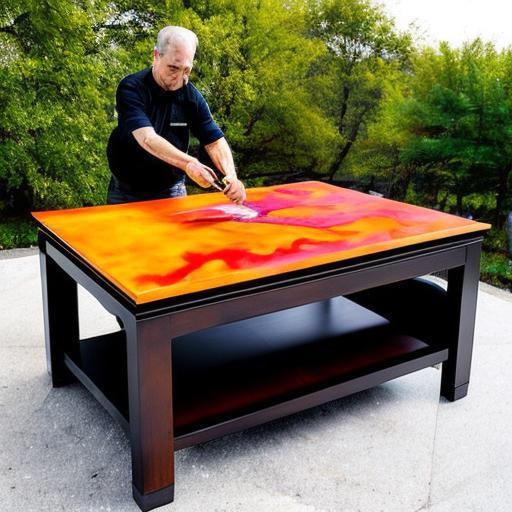} & % Shape 5 image
\includegraphics[width=\imagewidth]{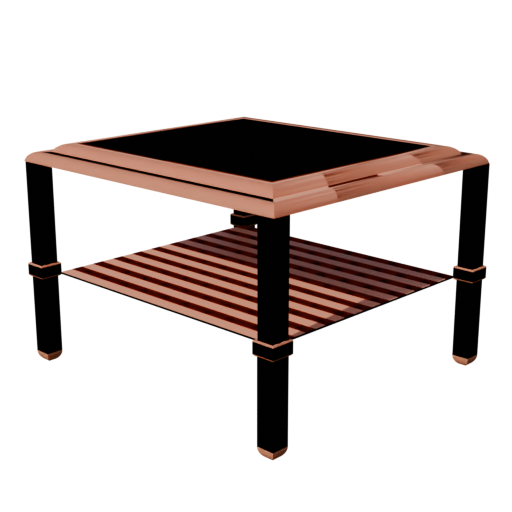}
\\ % Shape 6 image

% Subsequent Rows: Prompts and generated images
%`A toy \textbf{guitar} in a box'  &
%\includegraphics[width=\imagewidth]{figures/img/quality_3/03467517_16afb97fe3882483cf45555a18e7351a/depth.jpg} & % Shape 1 image
%\includegraphics[width=\imagewidth]{figures/img/quality_3/03467517_16afb97fe3882483cf45555a18e7351a/controlnet30.jpg} & % Shape 2 image
%\includegraphics[width=\imagewidth]{figures/img/quality_3/03467517_16afb97fe3882483cf45555a18e7351a/controlnet1.0.jpg} & % Shape 3 image
%\includegraphics[width=\imagewidth]{figures/img/quality_3/03467517_16afb97fe3882483cf45555a18e7351a/ctrl3.jpg} & % Shape 4 image
%\includegraphics[width=\imagewidth]{figures/img/quality_3/03467517_16afb97fe3882483cf45555a18e7351a/ctrl6.jpg} & % Shape 4 image

%\includegraphics[width=\imagewidth]{figures/img/quality_3/03467517_16afb97fe3882483cf45555a18e7351a/ours.jpg} & % Shape 5 image
%\includegraphics[width=\imagewidth]{figures/img/quality_3/03467517_16afb97fe3882483cf45555a18e7351a/render.jpg}
%\\ % Shape 6 image

%\bottomrule
\end{tabular}
\vspace{-3mm}
\caption{\textbf{Generalization to compositional prompts.} Over-constrained depth conditioned baselines (e.g. ControlNet and Ctrl-X@60) ignore prompt composition. Under-constrained depth baselines (e.g. CNet-Stop@30 or Ctrl-X@30) stray too much from target geometry. Our method provides non-superficial generalization to compositional prompts and strong adherence to target shape. We stress out that depth is only provided as input for baselines and \textbf{ShapeWords don't use depth input}. We provide more examples in the supplement.    }
\label{fig:comp_qual_eval}
\end{figure*}

\begin{comment}
\begin{table*}[htbp]
\centering
\setlength{\tabcolsep}{1pt}
\begin{tabularx}{\textwidth}{>{\centering\arraybackslash}m{0.14\textwidth}*{6}{>{\centering\arraybackslash}X}}
\toprule
\textbf{Prompt} & \textbf{Input Depth} & \textbf{ControlNet} & \textbf{ControlNet@30} & \textbf{Ctrl-X} & \textbf{Ours} & \textbf{GT Shape} \\
\midrule
`A photo of an \textbf{object} on a beach' &
\includegraphics[width=\linewidth]{example-image-a} &
\includegraphics[width=\linewidth]{example-image-a} &
\includegraphics[width=\linewidth]{example-image-a} &
\includegraphics[width=\linewidth]{example-image-a} &
\includegraphics[width=\linewidth]{example-image-a} &
\includegraphics[width=\linewidth]{example-image-a} \\

`A photo of an \textbf{object} on a beach' &
\includegraphics[width=\linewidth]{example-image-a} &
\includegraphics[width=\linewidth]{example-image-a} &
\includegraphics[width=\linewidth]{example-image-a} &
\includegraphics[width=\linewidth]{example-image-a} &
\includegraphics[width=\linewidth]{example-image-a} &
\includegraphics[width=\linewidth]{example-image-a} \\
\bottomrule
\end{tabularx}
\caption{Caption of the table.}
\label{tab:example}
\end{table*}
\end{comment}

%% file: figures/prompt_shape_comparison.tex
\newlength{\styleimagewidth}
\setlength{\styleimagewidth}{0.07\textwidth} % Adjust this value as needed

% Define a new column type C based on \imagewidth
\newcolumntype{C}{>{\centering\arraybackslash}m{\styleimagewidth}}

% Set the color of table lines to grey
%\arrayrulecolor{gray}

\begin{figure}[t!]
\centering
\setlength{\tabcolsep}{1pt} % Reduce horizontal padding between columns
\begin{tabular}{C*{6}{C}}
%\toprule
% First Row: Empty cell and Shape labels with images
& % Empty cell (upper-left corner)
%\parbox[c]{\styleimagewidth}{\centering \textbf{Shape 1}\\ \includegraphics[width=\styleimagewidth]{example-image-a}} &
\parbox[c]{\styleimagewidth}{\centering \includegraphics[width=\styleimagewidth]{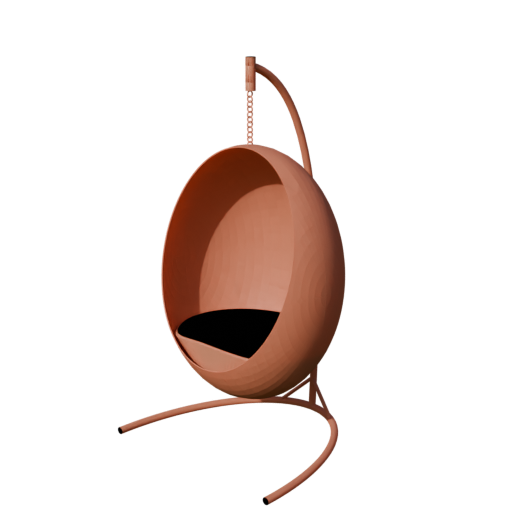}} &
\parbox[c]{\styleimagewidth}{\centering \includegraphics[width=\styleimagewidth]{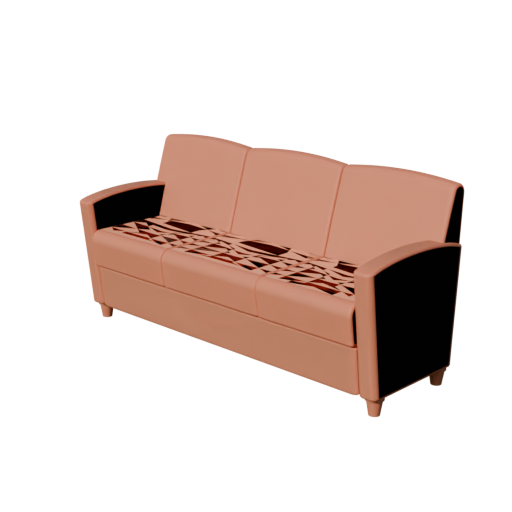}} &
\parbox[c]{\styleimagewidth}{\centering \includegraphics[width=\styleimagewidth]{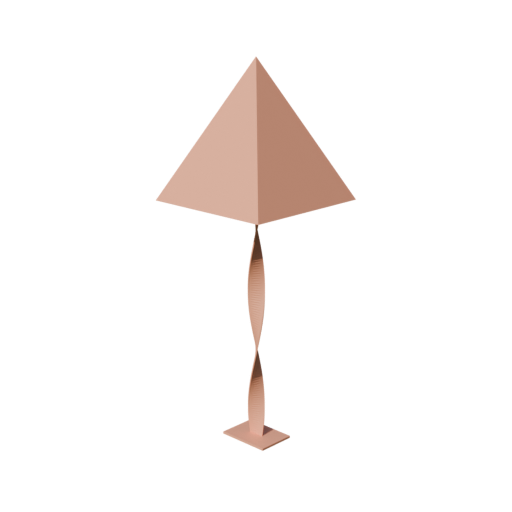}} &
\parbox[c]{\styleimagewidth}{\centering \includegraphics[width=\styleimagewidth]{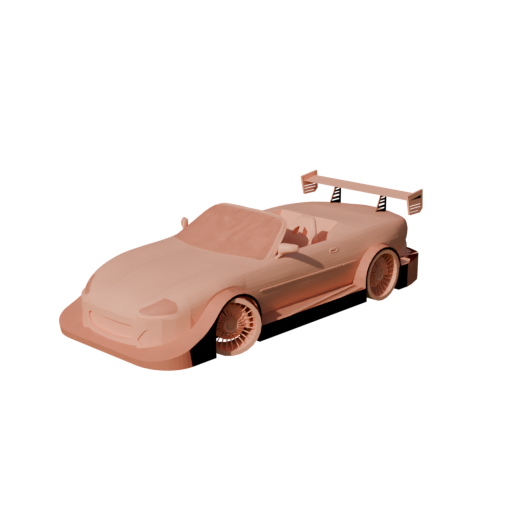}} &
\parbox[c]{\styleimagewidth}{\centering \includegraphics[width=\styleimagewidth]{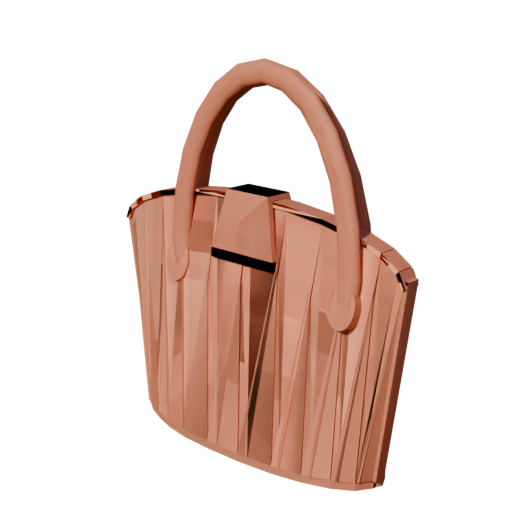}} \\
%\midrule
% Subsequent Rows: Prompts and generated images
\footnotesize{Photo on a beach}  &
\includegraphics[width=\styleimagewidth]{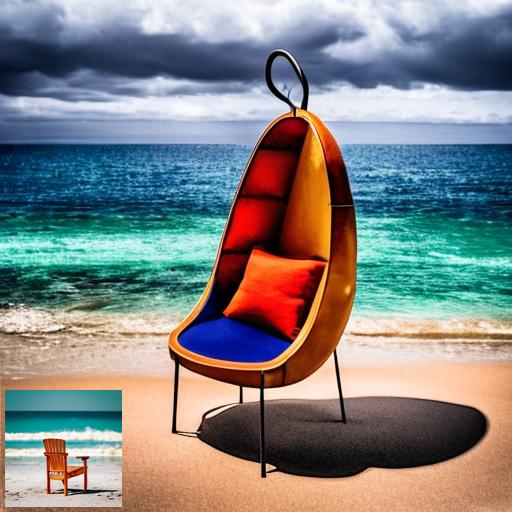} & % Shape 2 image
\includegraphics[width=\styleimagewidth]{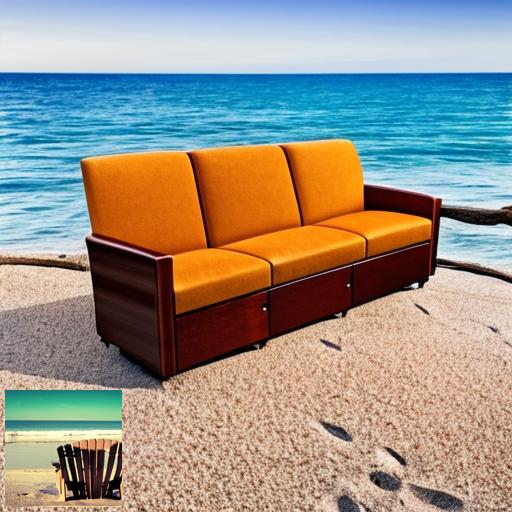} &% Shape 3 image
\includegraphics[width=\styleimagewidth]{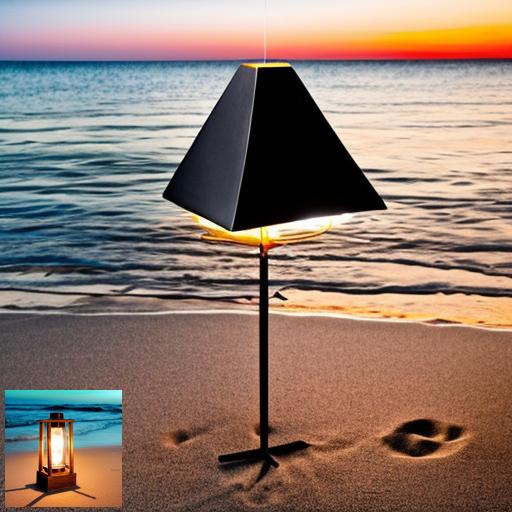} & % Shape 4 image
\includegraphics[width=\styleimagewidth]{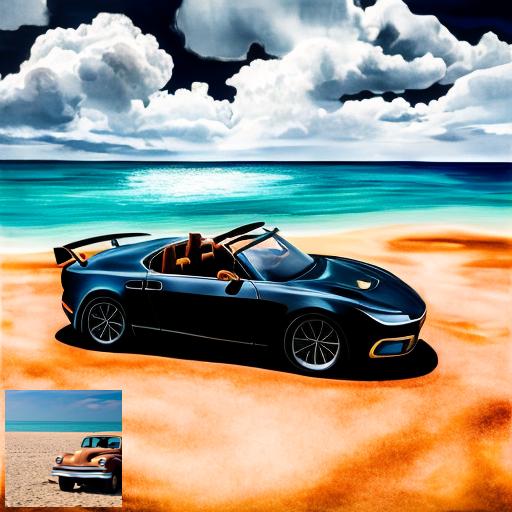} & % Shape 5 image
\includegraphics[width=\styleimagewidth]{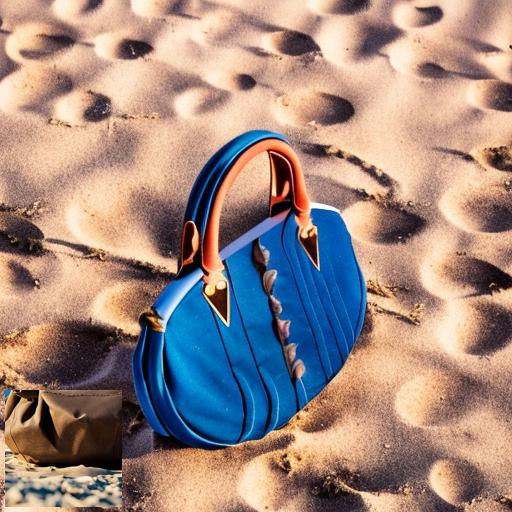}
\\ % Shape 6 image
%\midrule
\footnotesize{Charcoal drawing} &
\includegraphics[width=\styleimagewidth]{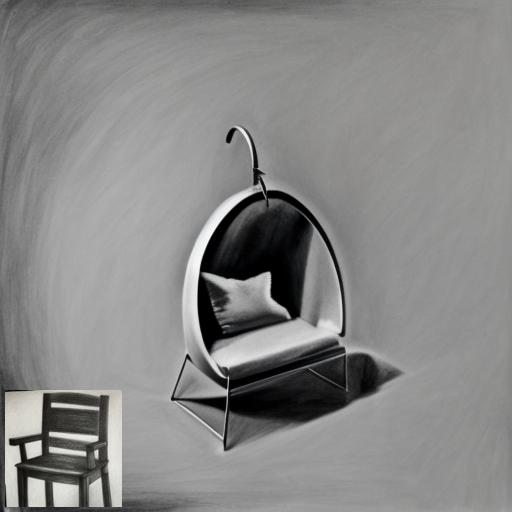} &
\includegraphics[width=\styleimagewidth]{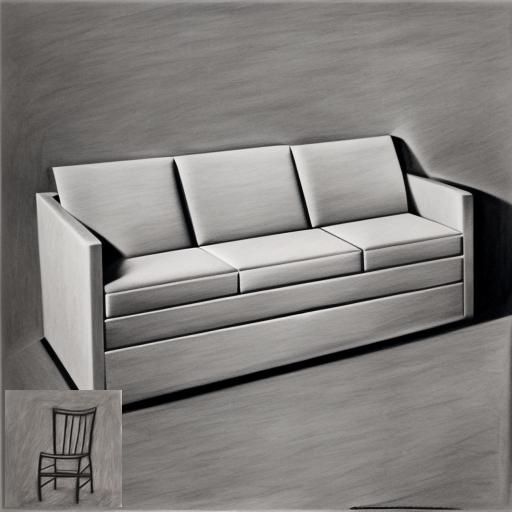} & 
\includegraphics[width=\styleimagewidth]{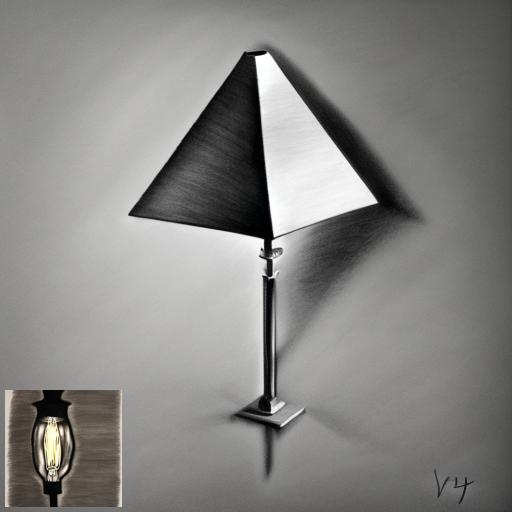} &
\includegraphics[width=\styleimagewidth]{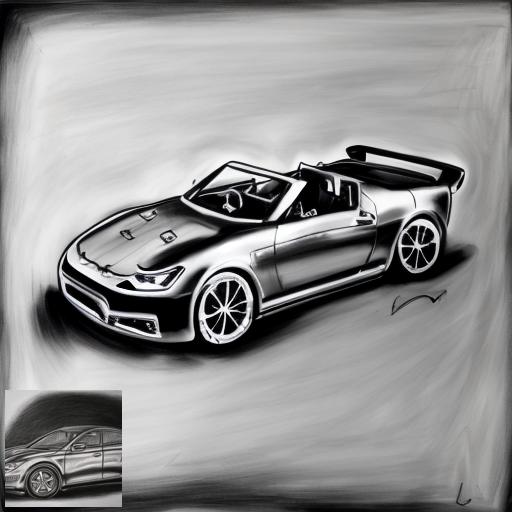} &
\includegraphics[width=\styleimagewidth]{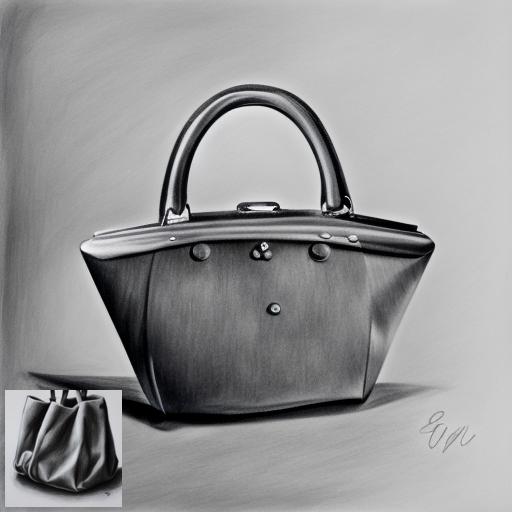} \\
\footnotesize{Hieronymus Bosch's painting} &
\includegraphics[width=\styleimagewidth]{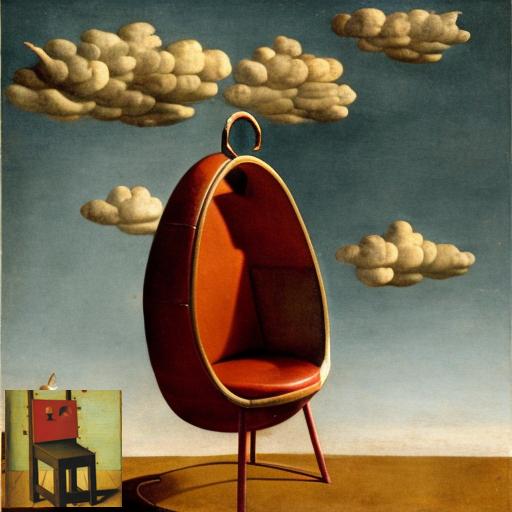} &
\includegraphics[width=\styleimagewidth]{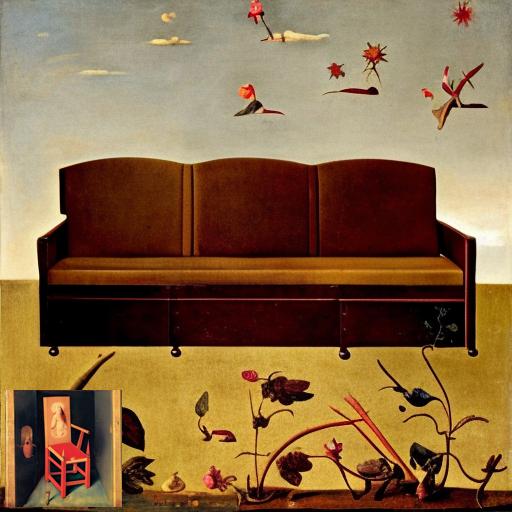} &
\includegraphics[width=\styleimagewidth]{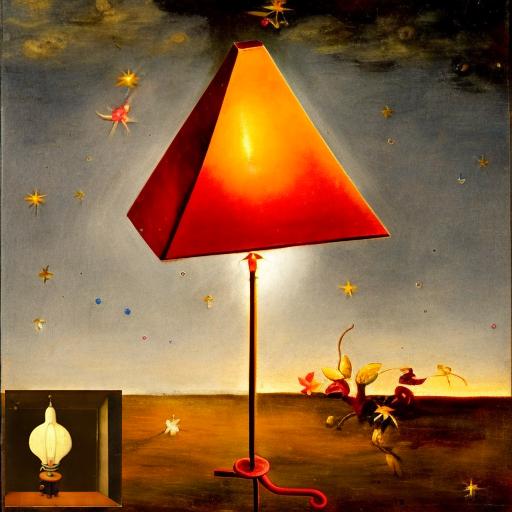} &
\includegraphics[width=\styleimagewidth]{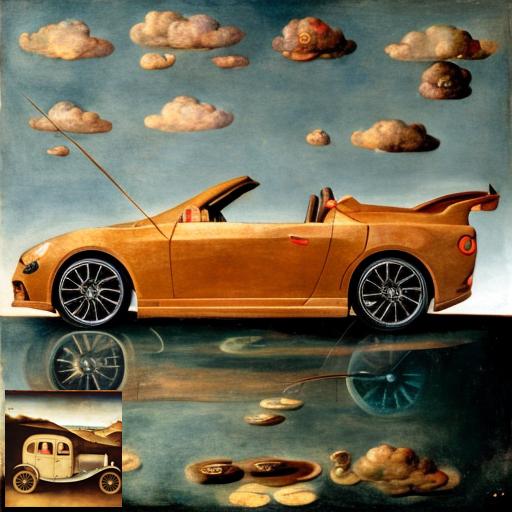} &
\includegraphics[width=\styleimagewidth]{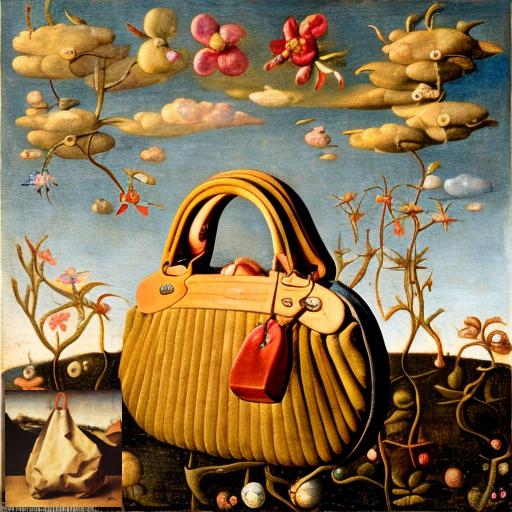} \\
%\bottomrule
\end{tabular}
\vspace{-3mm}
\caption{\textbf{Generalization across shapes and styles.} ShapeWords generalizes across styles and contexts and correctly incorporates stylistic prompt cues into target geometry, if needed. }
\label{fig:shapes_and_styles}
\end{figure}

%% file: figures/guidance_strength.tex
\begin{figure}[t!]
\centering
\setlength{\tabcolsep}{1pt} % Adjust horizontal spacing between columns

% Define column widths directly
\newlength{\controlcolwidth}
\setlength{\controlcolwidth}{0.092\textwidth} % Width of each control column
\newlength{\guidancecolwidth}
\setlength{\guidancecolwidth}{0.092\textwidth} % Width of guidance column

\begin{tabular}{%
    *{4}{>{\centering\arraybackslash}m{\controlcolwidth}}%
    >{\centering\arraybackslash}m{\guidancecolwidth}%
}
% Arrow spanning control columns
\multicolumn{5}{c}{
\begin{tikzpicture}
\begin{scope}[x=1.05\controlcolwidth, y=0.5cm]
\draw[->, thick] (0,0) -- (4,0);
\foreach \x/\label in {0/0, 1/0.33, 2/0.67, 3/1.0}
{
    \draw (\x,0.1) -- (\x,-0.1);
    \node[below] at (\x,-0.1) {\label};
}
\node[below right] at (3.5,-0.1) {\small{strength $\lambda$}};
\end{scope}
\end{tikzpicture}
} \\[2pt]

% First row of images
\includegraphics[width=\linewidth]{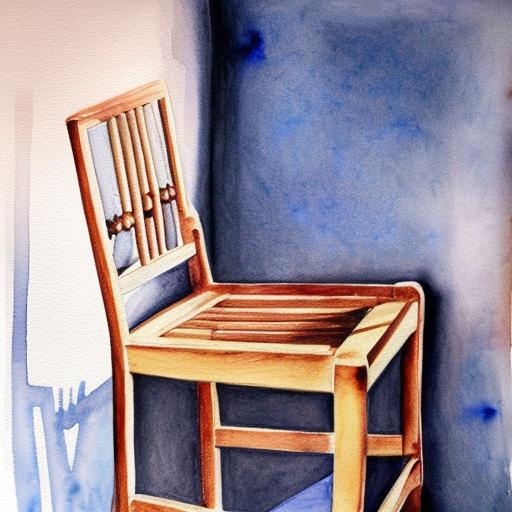} &
\includegraphics[width=\linewidth]{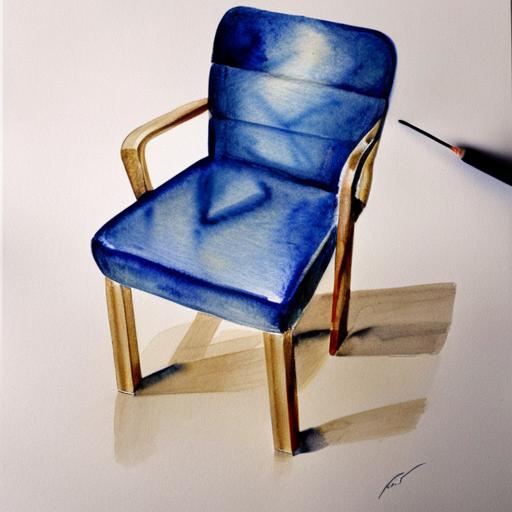} &
\includegraphics[width=\linewidth]{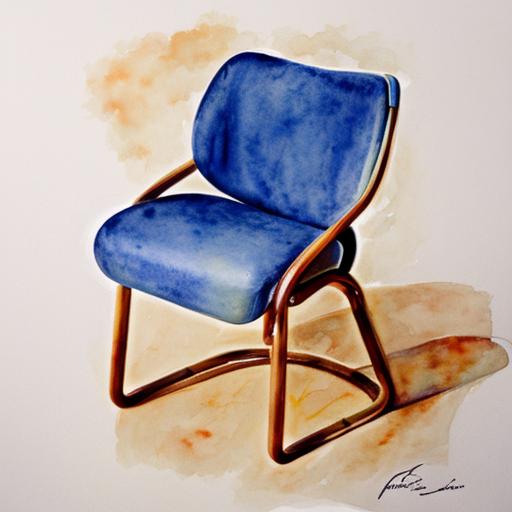} &
\includegraphics[width=\linewidth]{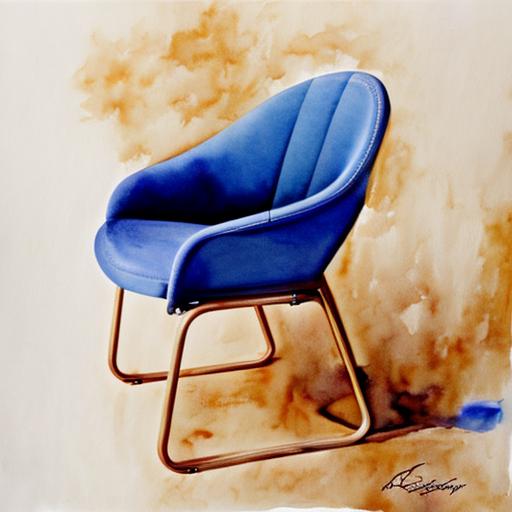} &
\includegraphics[width=\linewidth]{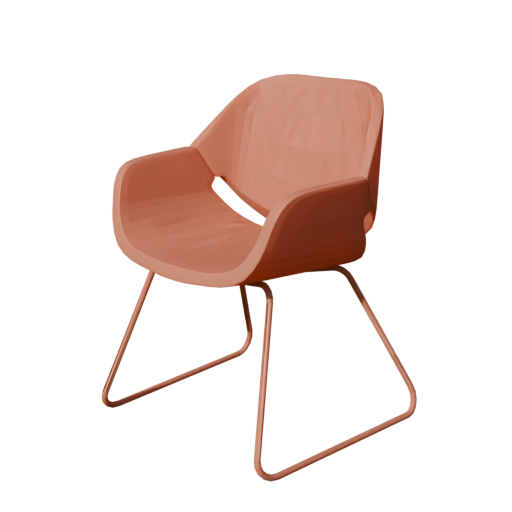} \\[2pt]

% Second row of images
\includegraphics[width=\linewidth]{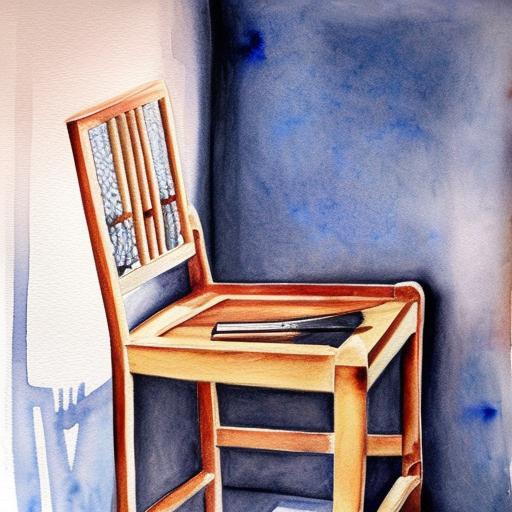} &
\includegraphics[width=\linewidth]{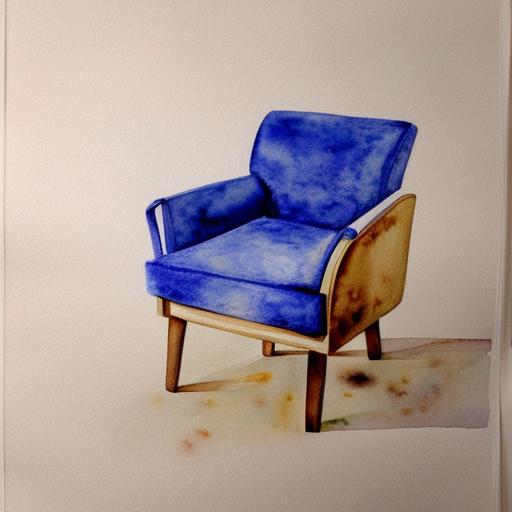} &
\includegraphics[width=\linewidth]{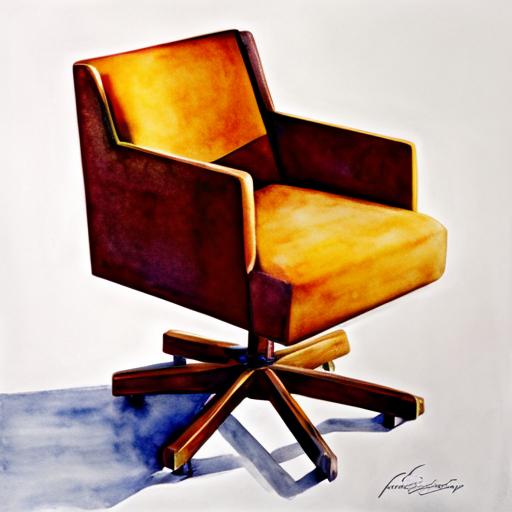} &
\includegraphics[width=\linewidth]{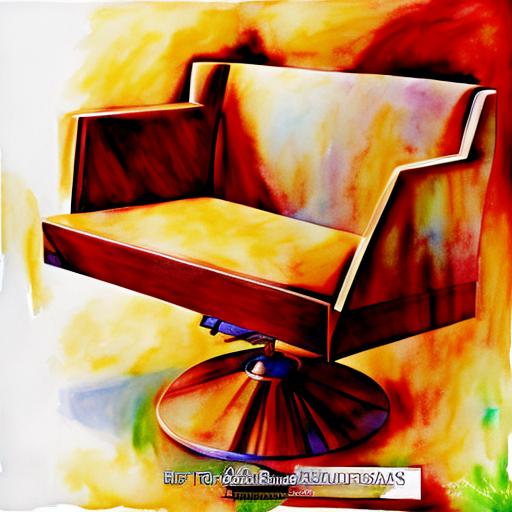} &
\includegraphics[width=\linewidth]{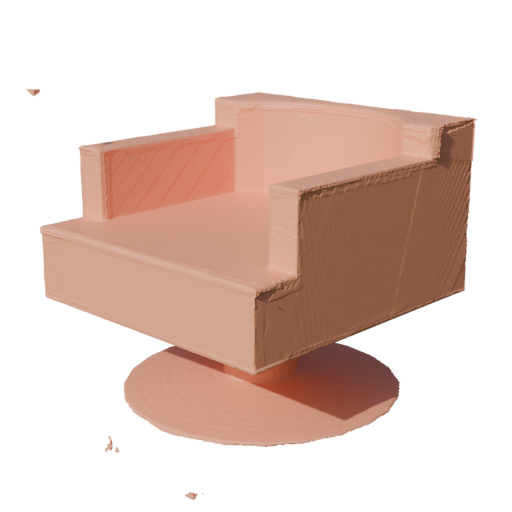} \\[2pt]

% Third row of images
\includegraphics[width=\linewidth]{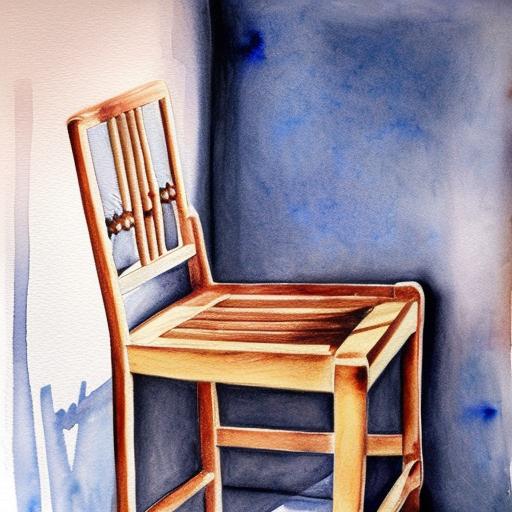} &
\includegraphics[width=\linewidth]{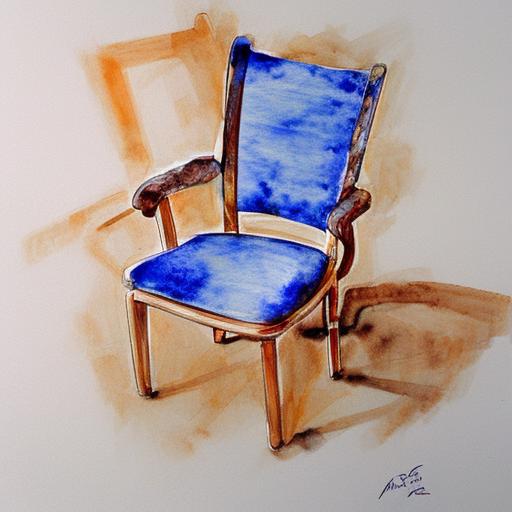} &
\includegraphics[width=\linewidth]{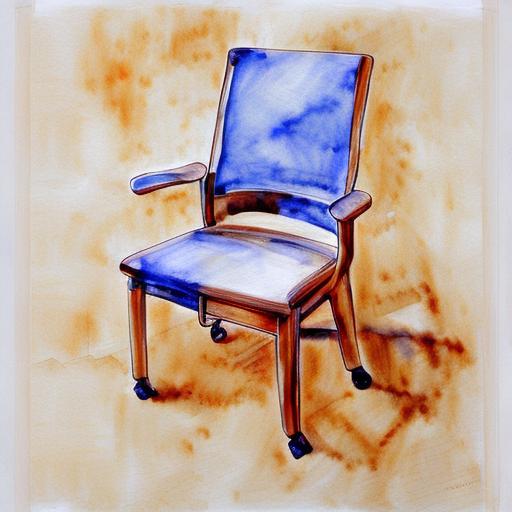} &
\includegraphics[width=\linewidth]{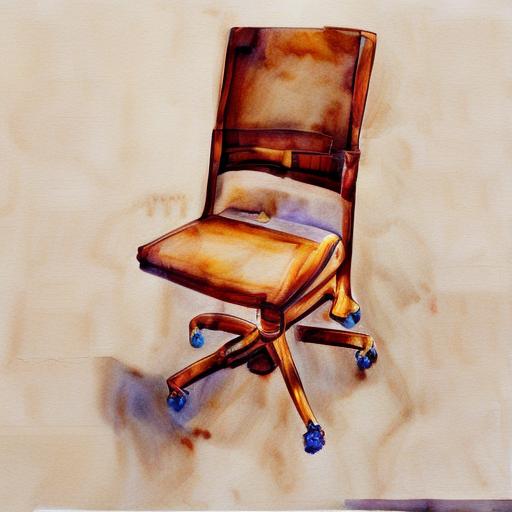} &
\includegraphics[width=\linewidth]{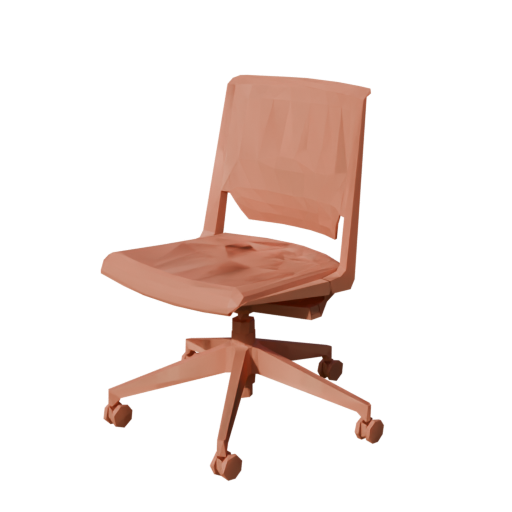} \\

% Add more rows as needed
\end{tabular}
\vspace{-3mm}
\caption{\textbf{Ablation on guidance strength $\lambda$.} Here we vary 
 $\lambda$ from $0$ to $1$ for the prompt  
``aquarelle drawing of a [SHAPE-ID]'' (random seed is fixed, so results for $\lambda=0$ are the same).
 Increasing level of guidance strength increases resemblance to target shape. The intermediate results still remain aesthetically plausible. }
 \vspace{-2mm}
\label{fig:strengh_ablation}
\end{figure}
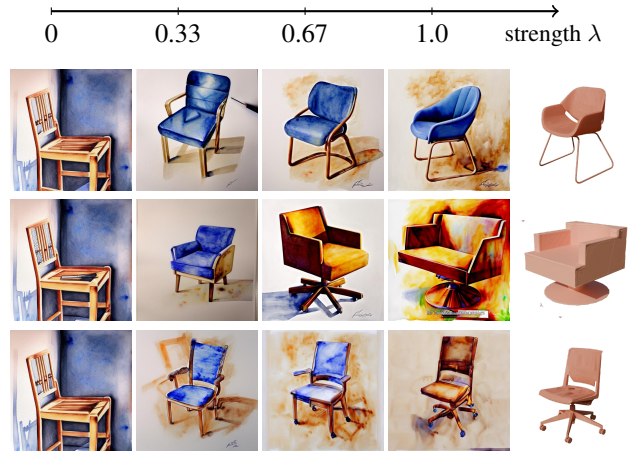

%% file: figures/concept_art.tex
\newcommand{\conceptimgwidth}{0.09\textwidth}

\begin{figure}[t!]
\centering
\setlength{\tabcolsep}{1pt} % Adjust horizontal spacing
\renewcommand{\arraystretch}{1} % Adjust vertical spacing
\begin{tabular}{ccccc}
Guidance & \multicolumn{4}{c}{\small{\textit{`Concept art of futuristic \textbf{furniture}'}}} \\
\includegraphics[width=\conceptimgwidth]{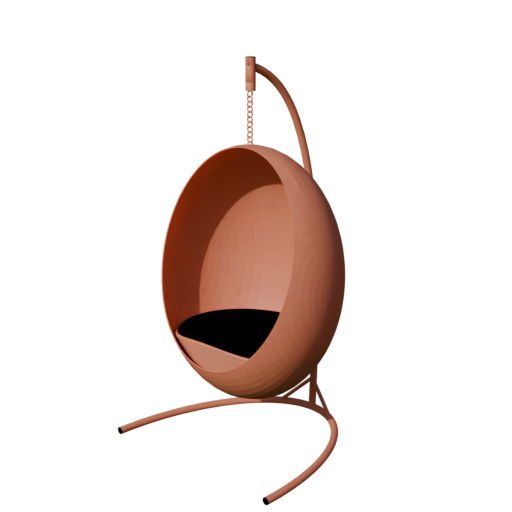} &
\includegraphics[width=\conceptimgwidth]{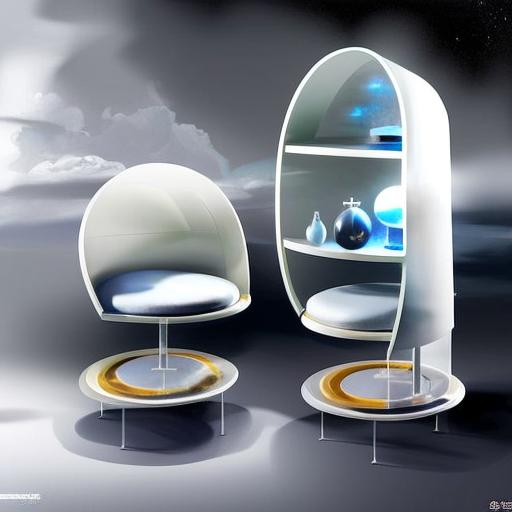} &
\includegraphics[width=\conceptimgwidth]{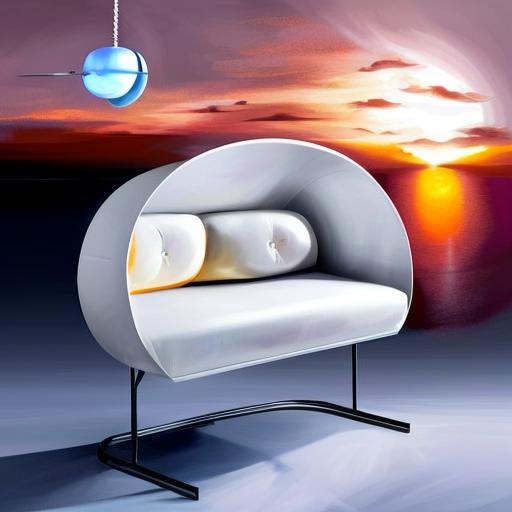} &
\includegraphics[width=\conceptimgwidth]{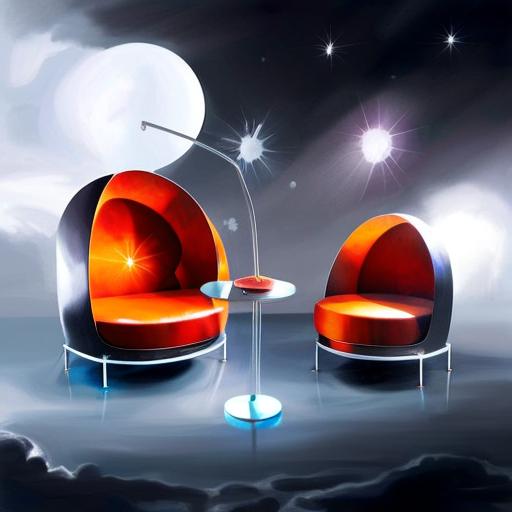}&
\includegraphics[width=\conceptimgwidth]{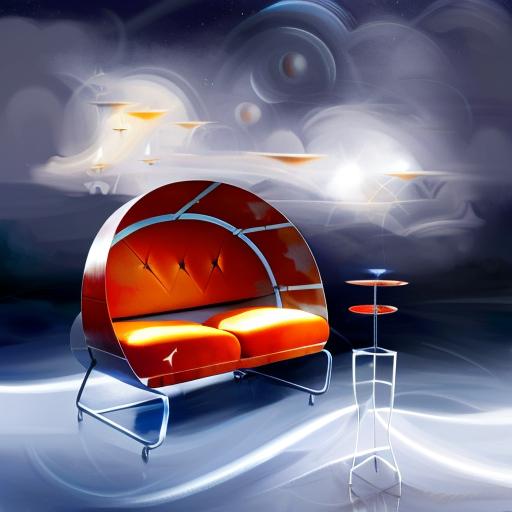} \\

%---------------------
\includegraphics[width=\conceptimgwidth]{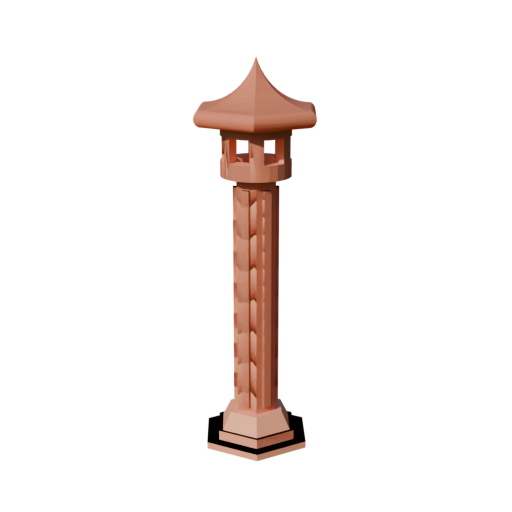} &
\includegraphics[width=\conceptimgwidth]{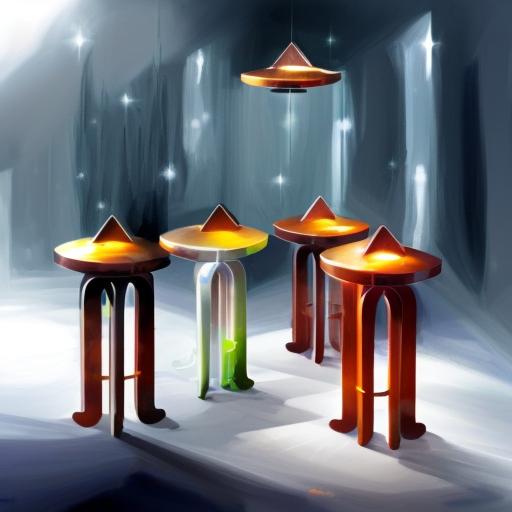} &
\includegraphics[width=\conceptimgwidth]{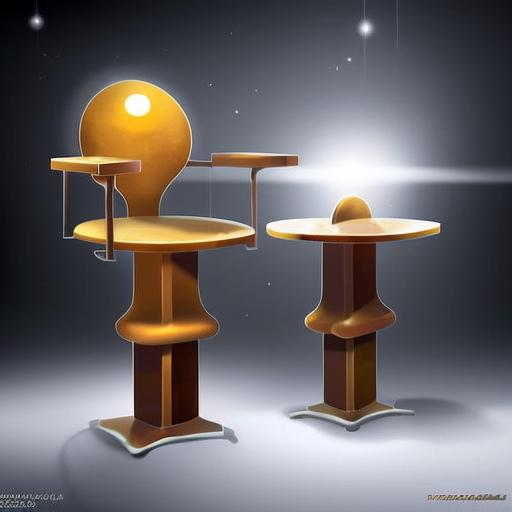} &
\includegraphics[width=\conceptimgwidth]{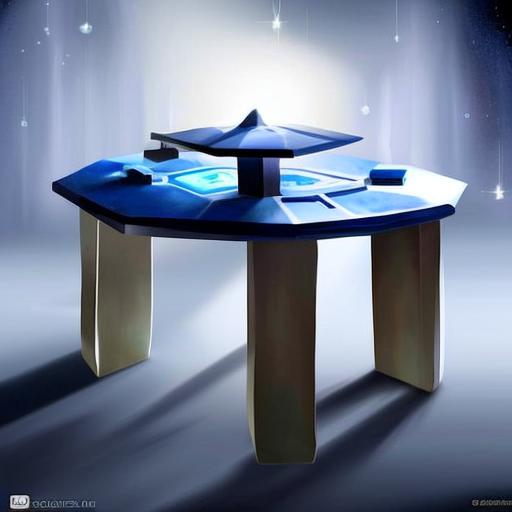} &
\includegraphics[width=\conceptimgwidth]{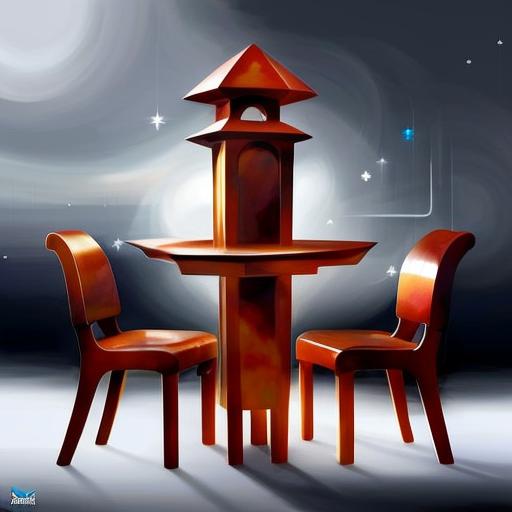} \\
%---------------------

\end{tabular}
\vspace{-3mm}
\caption{\textbf{Soft geometry guidance via ShapeWords.} Our method can produce diverse variations of a target prompt that adhere to guidance geometry via variation of $\lambda$. It can also produce several diverse objects in the image that still resemble the target geometry. }
\vspace{-3mm}
\label{fig:concept_art}
\end{figure}

%% file: sec/5_discussion.tex
\section{Conclusion}

We have presented ShapeWords, the first to our knowledge method that allows for geometric manipulation of text-to-image models via mapping of 3D shapes into the space of text embeddings (CLIP). Our experiments showed that our method shows both strong shape adherence and good generalization to compositional prompts. 
\vspace{-2mm}
\paragraph{Limitations.} Our supplement shows failure cases for our method. In summary, it struggles with  capturing challenging fine-grained geometry, especially in areas of thin parts. Generalization to much more complex prompts would also requiring training on much larger scale datasets. Finally, our method exhibits certain appearance biases (e.g. oversaturation of some images, or preference of certain colors), which could be avoided with better score distillation variants. Combining our shape tokens with viewpoint and other 3D attribute tokens would also be an interesting future direction \cite{viewneti,cheng2024learning}.

\paragraph{Acknowledgements}
We thank Vikas~Thamizharasan, Cecilia Ferrando, Deep Chakraborty and Matheus Gadelha for providing useful feedback and input to our project. This project has received funding from the European Research Council (ERC) under the Horizon research and innovation programme (Grant agreement No. 101124742). We also acknowledge funding from the EU H2020 Research and Innovation Programme and the Republic of Cyprus through the Deputy Ministry of Research, Innovation and Digital Policy (GA 739578).

%% file: sec/X_suppl.tex
\clearpage
\setcounter{page}{1}
\maketitlesupplementary

\section{Implementation details}

\subsection{Data generation details}

For depth images, we used the inverted ShapeNet data provided by the ULIP authors \cite{xue2023ulip}. As stated in the main paper, for each depth image, we applied a randomly selected
prompt from a  set of $13,716$ prompts for ControlNet conditioning. The ControlNet-based generation was done for $50$ steps with control strength of $2$. To promote adherence to shape geometry and reduce appearance biases during training, we additionally used the Stable Diffusion 2.1 inpainting model \cite{rombach2022stable} for 50 steps to modify the backgrounds while preserving the foreground objects. The inpainting strength was set to $0.5$. 

\subsection{SDS weighting function}

For the SDS optimization loss of Eq. 6, we use the weighting function \(W(t)\), proposed by DreamTime~\cite{huang2024dreamtime}, that enhances training stability:
$$
W(t) = \frac{1}{Z}\, \sqrt{\frac{1 - \hat{\alpha}_t}{\hat{\alpha}_t}}\, \exp\left( -\frac{(t - m)^2}{2 s^2} \right),
$$
where \(m\) and \(s\) are hyperparameters controlling the weight distribution at each time step; $\hat{\alpha}_t$ is the noise scale for step $t$; and \(Z\) is a normalization constant ensuring that the weights sum to one over all timesteps. We set \(m = 500\) and \(s = 250\), which provide a good balance between high-frequency details (fine geometry) and low-frequency details (coarse geometry).

\subsection{Training}

We trained the model for $55$ epochs on four NVIDIA A5000 GPUs with batch size 24 per GPU. The learning rate was set to $0.0005$ with $1,000$ warm-up steps to help stabilizing the training process. Similarly to textual inversion pipelines, we randomly crop and resize the training images to prevent overfitting of the model to spatial positions. The maximum scale of the crop was set to $0.8$. 

\input{figures/eos_ablation}

\input{figures/failure_cases_geometry}

During training, the guidance prompt delta $\deltaT$ is applied to \textit{all 77 word embeddings} (padding was set to max sequence length). We empirically found that this strategy during training helps the model to better generalize  compared to adding the guidance delta to the object and EOS tokens only. We suspect that the usage of deltas on all token embeddings during training helps the model to diffuse training appearance biases across all tokens, which in turn reduces the overall appearance biases distilled in the object and EOS tokens. 
%Exploration of different training strategies for our method is a promising direction for future work.

\input{figures/failure_prompts}

\section{Running times}

We note that ShapeWords and ControlNet-based baselines rely on the same Stable Diffusion model (Stable Diffusion 2.1 base) and have similar computation requirements at test time: given a text prompt or/and depth image, it takes a few seconds to generate an image with $100$ diffusion steps on a single GPU: $6.79$s for ControlNet; and $5.00$s for Stable Diffusion 2.1 with ShapeWords. The forward pass of the Shape2CLIP module takes $0.003$s with pre-computed PointBERT embeddings. All measurements were done on single A40 GPU with batch size $1$ and based on an average across 20 runs. For details on PointBERT computational costs, we refer to \cite{yu2022pointbert}.

\section{Token replacement strategies}

We qualitatively compare token replacement strategies in Figure \ref{fig:supp_role_of_eos} at test time. Adding the guidance prompt delta $\deltaT$ to all tokens in the prompt yields overly smooth images that do not adher well to stylistic cues provided in the text or the target geometry. Adding $\deltaT$ only to object token without addition to the EOS token results in good geometry but still poor adherence to the stylistic cues in the prompt. Conversely, modifying only the EOS token results in good stylistic adherence but poor geometry. The strategy described in the main text, which is to add $\deltaT$ to both the object and EOS tokens, yields the best balance of textual and target shape adherence.

\section{Additional quantitative results}

We provide additional quantitative results in Table \ref{table:supp_quant_eval}. Our model consistently outperforms ControlNet-Stop@K variants in terms of aesthetic score. In terms of CLIP score, we outperform all ControlNet-Stop@K variants, except for ControlNet-Stop@30 that
matches the CLIP score of our method. Yet, as we discussed in our experiments in the ``simple prompts dataset'' as well as our perceptual user study in the ``compositional prompts dataset'', this variant severely underperforms in terms of shape adherence compared to our method. According to our user study, it also underperforms with respect to textual cue matching, when this is evaluated perceptually.

\input{figures/quant_visual_eval_supp}

\section{Additional qualitative results}

We provide additional qualitative results for shape and prompt adherence in Figures \ref{fig:supp_geom_qual_eval} and \ref{fig:supp_prompt_qual_eval}, respectively. 

\section{Failure cases}

We observed that the failure cases for our model fall in two modes. First, it struggles with capturing details of challenging fine-grained geometry (Figure \ref{fig:failure_geometry}). In such cases, ShapeWords correctly captures coarse shape structure but struggles to reproduce fine geometric details. Our hypothesis is that the geometric precision of ShapeWords is likely to be bound by the image resolution of OpenCLIP model (ViT-H/14, 224px) which we used to train ShapeWords, and the ability of PointBert to capture such fine-scale geometric details. Training ShapeWords with variants of CLIP of higher resolution might yield better geometric precision -- we consider that this is a promising direction for future work.

Second, our model struggles to generalize to largely out-of-distribution text prompts. We illustrate this issue in Figure \ref{fig:faulure_prompts}. For example, the prompt 'an origami of a chair' requires both adjustment of texture and local geometry. Our model struggles to do both, especially for high values of guidance strength. We think this issue arises from a combination of two factors: a) our set of prompts is biased towards smoother appearances (e.g. `photo', `sketch', `illustration'), b) our supervisory images come from ControlNet that also tends to produce smooth surfaces following depth maps. However, results for intermediate guidance strength suggest that our model can still generalize to such prompts to some extent. We suspect that this issue could potentially be alleviated by using more diverse training data.

\input{figures/comparison_by_k}

\input{figures/comp_eval_qualitative_supp}

%% file: figures/eos_ablation.tex
\newcommand{\eosimgwidth}{0.09\textwidth}

\begin{figure}[t!]
\centering
\setlength{\tabcolsep}{1pt} % Adjust horizontal spacing
\renewcommand{\arraystretch}{1} % Adjust vertical spacing
\begin{tabular}{ccccc}
\small{Guidance} & \small{All tokens} & \small{Obj. only} & \small{EOS only} & \small{Obj.\&EOS} \\
\includegraphics[width=\eosimgwidth]{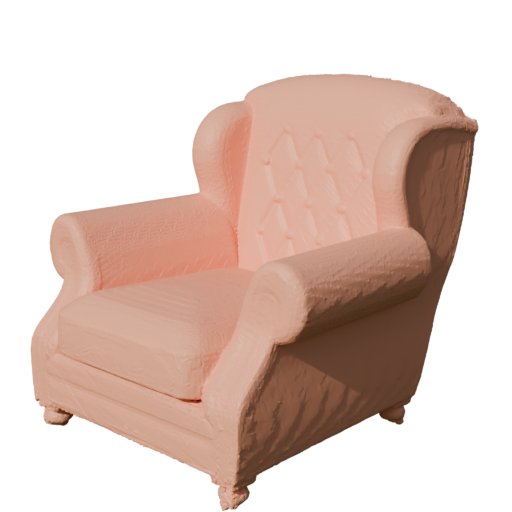} &
\includegraphics[width=\eosimgwidth]{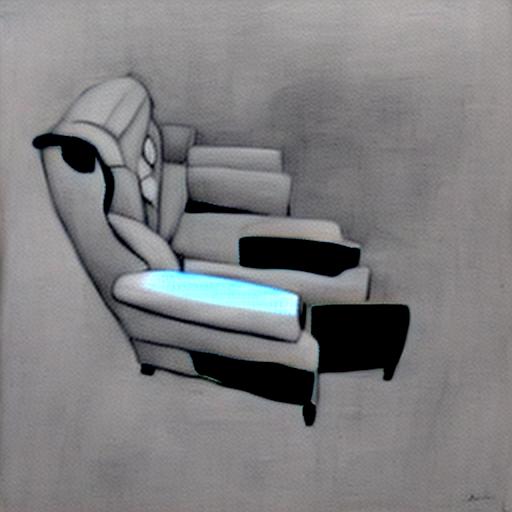} &
\includegraphics[width=\eosimgwidth]{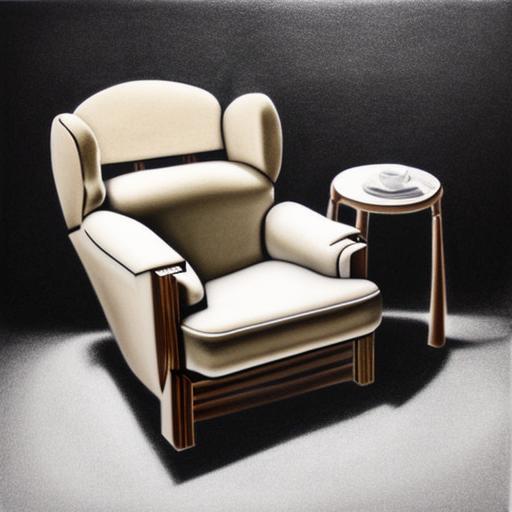} &
\includegraphics[width=\eosimgwidth]{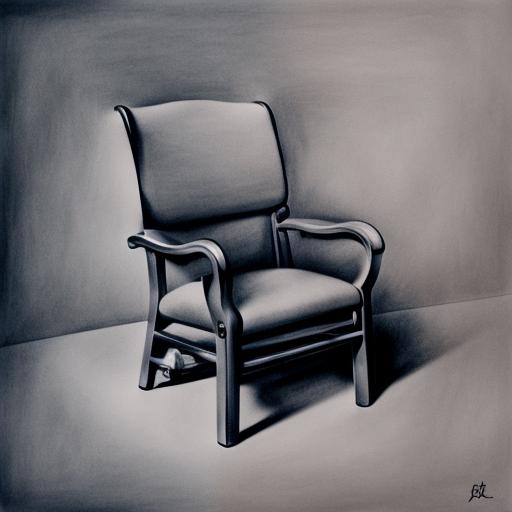}&
\includegraphics[width=\eosimgwidth]{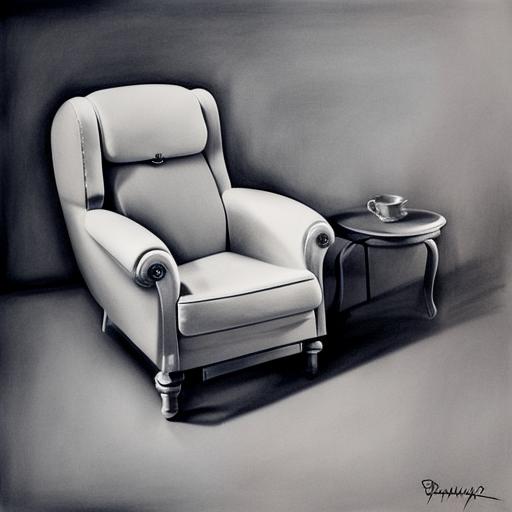} \\

%---------------------
\includegraphics[width=\eosimgwidth]{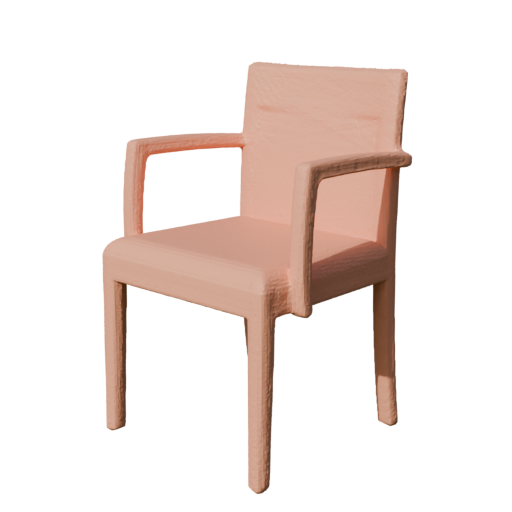} &
\includegraphics[width=\eosimgwidth]{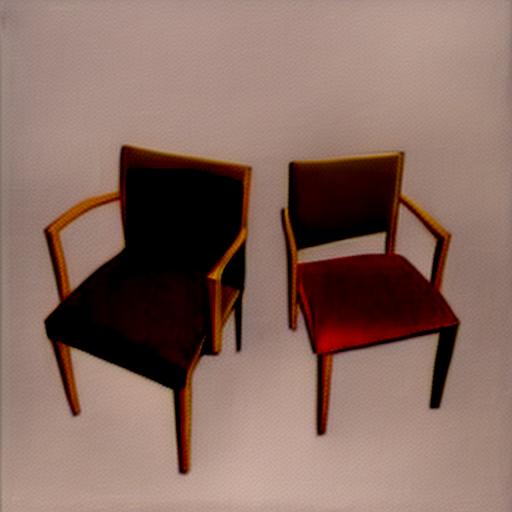} &
\includegraphics[width=\eosimgwidth]{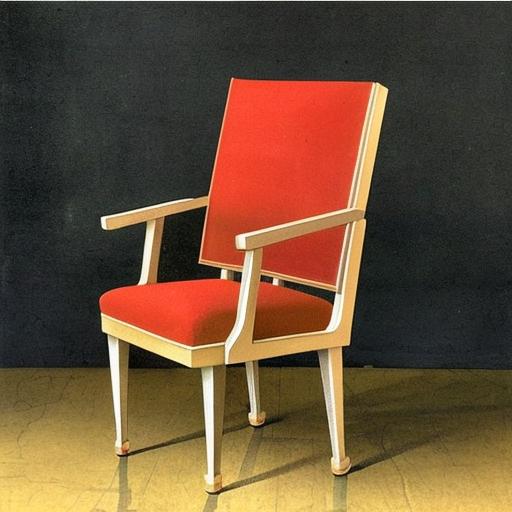} &
\includegraphics[width=\eosimgwidth]{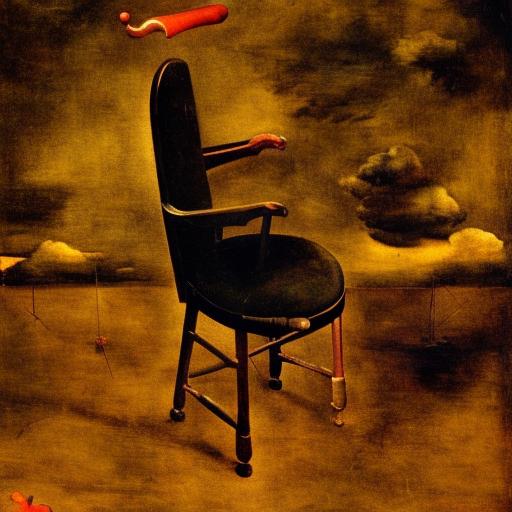} &
\includegraphics[width=\eosimgwidth]{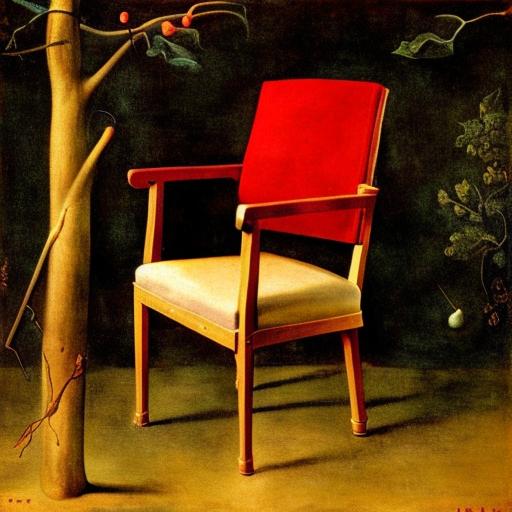} \\
%---------------------
%---------------------
\includegraphics[width=\eosimgwidth]{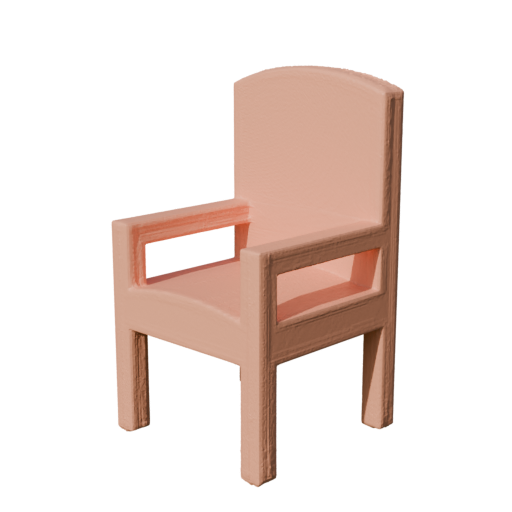} &
\includegraphics[width=\eosimgwidth]{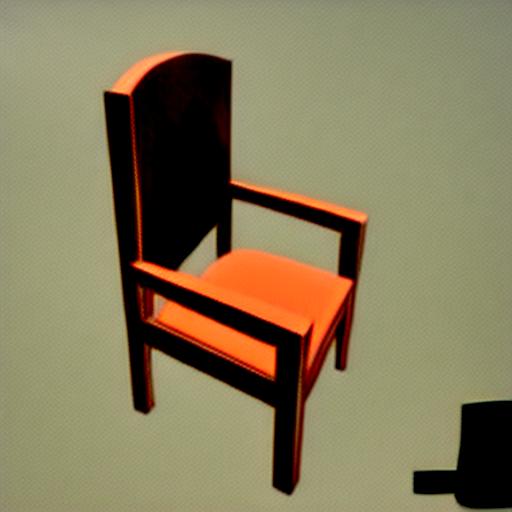} &
\includegraphics[width=\eosimgwidth]{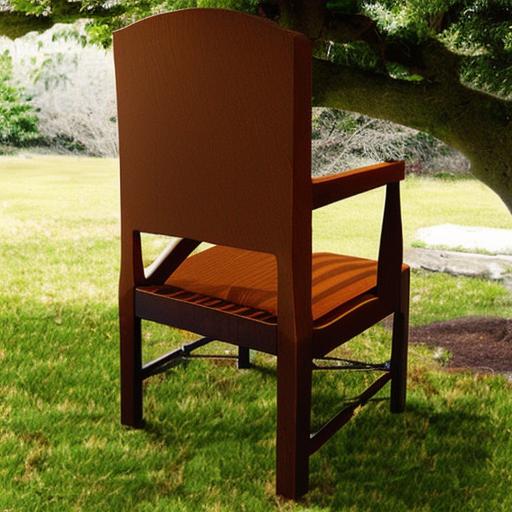} &
\includegraphics[width=\eosimgwidth]{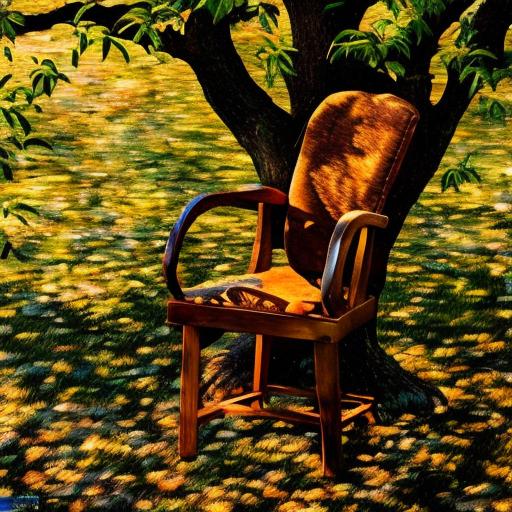} &
\includegraphics[width=\eosimgwidth]{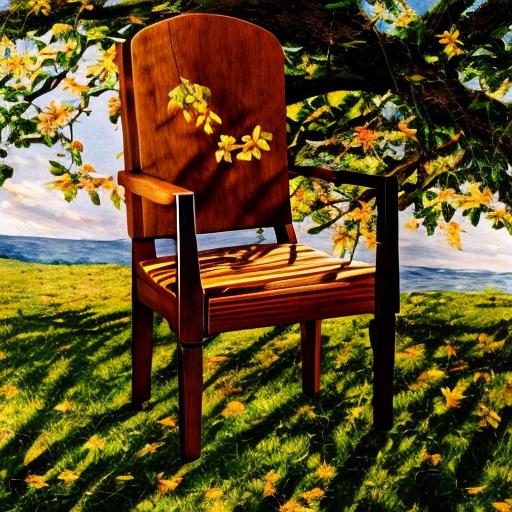} \\
%---------------------

\end{tabular}
\vspace{-3mm}
\caption{\textbf{Token replacement strategies.} We qualitatively compare the following strategies for guidance: adding the prompt delta $\deltaT$ to all prompt embeddings; adding it to only the object word embedding; adding it only to EOS token embedding; adding it to both EOS and object token embeddings (as done in the main paper). Prompts are: `a charcoal drawing of \textbf{chair}' (top row), `Hieronymus Bosch's painting of a \textbf{chair}' (middle row), `a \textbf{chair} under a tree' (bottom row). Target shapes are shown on the left. Compared to modifying the object \& EOS tokens, the ``all tokens'' strategy produces over-smoothed images; the ``object only token'' strategy struggles to incorporate stylistic cues from the text into geometry; and the ``EOS token'' strategy struggles with preserving the target shape geometry. }
\vspace{-3mm}
\label{fig:supp_role_of_eos}
\end{figure}

%% file: figures/failure_cases_geometry.tex
\newlength{\failgemimagewidth}
\setlength{\failgemimagewidth}{0.09\textwidth} % Adjust this value as needed

% Define a new column type C based on \imagewidth
\newcolumntype{C}{>{\centering\arraybackslash}m{\failgemimagewidth}}

% Set the color of table lines to grey
%\arrayrulecolor{gray}

\begin{figure}[t!]
\centering
\setlength{\tabcolsep}{1pt} % Reduce horizontal padding between columns
\begin{tabular}{C*{5}{C}}
%\toprule
% First Row: Empty cell and Shape labels with images
%& % Empty cell (upper-left corner)
%\parbox[c]{\failgemimagewidth}{\centering \textbf{Shape 1}\\ \includegraphics[width=\failgemimagewidth]{example-image-a}} &
\parbox[c]{\failgemimagewidth}{\centering \includegraphics[width=\failgemimagewidth]{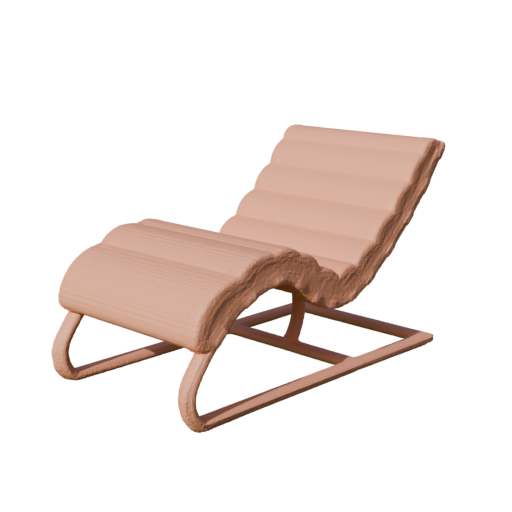}} &
\parbox[c]{\failgemimagewidth}{\centering \includegraphics[width=\failgemimagewidth]{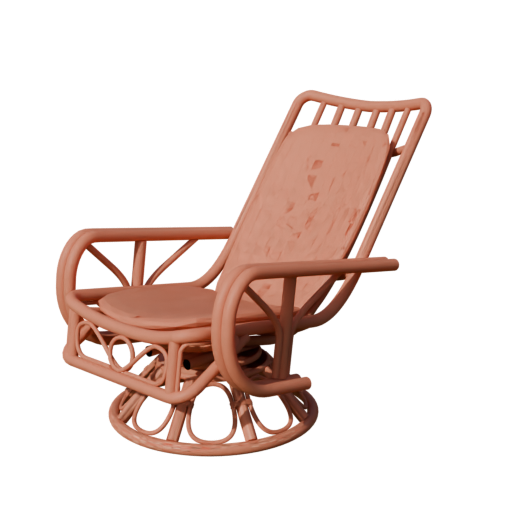}} &
\parbox[c]{\failgemimagewidth}{\centering \includegraphics[width=\failgemimagewidth]{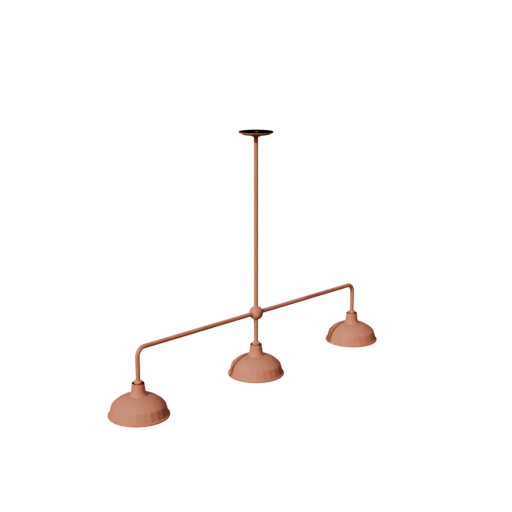}} &
\parbox[c]{\failgemimagewidth}{\centering \includegraphics[width=\failgemimagewidth]{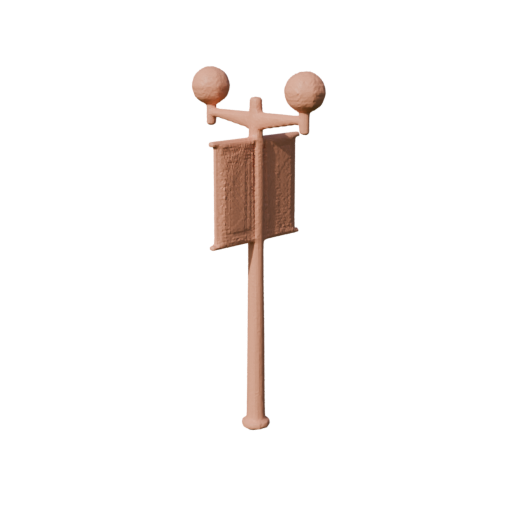}} &
\parbox[c]{\failgemimagewidth}{\centering \includegraphics[width=\failgemimagewidth]{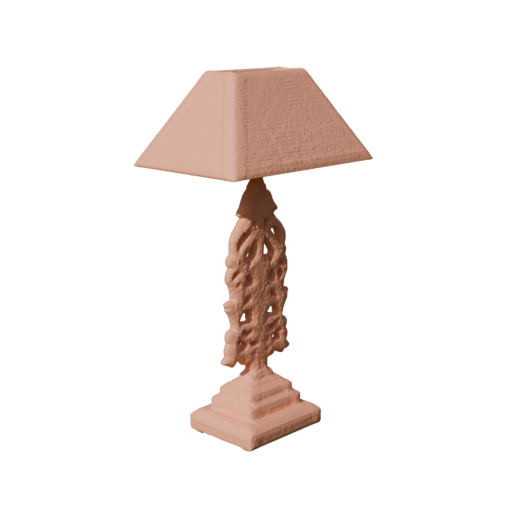}} \\
%\midrule
% Subsequent Rows: Prompts and generated images
%\footnotesize{Photo on a beach}  &
%\includegraphics[width=\failgemimagewidth]{figures/img/prompt_shape_grid/beach/03001627_1a38407b3036795d19fb4103277a6b93/seed_0_str_0.95_res_4.jpg} & % Shape 1 image
\includegraphics[width=\failgemimagewidth]{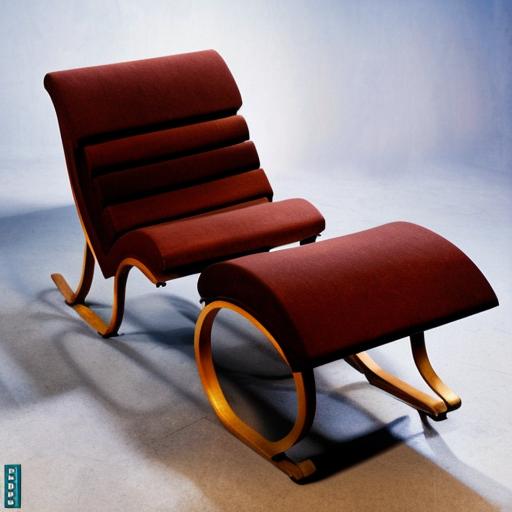} & % Shape 2 image
\includegraphics[width=\failgemimagewidth]{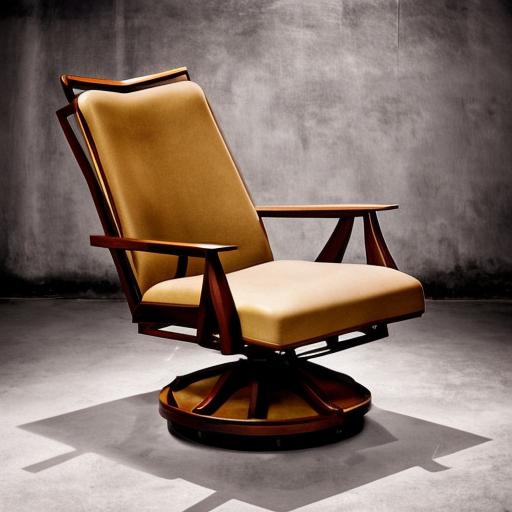} &% Shape 3 image
\includegraphics[width=\failgemimagewidth]{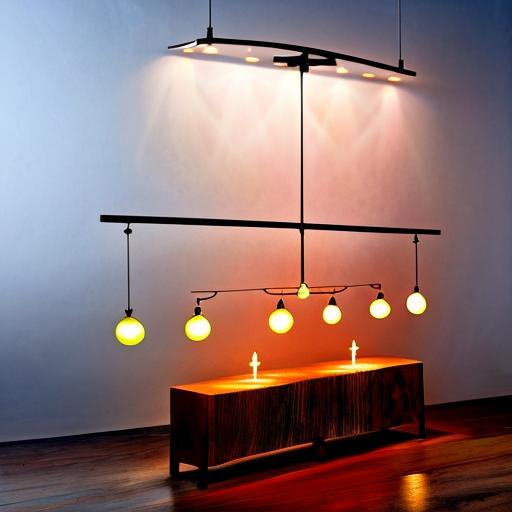} & % Shape 4 image
\includegraphics[width=\failgemimagewidth]{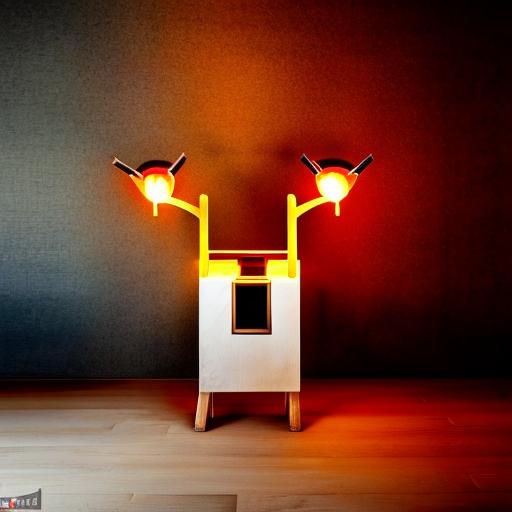} & % Shape 5 image
\includegraphics[width=\failgemimagewidth]{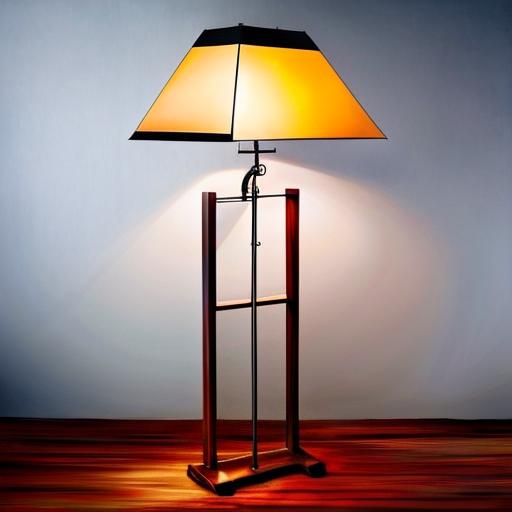}
\\ % Shape 6 image

%\bottomrule
\end{tabular}
\vspace{-3mm}
\caption{\textbf{Failure cases -- shape adherence.} Our model struggles to generalize to shapes with complex fine-grained geometries (e.g. a lot of thin parts or lot of holes). 
Prompts for the shapes are: `a \textbf{chair}' (first two shapes); `a \textbf{lamp}' (last three shapes). Target shapes are shown on the top.      
}
\label{fig:failure_geometry}
\end{figure}

%% file: figures/failure_prompts.tex
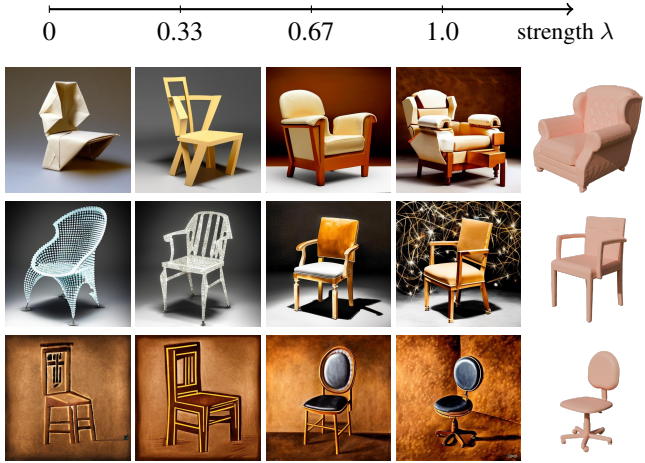
\begin{figure}[t!]
\centering
\setlength{\tabcolsep}{1pt} % Adjust horizontal spacing between columns

% Define column widths directly
\newlength{\failpromptcontrolcolwidth}
\setlength{\failpromptcontrolcolwidth}{0.095\textwidth} % Width of each control column
\newlength{\promptfailguidancecolwidth}
\setlength{\promptfailguidancecolwidth}{0.095\textwidth} % Width of guidance column

\begin{tabular}{%
    *{4}{>{\centering\arraybackslash}m{\failpromptcontrolcolwidth}}%
    >{\centering\arraybackslash}m{\promptfailguidancecolwidth}%
}
% Arrow spanning control columns
\multicolumn{5}{c}{
\begin{tikzpicture}
\begin{scope}[x=1.05\failpromptcontrolcolwidth, y=0.5cm]
\draw[->, thick] (0,0) -- (4,0);
\foreach \x/\label in {0/0, 1/0.33, 2/0.67, 3/1.0}
{
    \draw (\x,0.1) -- (\x,-0.1);
    \node[below] at (\x,-0.1) {\label};
}
\node[below right] at (3.5,-0.1) {\small{strength $\lambda$}};
\end{scope}
\end{tikzpicture}
} \\[2pt]

% First row of images
\includegraphics[width=\linewidth]{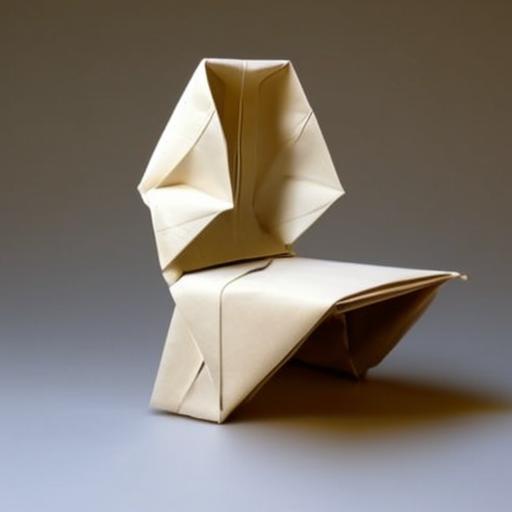} &
\includegraphics[width=\linewidth]{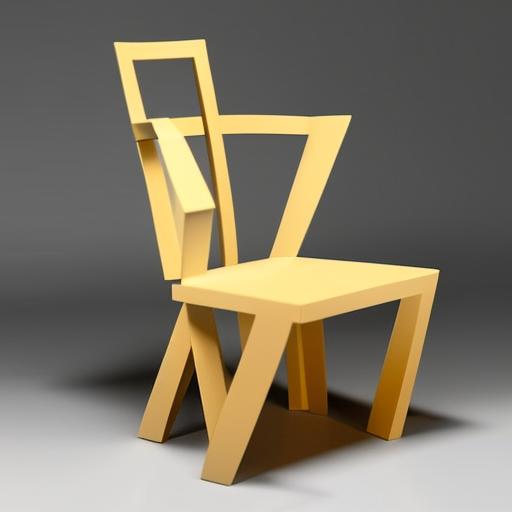} &
\includegraphics[width=\linewidth]{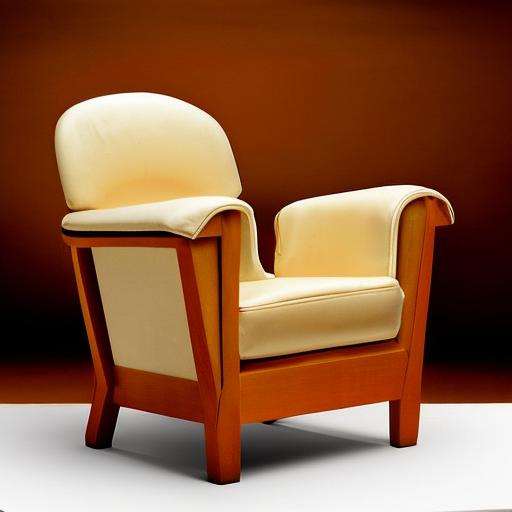} &
\includegraphics[width=\linewidth]{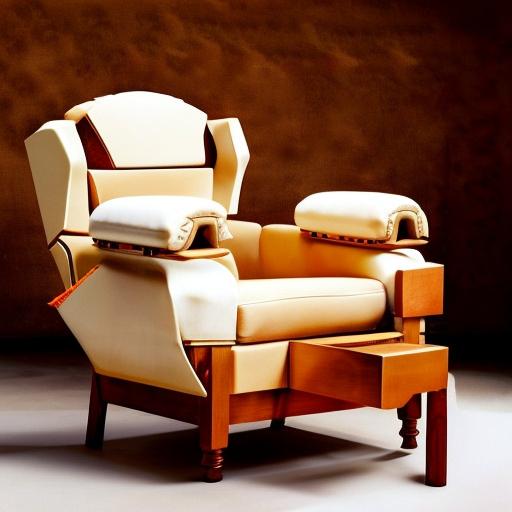} &
\includegraphics[width=\linewidth]{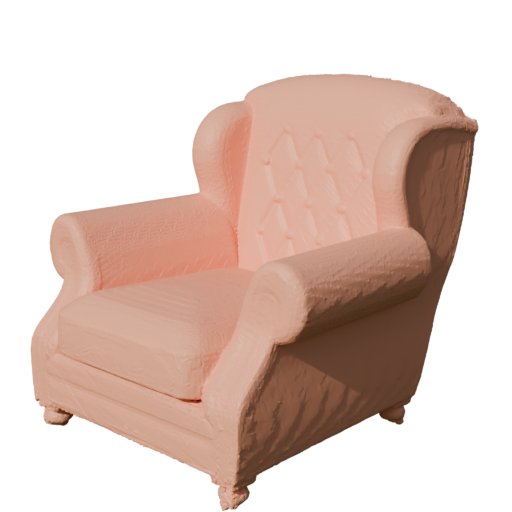} \\[2pt]

% Second row of images
\includegraphics[width=\linewidth]{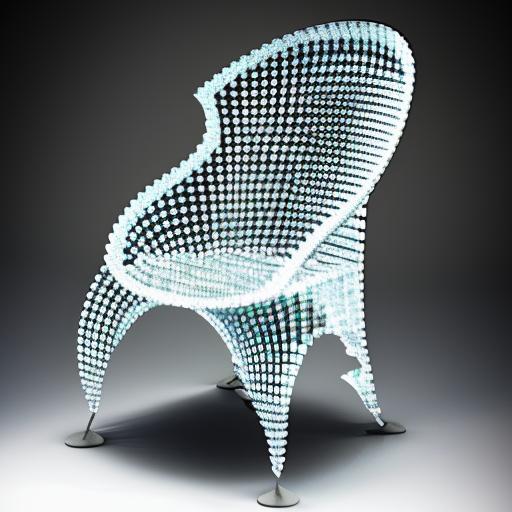} &
\includegraphics[width=\linewidth]{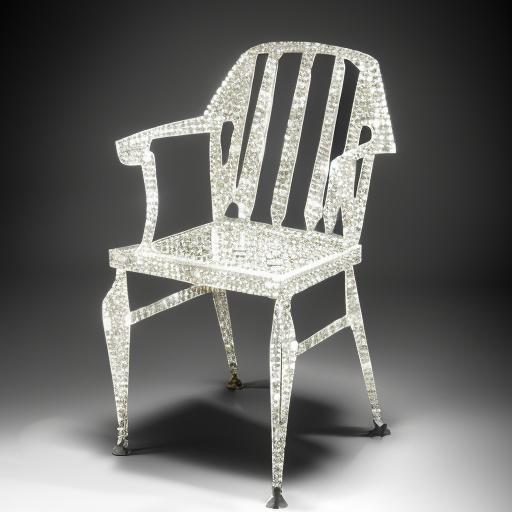} &
\includegraphics[width=\linewidth]{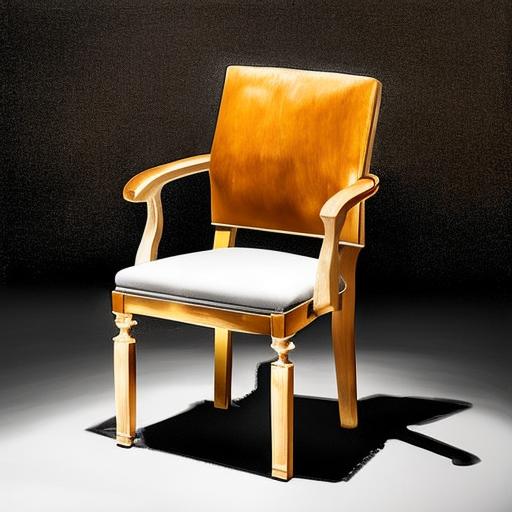} &
\includegraphics[width=\linewidth]{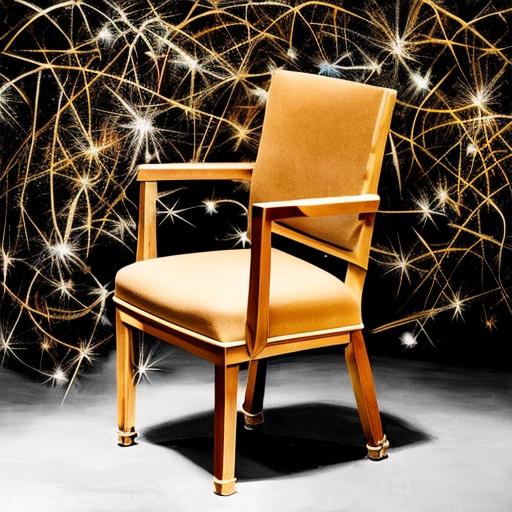} &
\includegraphics[width=\linewidth]{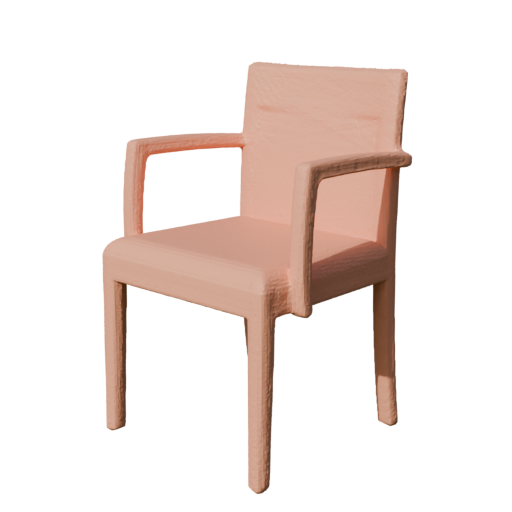} \\[2pt]

% Third row of images
\includegraphics[width=\linewidth]{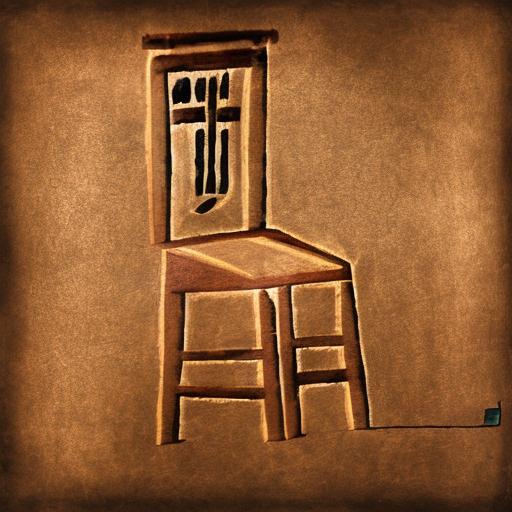} &
\includegraphics[width=\linewidth]{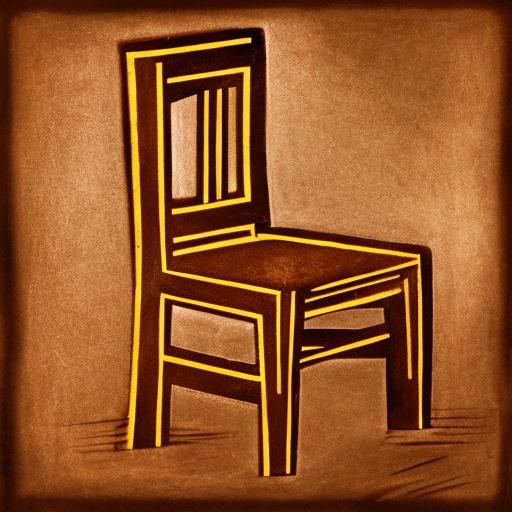} &
\includegraphics[width=\linewidth]{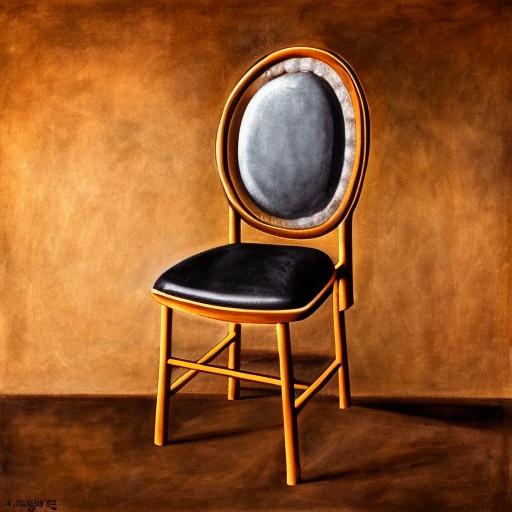} &
\includegraphics[width=\linewidth]{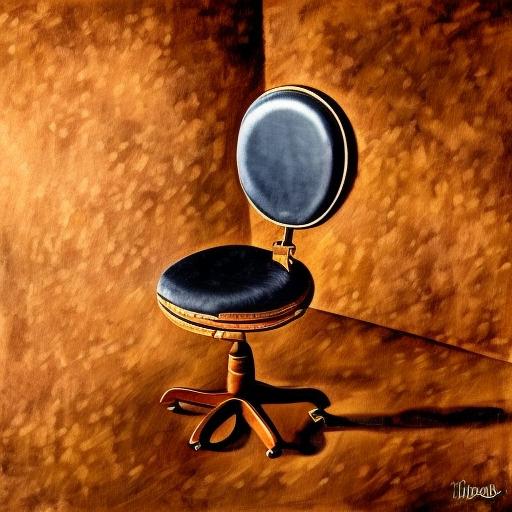} &
\includegraphics[width=\linewidth]{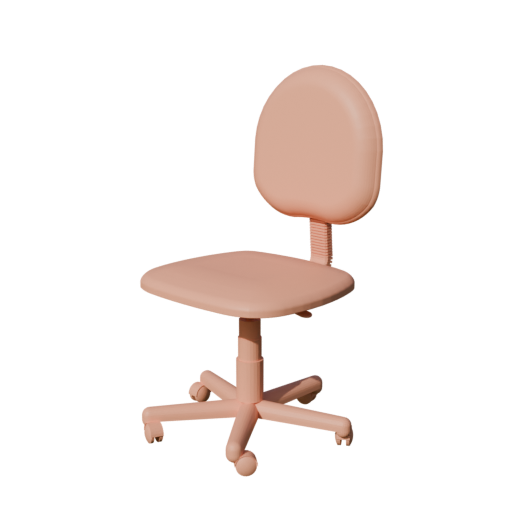} \\

% Add more rows as needed
\end{tabular}
\vspace{-3mm}
\caption{\textbf{Failure cases -- prompt adherence.} Our model struggles to generalize to out-of-distribution prompts that require local adjustments of surface geometry. Prompts are: 'an origami of a chair' (top); 'a diamond sculpture of a chair' (middle); 'a hieroglyph of a chair' (bottom). Target shapes are shown on the right. }
 \vspace{-2mm}
\label{fig:faulure_prompts}
\end{figure}

%% file: figures/quant_visual_eval_supp.tex
\begin{table}[th!]
\centering
\begin{tabular}{l|cc}
\toprule
\textbf{Model} &  \textbf{Aes. $\uparrow$} & \textbf{CLIP $\uparrow$} \\
\midrule
ControlNet &  5.24& 26.9\\
\midrule
CNet-Stop@20     &  5.15& 30.3\\
CNet-Stop@30     &  5.18& \textbf{31.5}\\
CNet-Stop@40     &  5.15& 30.3\\
CNet-Stop@60     & 5.20& 28.3\\
CNet-Stop@80     & 5.17& 27.5\\
ShapeWords           &  \textbf{5.45}& \textbf{31.5}\\
\bottomrule
\end{tabular}
\vspace{-3mm}
\caption{\textbf{Evaluation results on compositional prompts.} Taking into account both the Aesthetics score and the CLIP score (scaled by $100$), our method outperforms  ControlNet variants in the challenging compositional setting. Even if the CNet-Stop@30 variant matches the CLIP score of our method, it still severely underperforms in terms of shape adherence according to our user study and the rest of our experiments.
}
\label{table:supp_quant_eval}
\end{table}

%% file: figures/comparison_by_k.tex
\newcommand{\ablimgwidth}{0.09\textwidth}
\begin{figure*}[htbp]
\centering
\setlength{\tabcolsep}{1pt} % Adjust horizontal spacing
\renewcommand{\arraystretch}{1} % Adjust vertical spacing
\begin{tabular}{c|ccc|ccc|ccc}
& \multicolumn{2}{c}{\scriptsize{CNet-Stop@20}} & & \multicolumn{2}{c}{\scriptsize{CNet-Stop@40}} & & \multicolumn{2}{c}{\scriptsize{CNet-Stop@80}} & \\
\scriptsize{Depth} & \scriptsize{Category} & \scriptsize{Subcategory} & \small{\scriptsize{ShapeWords@20}} & \scriptsize{Category} & \scriptsize{Subcategory} & \small{\scriptsize{ShapeWords@40}} & \scriptsize{Category} & \scriptsize{Subcategory} & \small{\scriptsize{ShapeWords@80}} \\
\includegraphics[width=\ablimgwidth]{figures/img/comparison_by_k/03001627_cea21726757b5253c3648f83bb1262ce/060/input_depth.png} &
\includegraphics[width=\ablimgwidth]{figures/img/comparison_by_k/03001627_cea21726757b5253c3648f83bb1262ce/060/cat_k_0.2.jpg} &
\includegraphics[width=\ablimgwidth]{figures/img/comparison_by_k/03001627_cea21726757b5253c3648f83bb1262ce/060/subcat_k_0.2.jpg} &
\includegraphics[width=\ablimgwidth]{figures/img/comparison_by_k/03001627_cea21726757b5253c3648f83bb1262ce/060/ours_k_0.2.jpg} &
\includegraphics[width=\ablimgwidth]{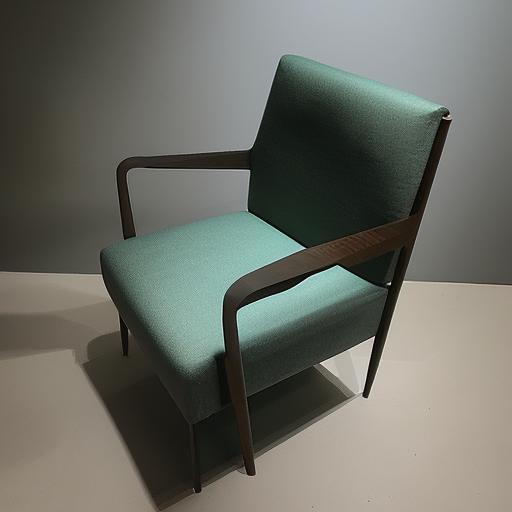} &
\includegraphics[width=\ablimgwidth]{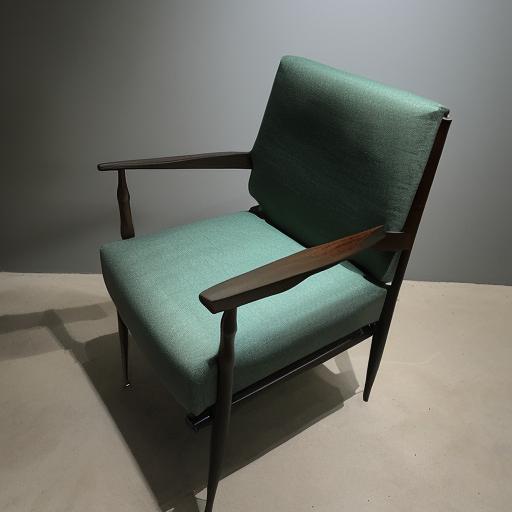} &
\includegraphics[width=\ablimgwidth]{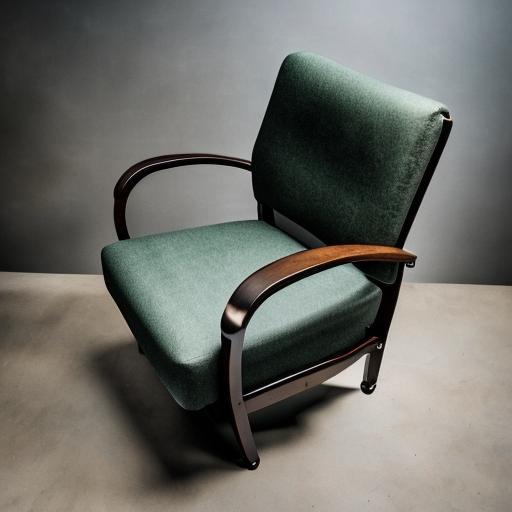} &
\includegraphics[width=\ablimgwidth]{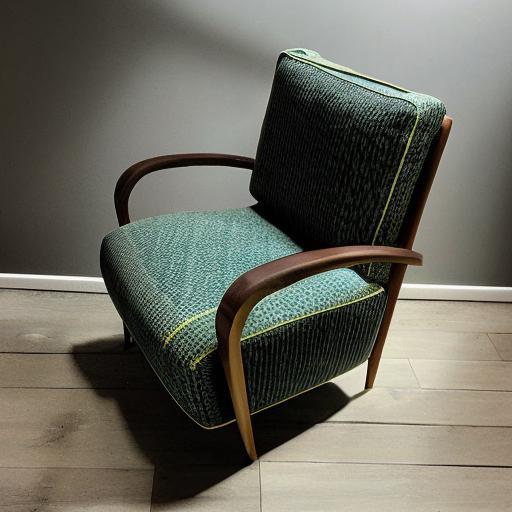} &
\includegraphics[width=\ablimgwidth]{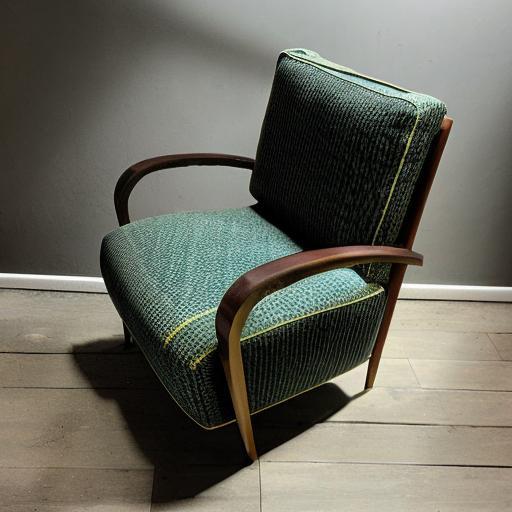} &
\includegraphics[width=\ablimgwidth]{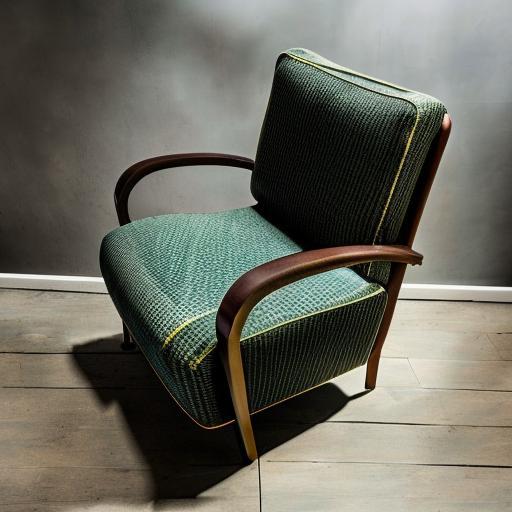} \\
%---------------------
\includegraphics[width=\ablimgwidth]{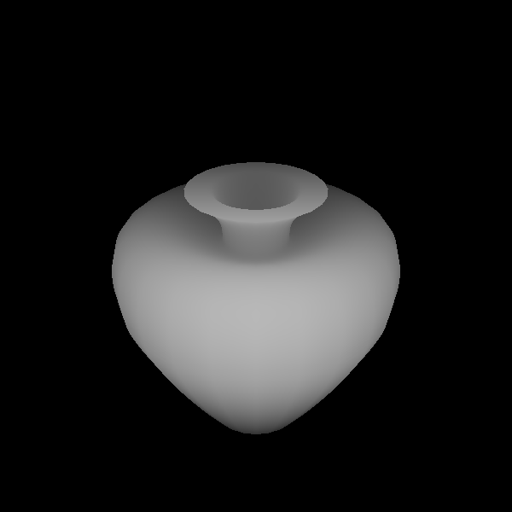} &
\includegraphics[width=\ablimgwidth]{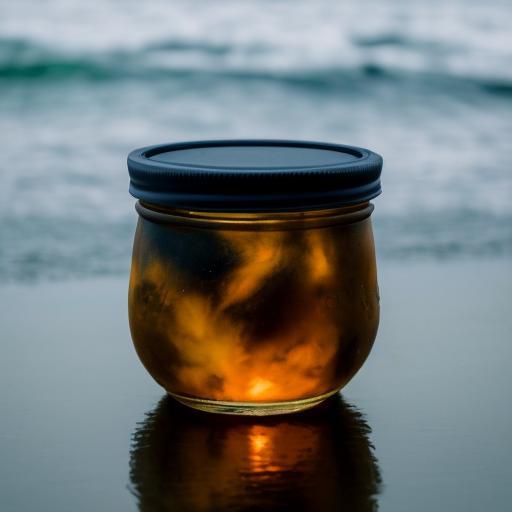} &
\includegraphics[width=\ablimgwidth]{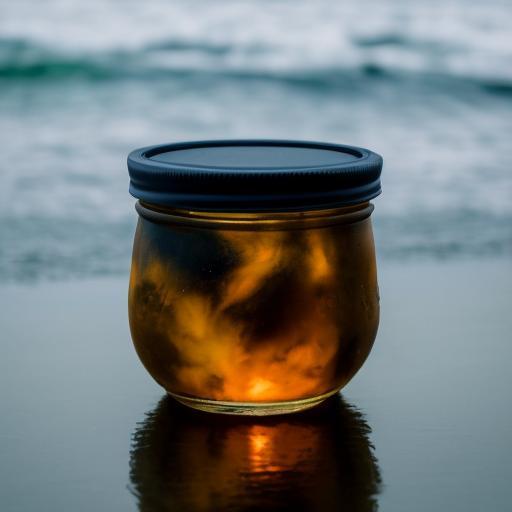} &
\includegraphics[width=\ablimgwidth]{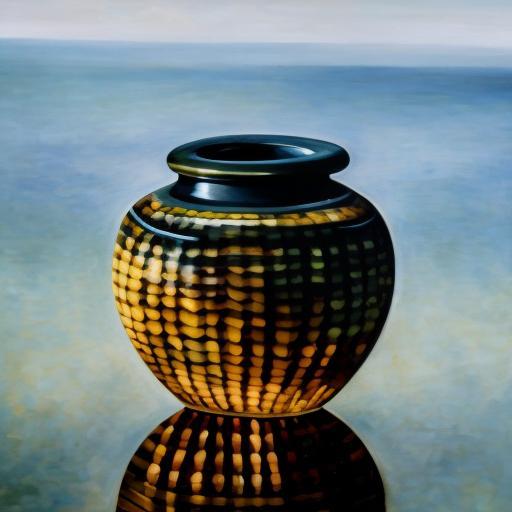} &
\includegraphics[width=\ablimgwidth]{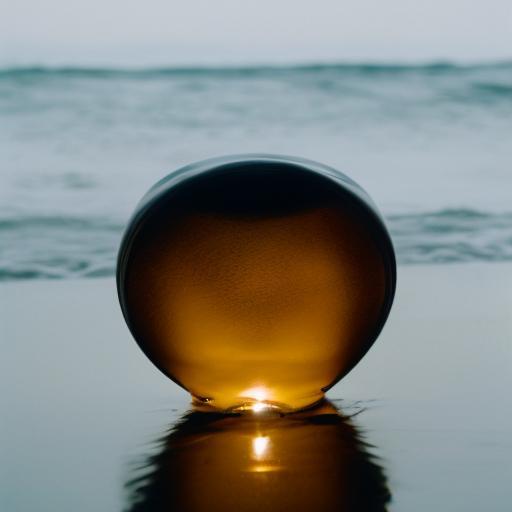} &
\includegraphics[width=\ablimgwidth]{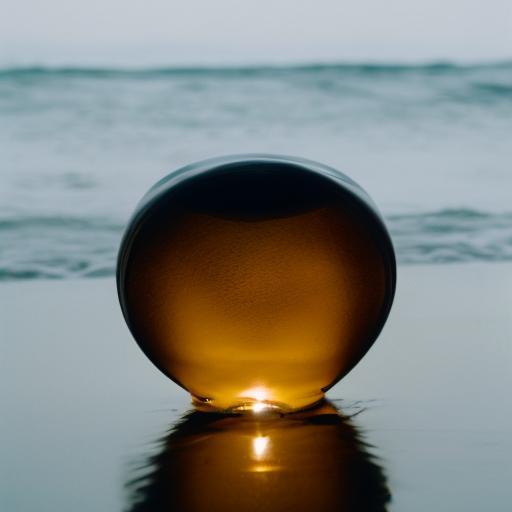} &
\includegraphics[width=\ablimgwidth]{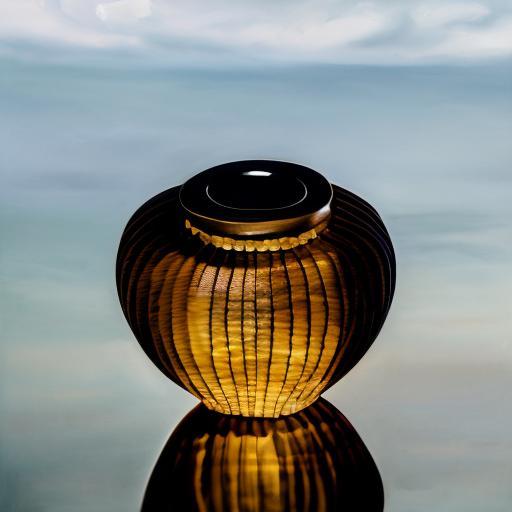} &
\includegraphics[width=\ablimgwidth]{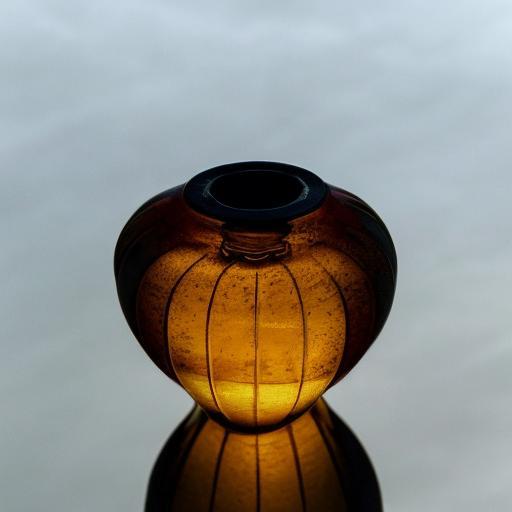} &
\includegraphics[width=\ablimgwidth]{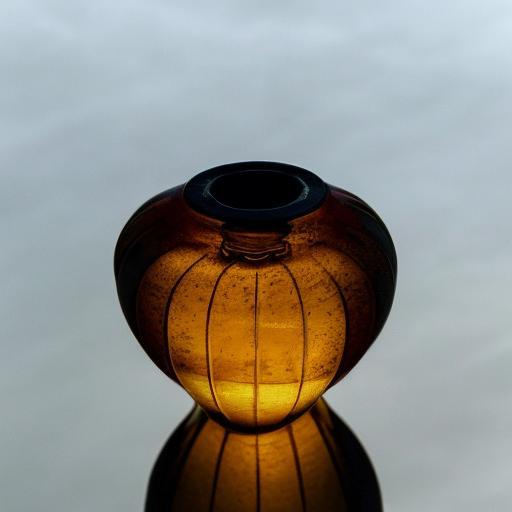} &
\includegraphics[width=\ablimgwidth]{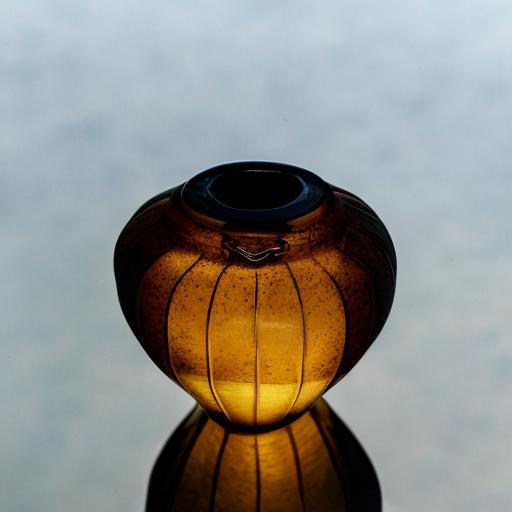} \\
%---------------------
\includegraphics[width=\ablimgwidth]{figures/img/comparison_by_k/04256520_9e18b198e803af3513f0322d0d7e53c7/060/input_depth.png} &
\includegraphics[width=\ablimgwidth]{figures/img/comparison_by_k/04256520_9e18b198e803af3513f0322d0d7e53c7/060/cat_k_0.2.jpg} &
\includegraphics[width=\ablimgwidth]{figures/img/comparison_by_k/04256520_9e18b198e803af3513f0322d0d7e53c7/060/subcat_k_0.2.jpg} &
\includegraphics[width=\ablimgwidth]{figures/img/comparison_by_k/04256520_9e18b198e803af3513f0322d0d7e53c7/060/ours_k_0.2.jpg} &
\includegraphics[width=\ablimgwidth]{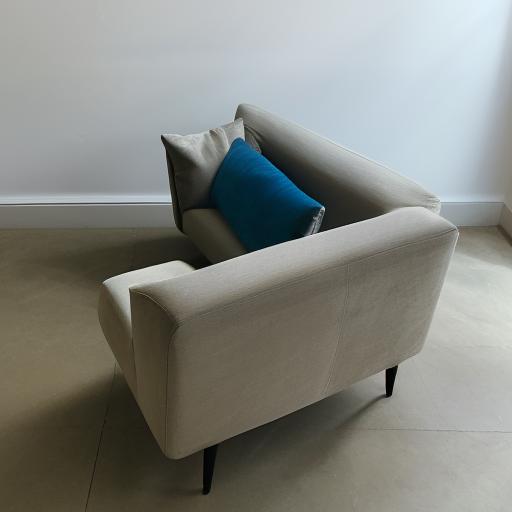} &
\includegraphics[width=\ablimgwidth]{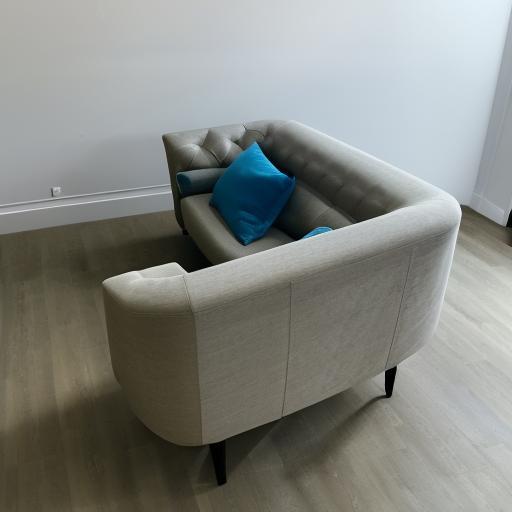} &
\includegraphics[width=\ablimgwidth]{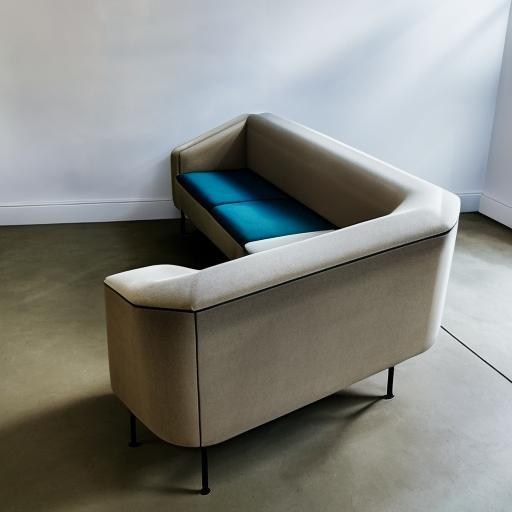} &
\includegraphics[width=\ablimgwidth]{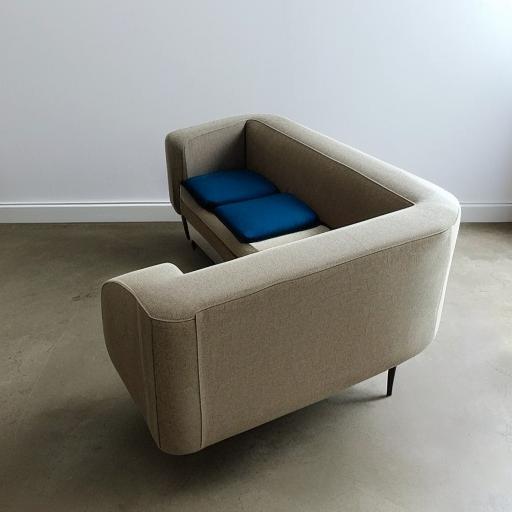} &
\includegraphics[width=\ablimgwidth]{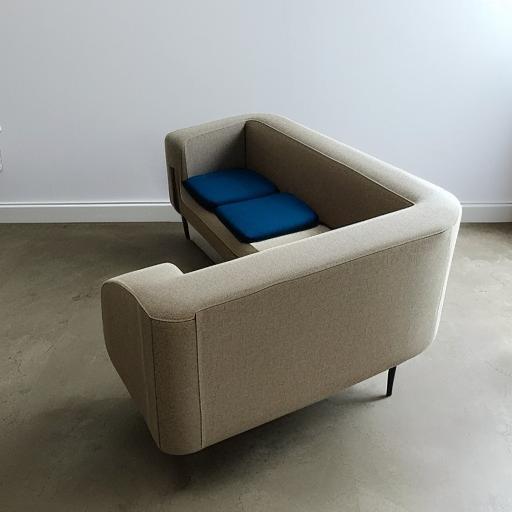} &
\includegraphics[width=\ablimgwidth]{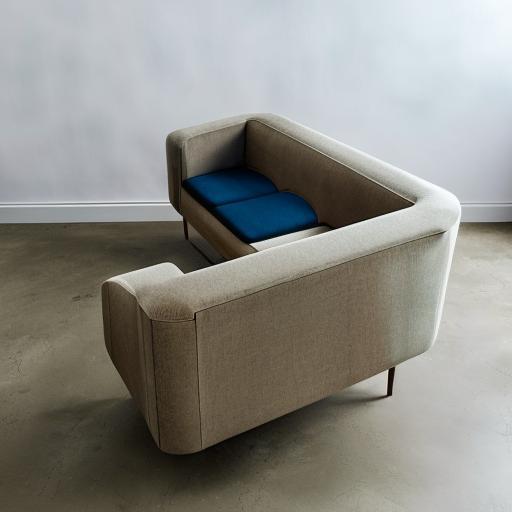} \\
%---------------------
\includegraphics[width=\ablimgwidth]{figures/img/comparison_by_k/02958343_8606a6353a2c0f7a453660f3d68cae6e/060/input_depth.png} &
\includegraphics[width=\ablimgwidth]{figures/img/comparison_by_k/02958343_8606a6353a2c0f7a453660f3d68cae6e/060/cat_k_0.2.jpg} &
\includegraphics[width=\ablimgwidth]{figures/img/comparison_by_k/02958343_8606a6353a2c0f7a453660f3d68cae6e/060/subcat_k_0.2.jpg} &
\includegraphics[width=\ablimgwidth]{figures/img/comparison_by_k/02958343_8606a6353a2c0f7a453660f3d68cae6e/060/ours_k_0.2.jpg} &
\includegraphics[width=\ablimgwidth]{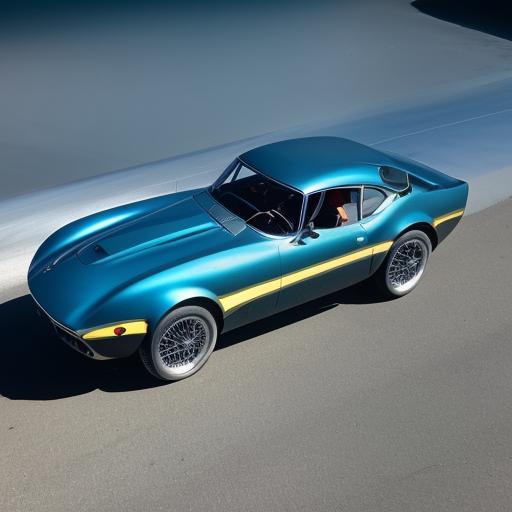} &
\includegraphics[width=\ablimgwidth]{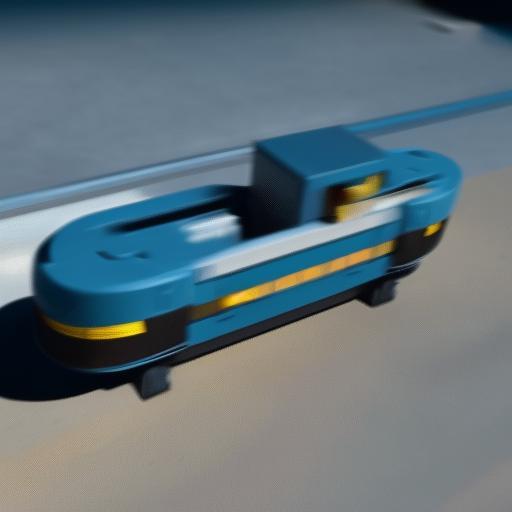} &
\includegraphics[width=\ablimgwidth]{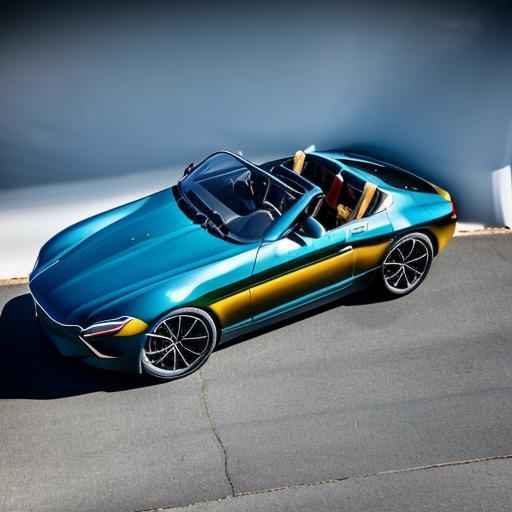} &
\includegraphics[width=\ablimgwidth]{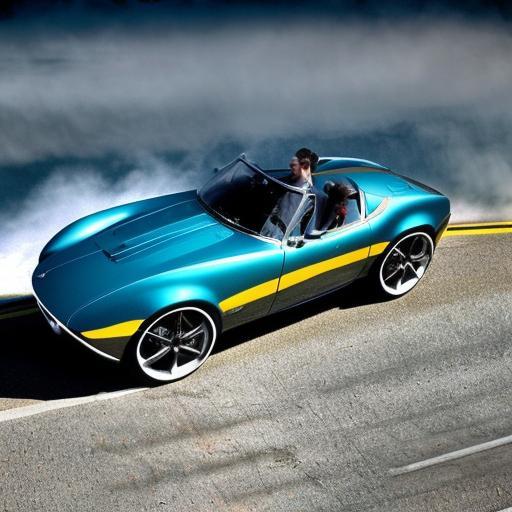} &
\includegraphics[width=\ablimgwidth]{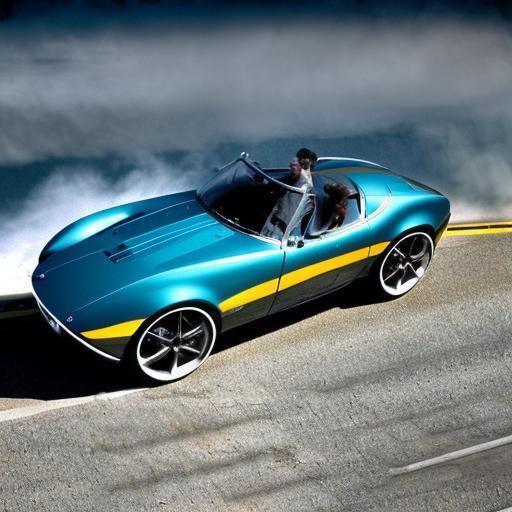} &
\includegraphics[width=\ablimgwidth]{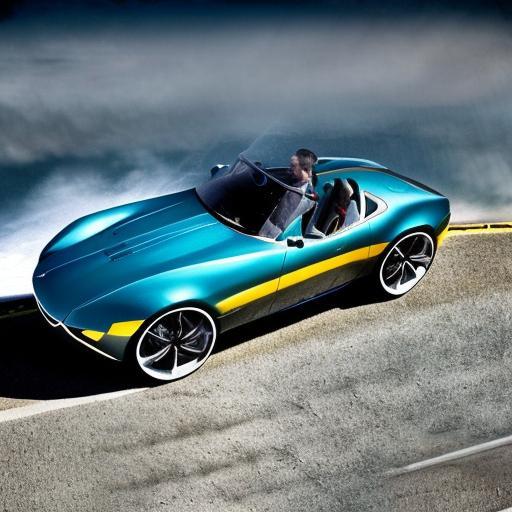} \\
%---------------------
\includegraphics[width=\ablimgwidth]{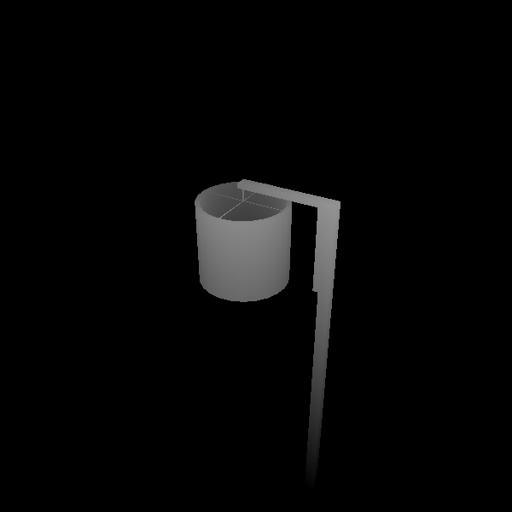} &
\includegraphics[width=\ablimgwidth]{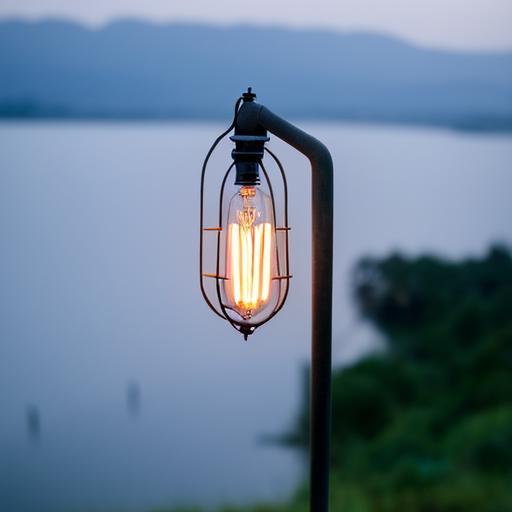} &
\includegraphics[width=\ablimgwidth]{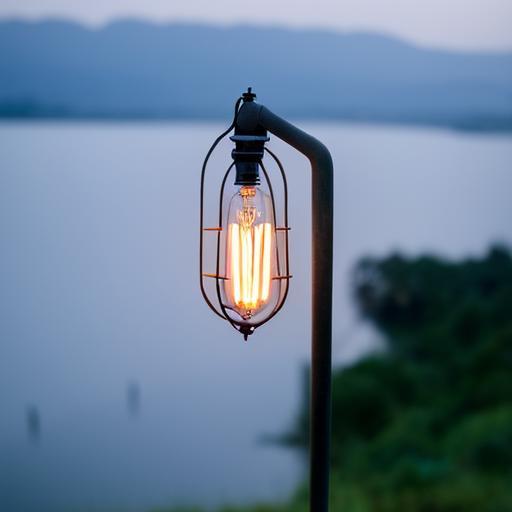} &
\includegraphics[width=\ablimgwidth]{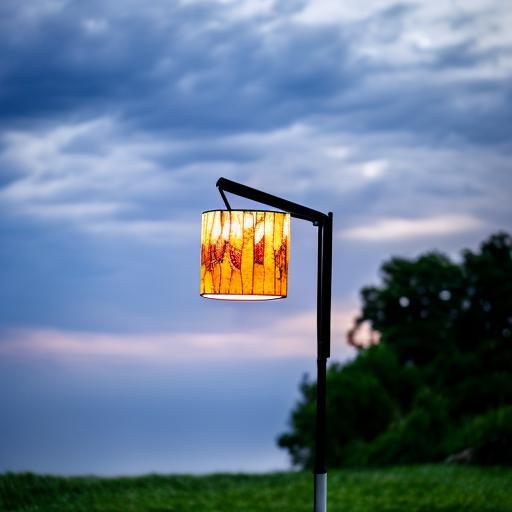} &
\includegraphics[width=\ablimgwidth]{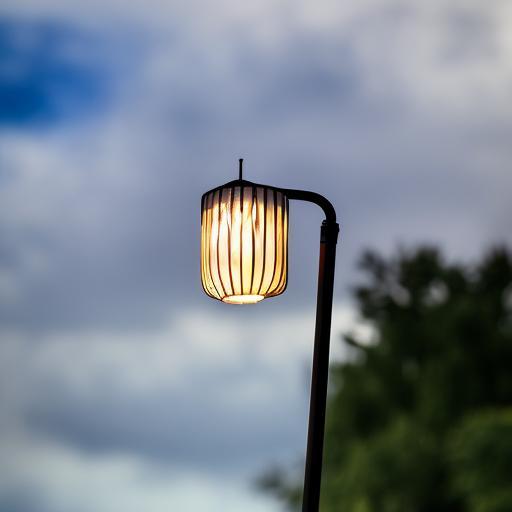} &
\includegraphics[width=\ablimgwidth]{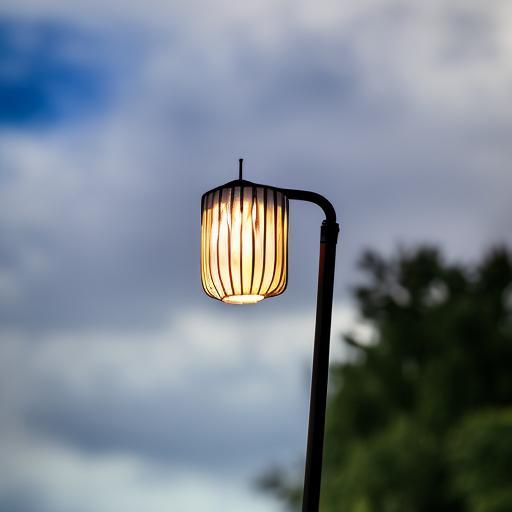} &
\includegraphics[width=\ablimgwidth]{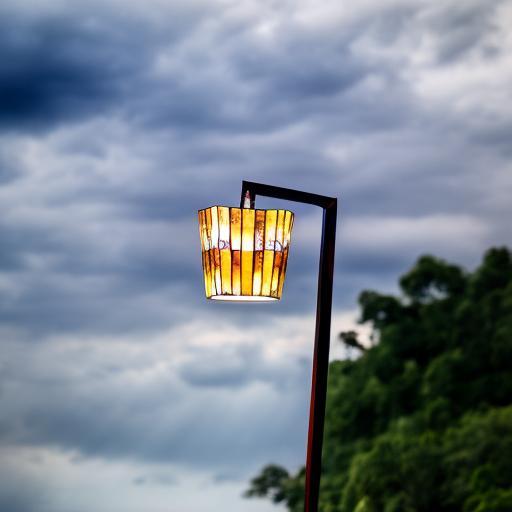} &
\includegraphics[width=\ablimgwidth]{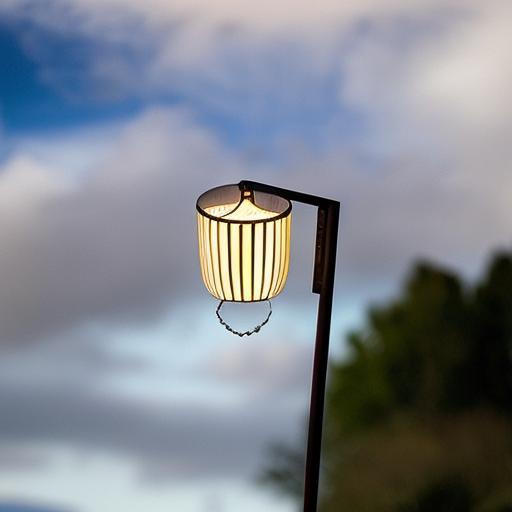} &
\includegraphics[width=\ablimgwidth]{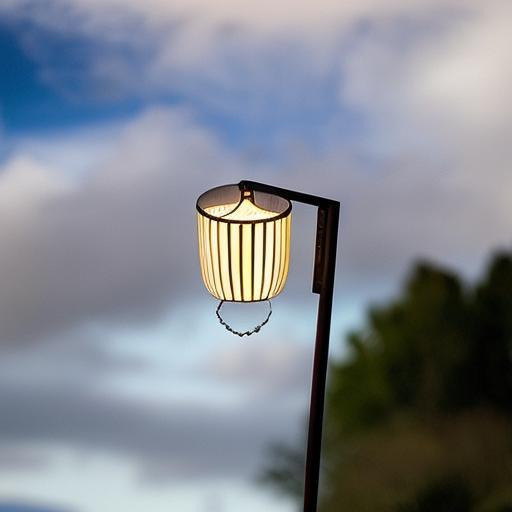} &
\includegraphics[width=\ablimgwidth]{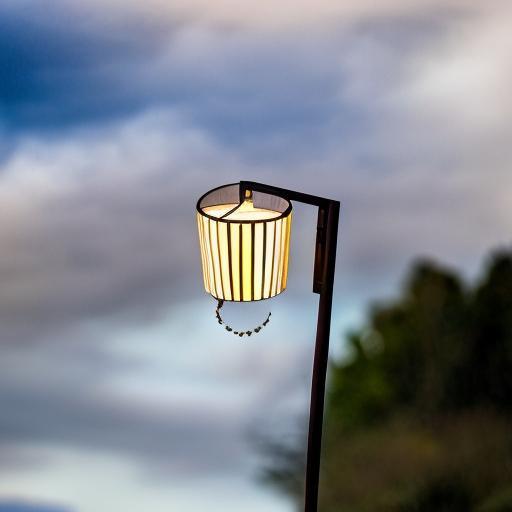} \\

%---------------------
\includegraphics[width=\ablimgwidth]{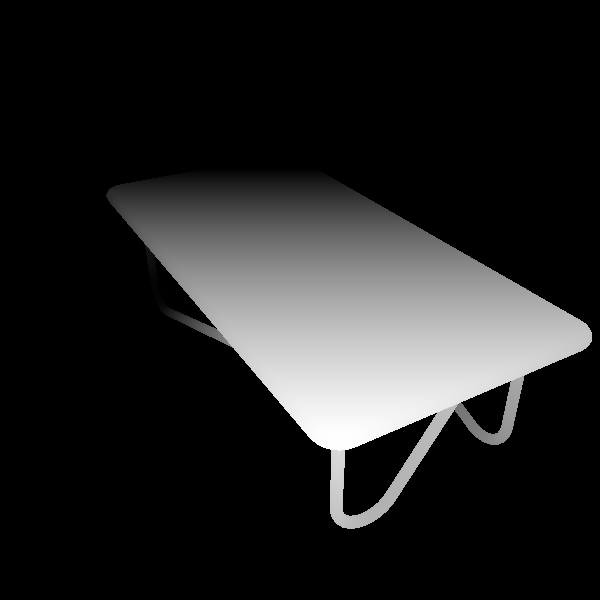} &
\includegraphics[width=\ablimgwidth]{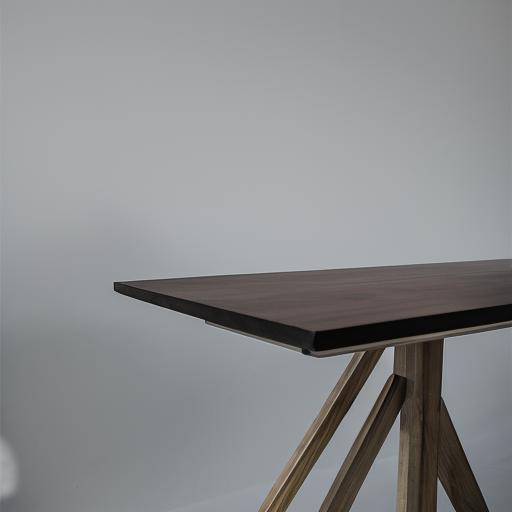} &
\includegraphics[width=\ablimgwidth]{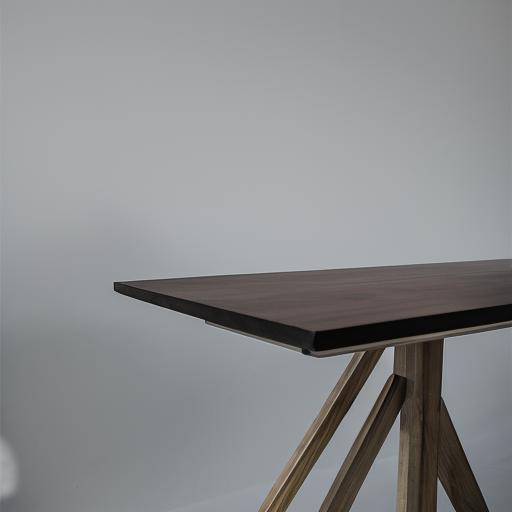} &
\includegraphics[width=\ablimgwidth]{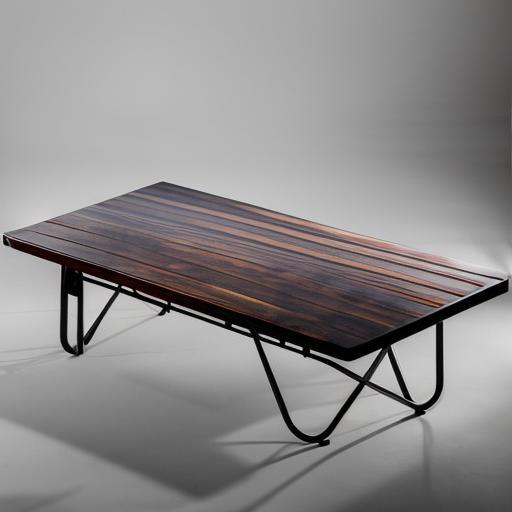} &
\includegraphics[width=\ablimgwidth]{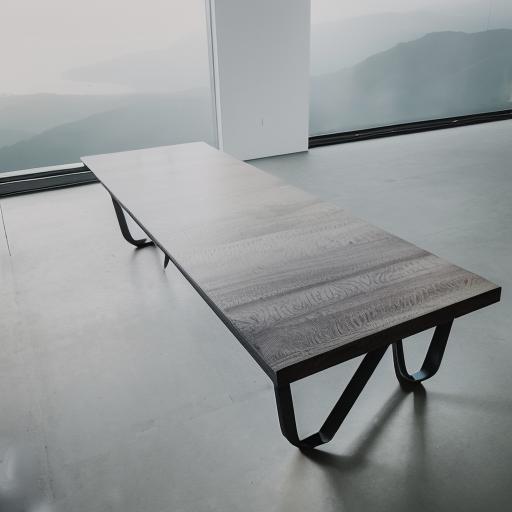} &
\includegraphics[width=\ablimgwidth]{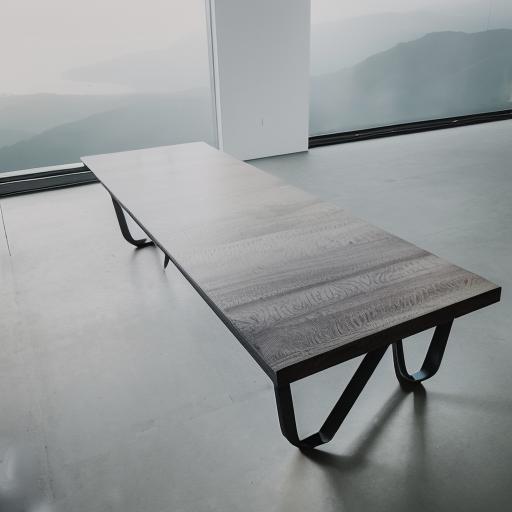} &
\includegraphics[width=\ablimgwidth]{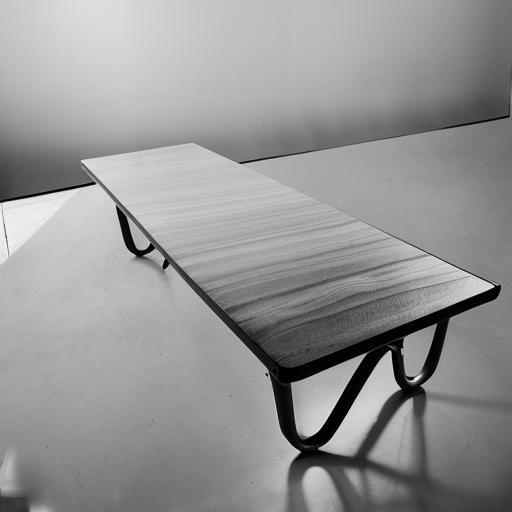} &
\includegraphics[width=\ablimgwidth]{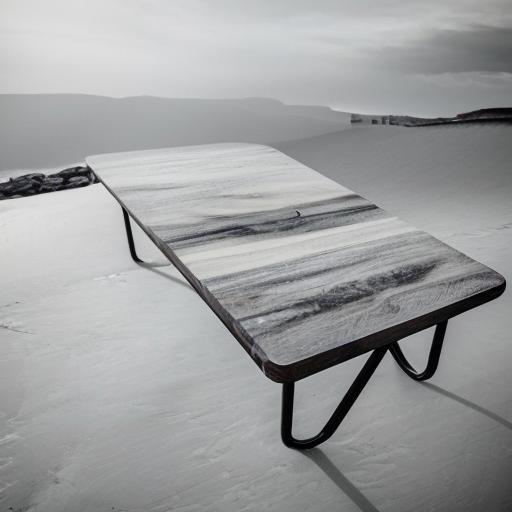} &
\includegraphics[width=\ablimgwidth]{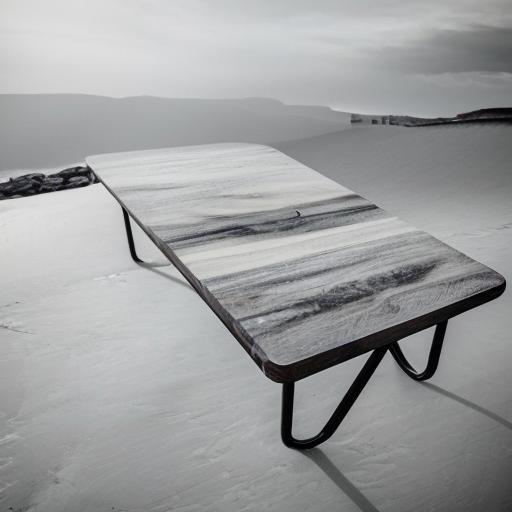} &
\includegraphics[width=\ablimgwidth]{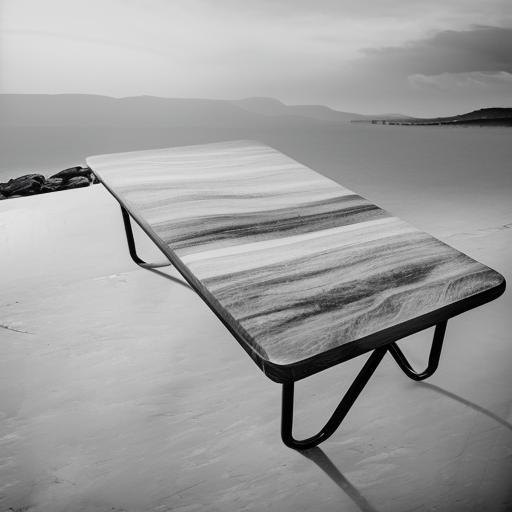} \\

%---------------------
\includegraphics[width=\ablimgwidth]{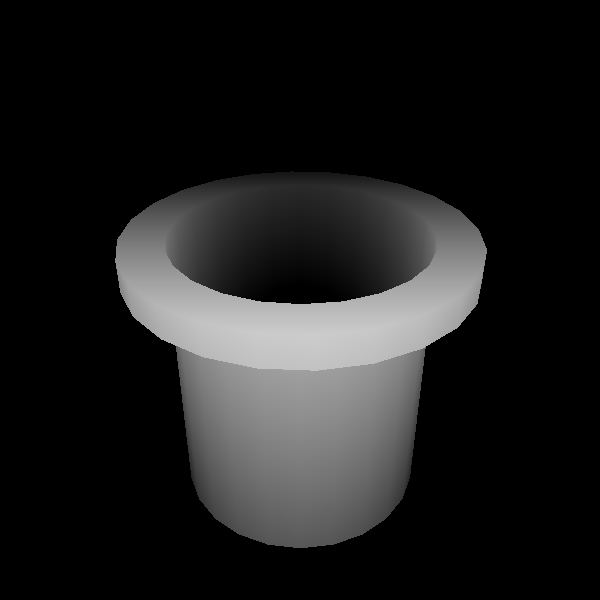} &
\includegraphics[width=\ablimgwidth]{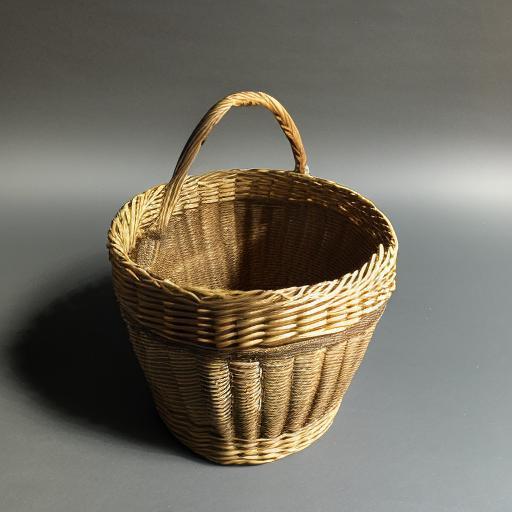} &
\includegraphics[width=\ablimgwidth]{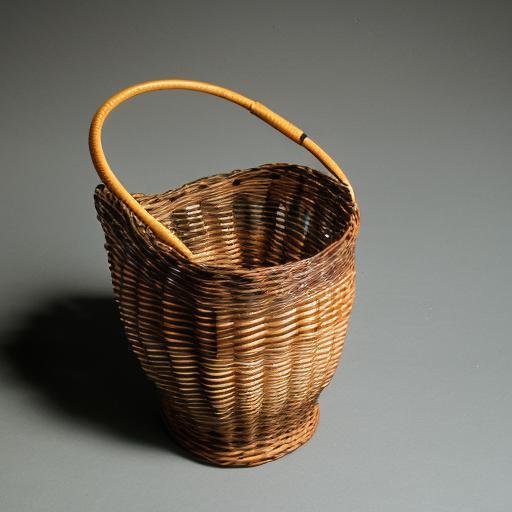} &
\includegraphics[width=\ablimgwidth]{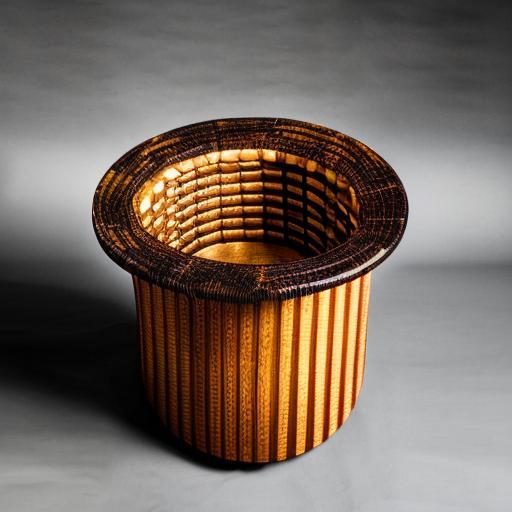} &
\includegraphics[width=\ablimgwidth]{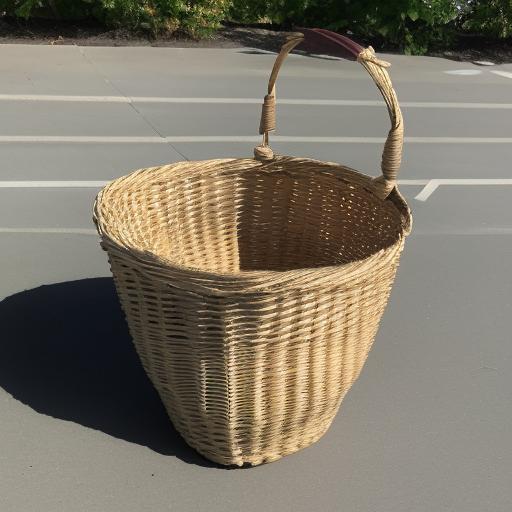} &
\includegraphics[width=\ablimgwidth]{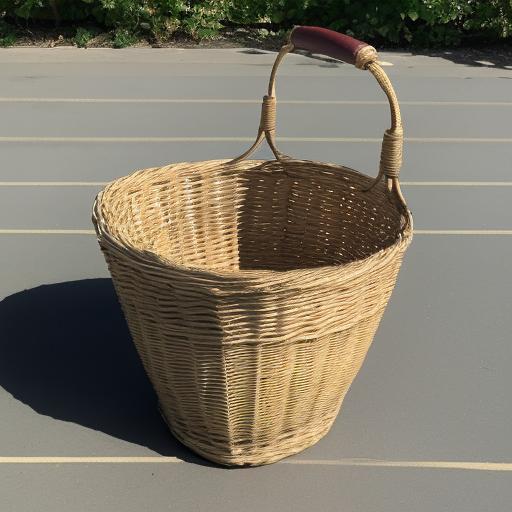} &
\includegraphics[width=\ablimgwidth]{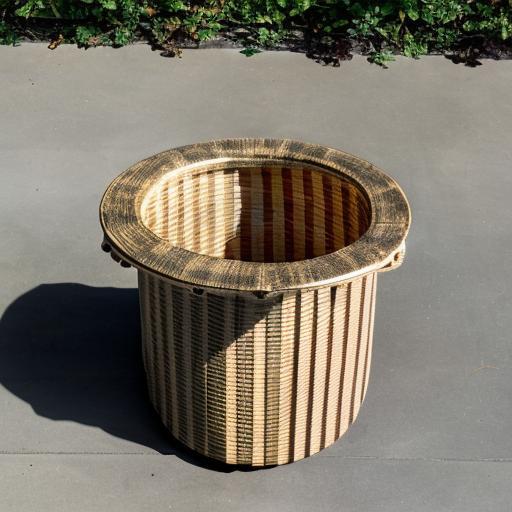} &
\includegraphics[width=\ablimgwidth]{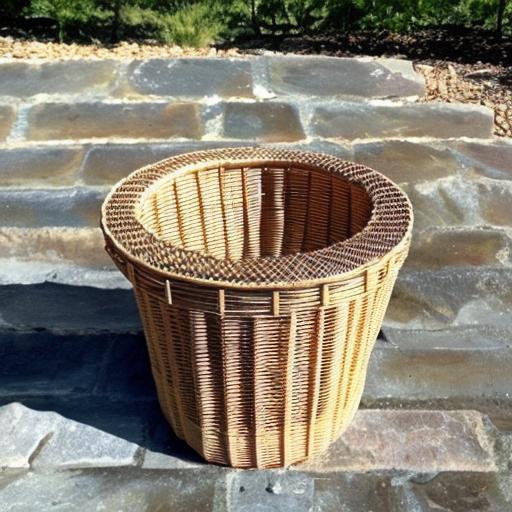} &
\includegraphics[width=\ablimgwidth]{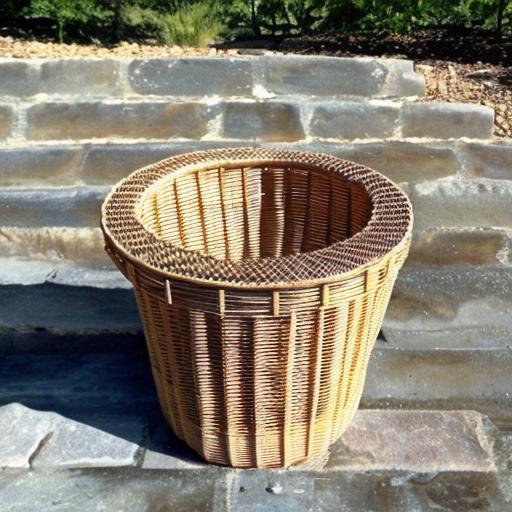} &
\includegraphics[width=\ablimgwidth]{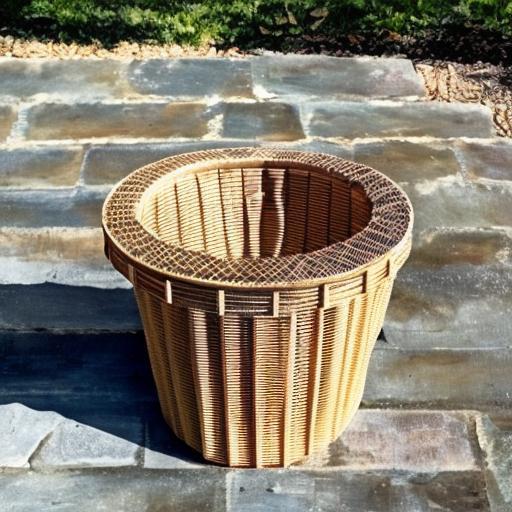} \\

%---------------------
\includegraphics[width=\ablimgwidth]{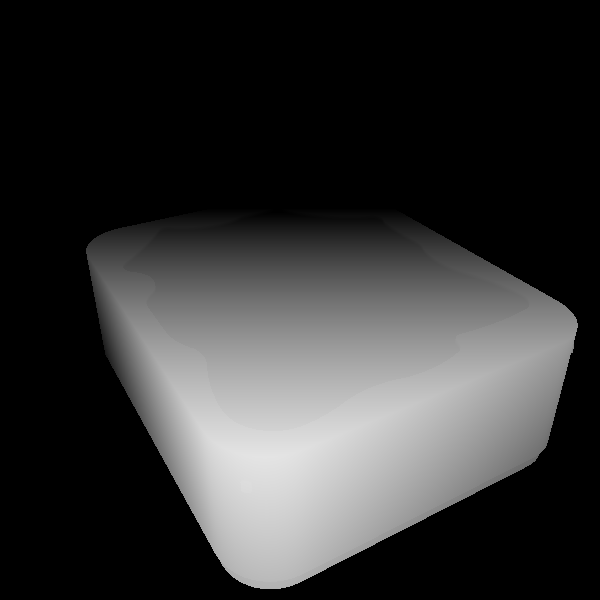} &
\includegraphics[width=\ablimgwidth]{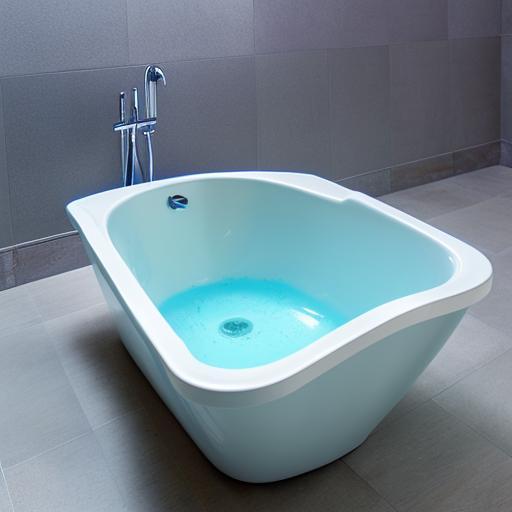} &
\includegraphics[width=\ablimgwidth]{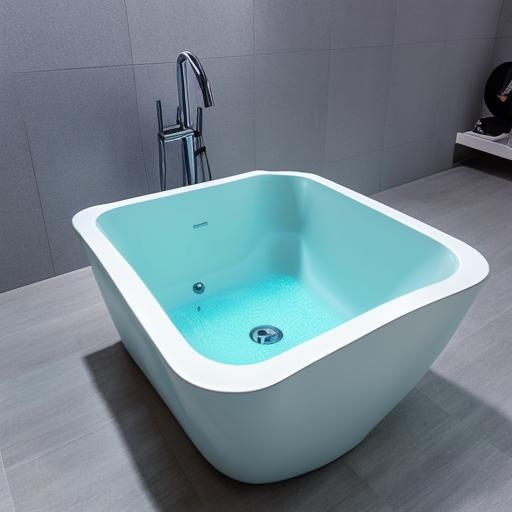} &
\includegraphics[width=\ablimgwidth]{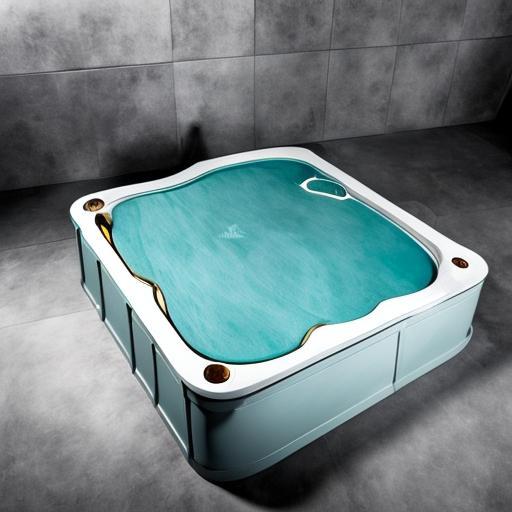} &
\includegraphics[width=\ablimgwidth]{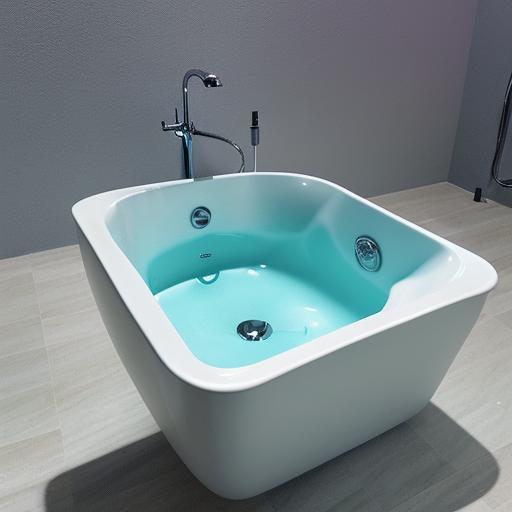} &
\includegraphics[width=\ablimgwidth]{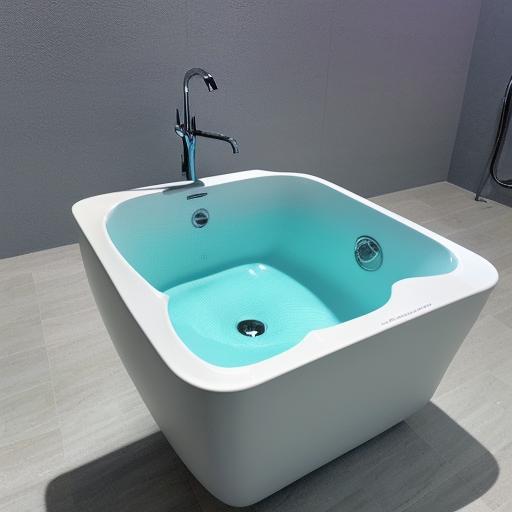} &
\includegraphics[width=\ablimgwidth]{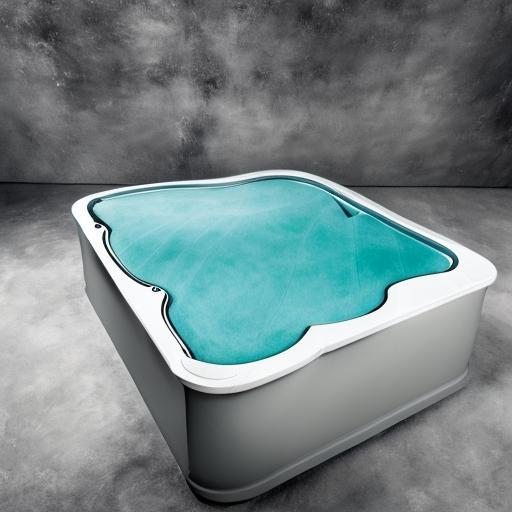} &
\includegraphics[width=\ablimgwidth]{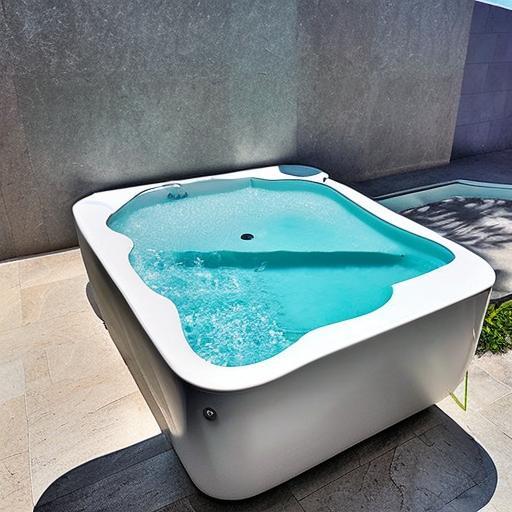} &
\includegraphics[width=\ablimgwidth]{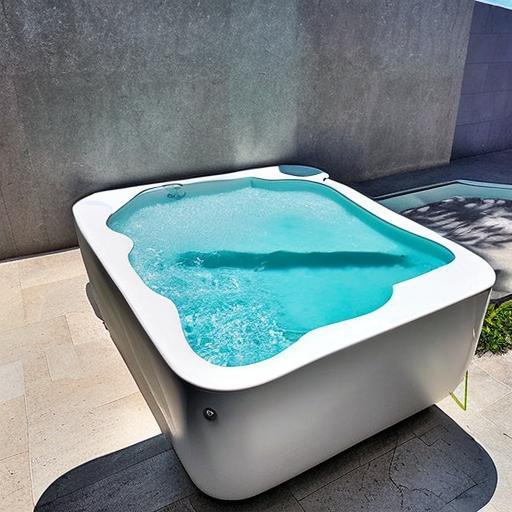} &
\includegraphics[width=\ablimgwidth]{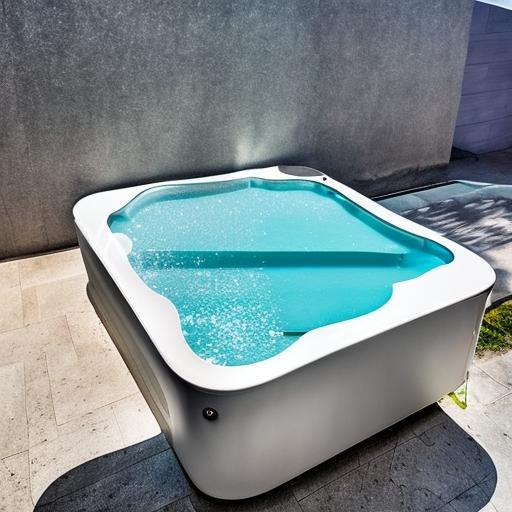} \\

%---------------------
\includegraphics[width=\ablimgwidth]{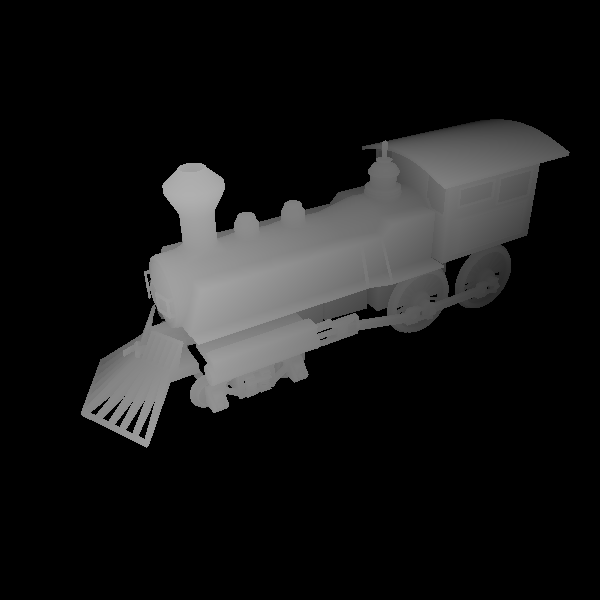} &
\includegraphics[width=\ablimgwidth]{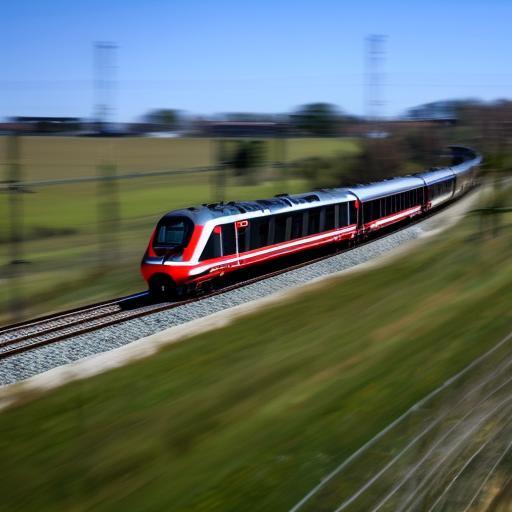} &
\includegraphics[width=\ablimgwidth]{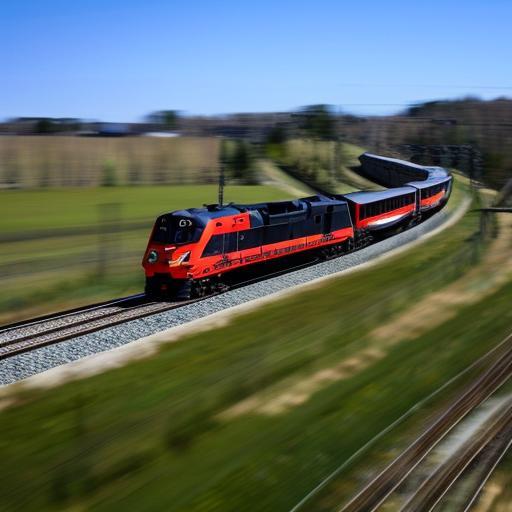} &
\includegraphics[width=\ablimgwidth]{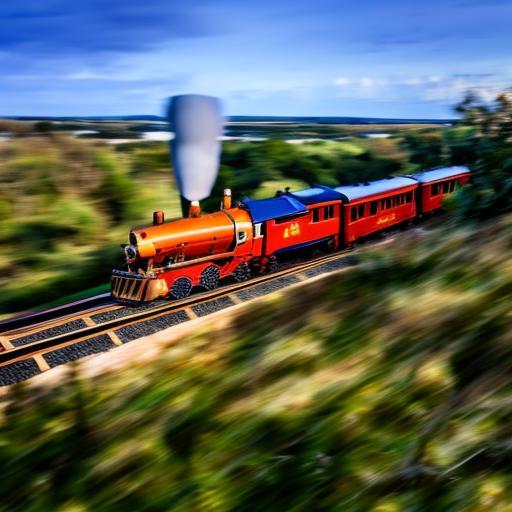} &
\includegraphics[width=\ablimgwidth]{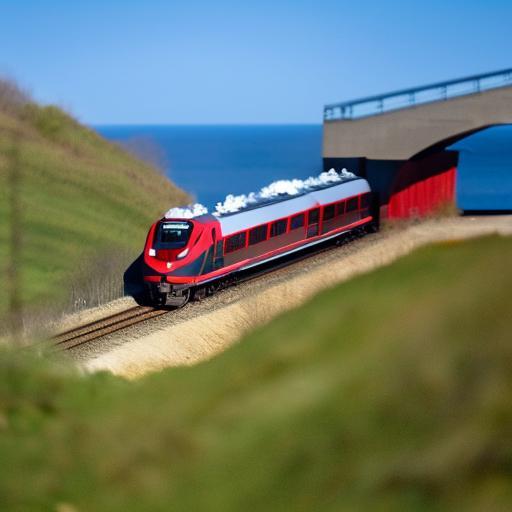} &
\includegraphics[width=\ablimgwidth]{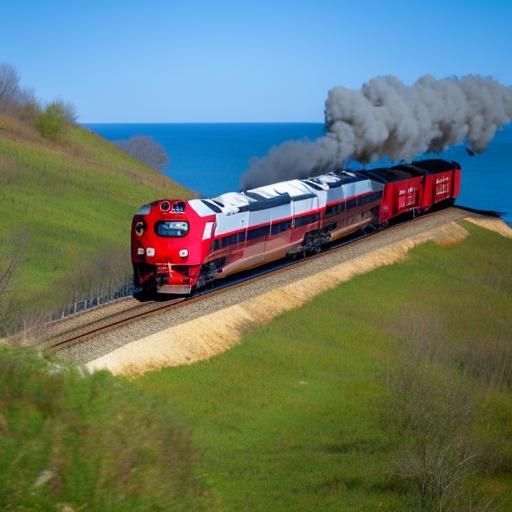} &
\includegraphics[width=\ablimgwidth]{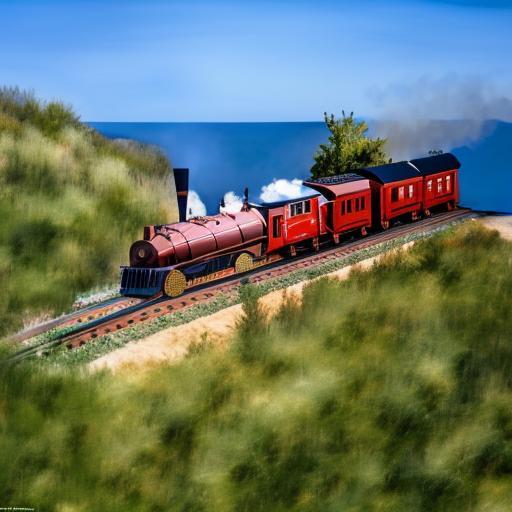} &
\includegraphics[width=\ablimgwidth]{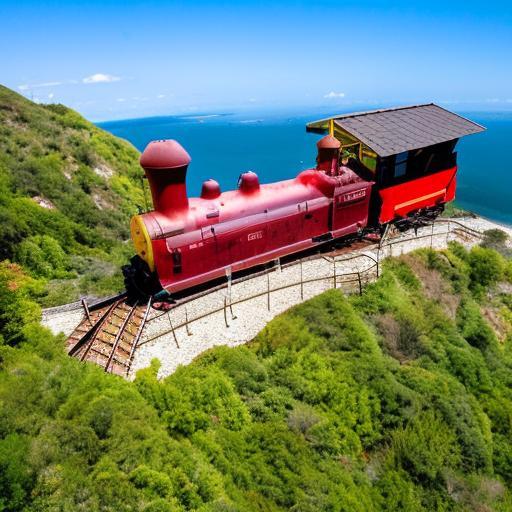} &
\includegraphics[width=\ablimgwidth]{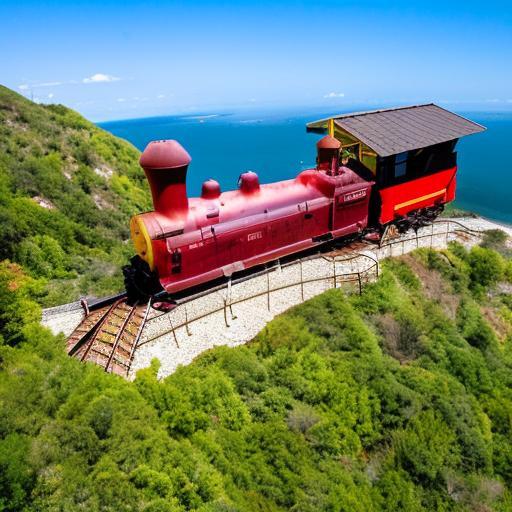} &
\includegraphics[width=\ablimgwidth]{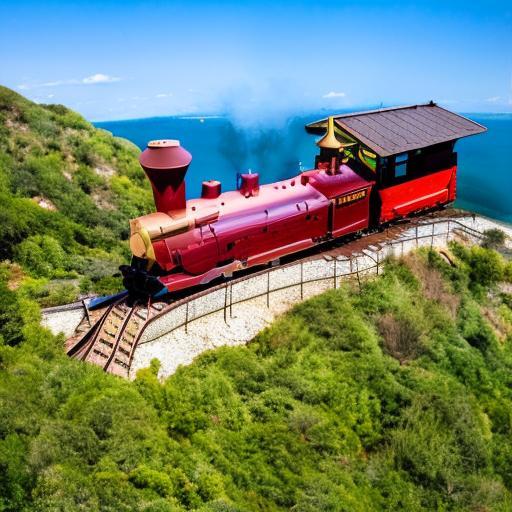} \\

%--------------------- 
\includegraphics[width=\ablimgwidth]{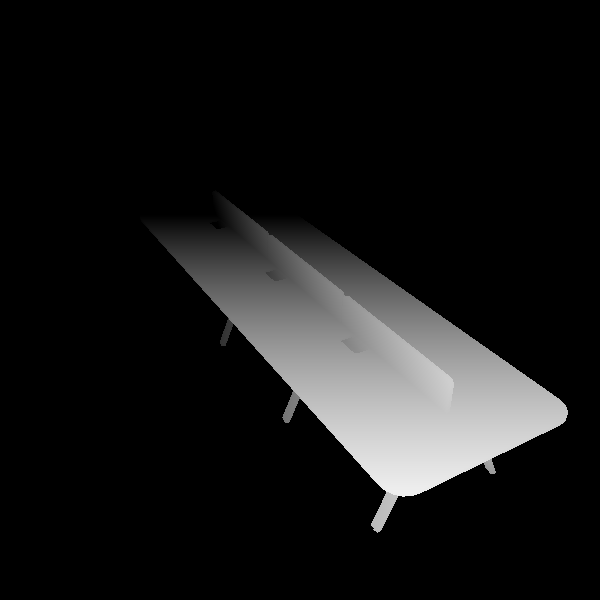} &
\includegraphics[width=\ablimgwidth]{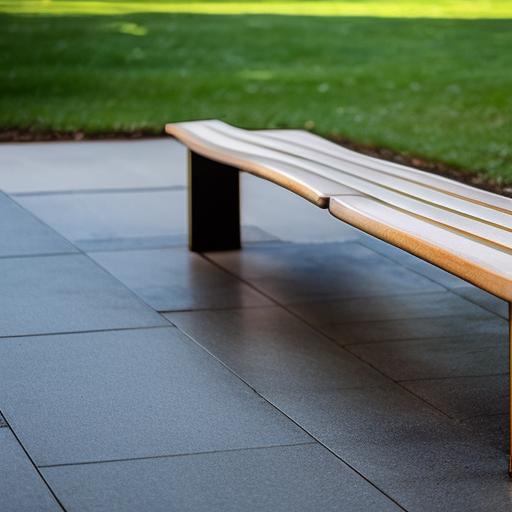} &
\includegraphics[width=\ablimgwidth]{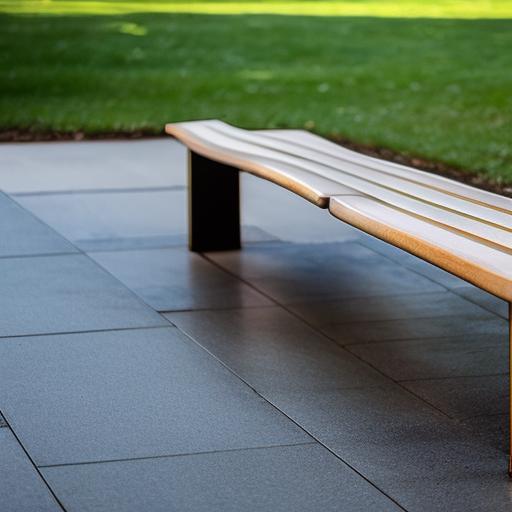} &
\includegraphics[width=\ablimgwidth]{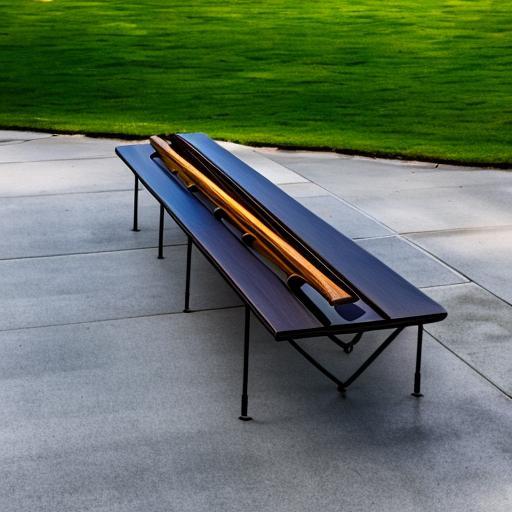} &
\includegraphics[width=\ablimgwidth]{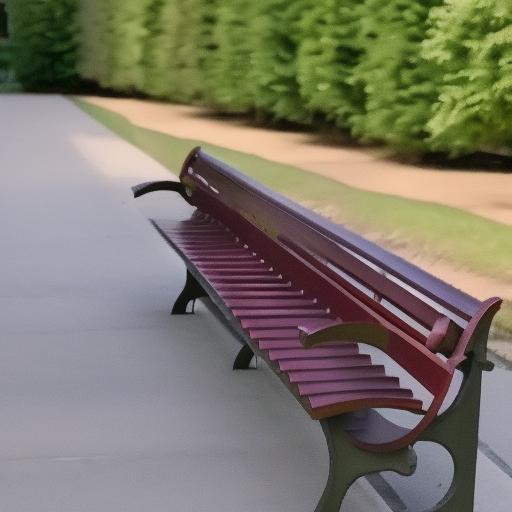} &
\includegraphics[width=\ablimgwidth]{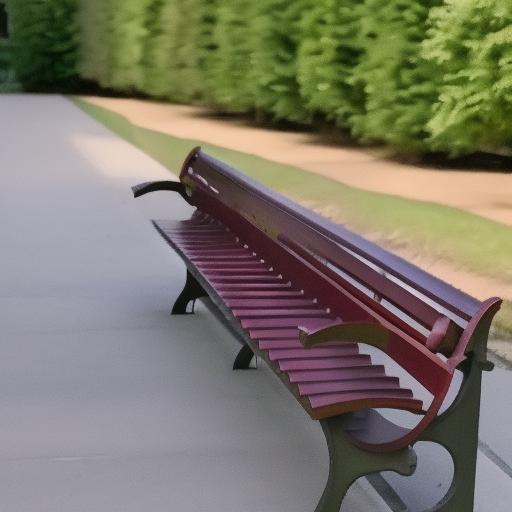} &
\includegraphics[width=\ablimgwidth]{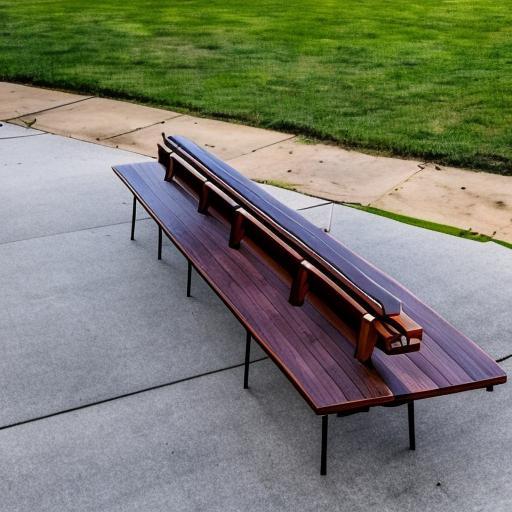} &
\includegraphics[width=\ablimgwidth]{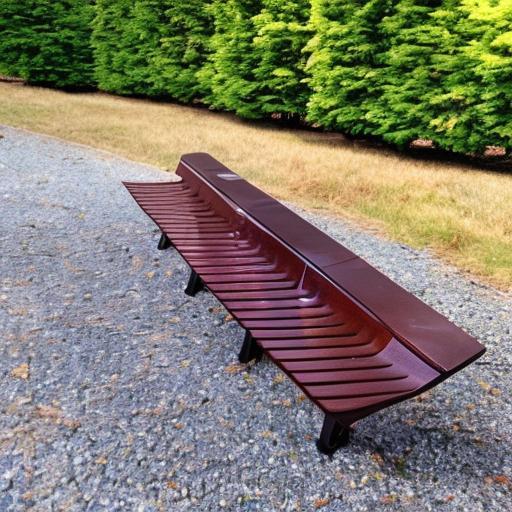} &
\includegraphics[width=\ablimgwidth]{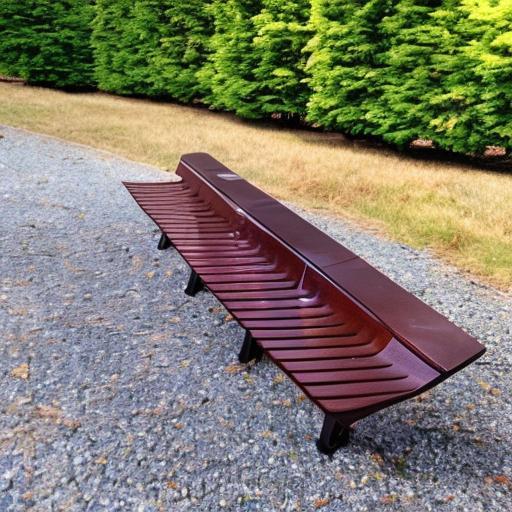} &
\includegraphics[width=\ablimgwidth]{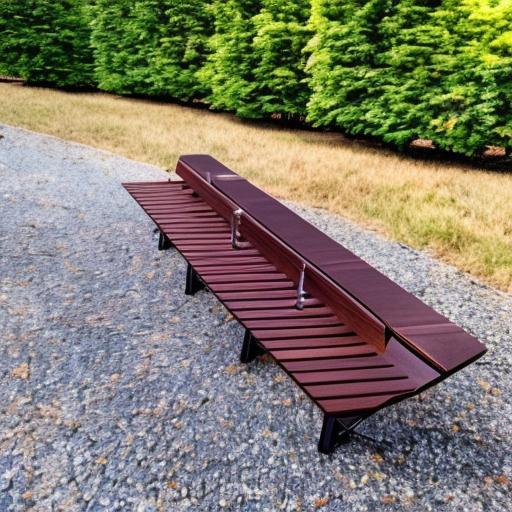} \\

\end{tabular}
\caption{\textbf{Shape adherence -- additional examples.} ShapeWords@20 produces shapes that are significantly more consistent with target shape geometry compared to the CNet-Stop@20 (conditioned either on category or subcategory prompts). ShapeWords@40 seems still more shape-adhering than CNet-Stop@40. In the setting of ShapeWords@80 and CNet-Stop@80, which both become more over-constrained by depth, differences become less noticeable. 
}
\label{fig:supp_geom_qual_eval}
\end{figure*}

%% file: figures/comp_eval_qualitative_supp.tex
\newlength{\suppcompimagewidth}
\setlength{\suppcompimagewidth}{0.11\textwidth} % Adjust this value as needed

% Define a new column type C based on \suppcompimagewidth
\newcolumntype{C}{>{\centering\arraybackslash}m{\suppcompimagewidth}}

% Set the color of table lines to grey
%\arrayrulecolor{gray}

\begin{figure*}[ht]
\centering
\setlength{\tabcolsep}{1pt} % Reduce horizontal padding between columns
\begin{tabular}{C*{7}{C}}
%\toprule
% First Row: Empty cell and Shape labels with images
\textbf{\small{Prompt}} & % Empty cell (upper-left corner)
\parbox[c]{\suppcompimagewidth}{\centering \textbf{\small{Depth}}} &
\parbox[c]{\suppcompimagewidth}{\centering \textbf{\small{CNet-Stop@30}}} &
\parbox[c]{\suppcompimagewidth}{\centering \textbf{\small{ControlNet}}} &
\parbox[c]{\suppcompimagewidth}{\centering \textbf{\small{CTRL-X@30}}} &
\parbox[c]{\suppcompimagewidth}{\centering \textbf{\small{CTRL-X@60}}} &
\parbox[c]{\suppcompimagewidth}{\centering \textbf{\small{ShapeWords}}} &
\parbox[c]{\suppcompimagewidth}{\centering \textbf{\small{Target Shape}}} \\
%\midrule
% Subsequent Rows: Prompts and generated images
% \small{`A \textbf{bag} under a tree'}  &
% \includegraphics[width=\suppcompimagewidth]{figures/img/quality_supp_final/02773838_f5800755a78fc83957be02cb1dc1e62/depth.jpg} & % Shape 1 image
% \includegraphics[width=\suppcompimagewidth]{figures/img/quality_supp_final/02773838_f5800755a78fc83957be02cb1dc1e62/controlnet30.jpg} & % Shape 2 image
% \includegraphics[width=\suppcompimagewidth]{figures/img/quality_supp_final/02773838_f5800755a78fc83957be02cb1dc1e62/controlnet1.0.jpg} & % Shape 3 image
% \includegraphics[width=\suppcompimagewidth]{figures/img/quality_supp_final/02773838_f5800755a78fc83957be02cb1dc1e62/ctrl3.jpg} & 
% \includegraphics[width=\suppcompimagewidth]{figures/img/quality_supp_final/02773838_f5800755a78fc83957be02cb1dc1e62/ctrl6.jpg} & 
% % Shape 4 image
% \includegraphics[width=\suppcompimagewidth]{figures/img/quality_supp_final/02773838_f5800755a78fc83957be02cb1dc1e62/ours.jpg} & % Shape 5 image
% \includegraphics[width=\suppcompimagewidth]{figures/img/quality_supp_final/02773838_f5800755a78fc83957be02cb1dc1e62/render.jpg}
% \\ % Shape 6 image

\small{`An artist painting a \textbf{bench}'}  &
\includegraphics[width=\suppcompimagewidth]{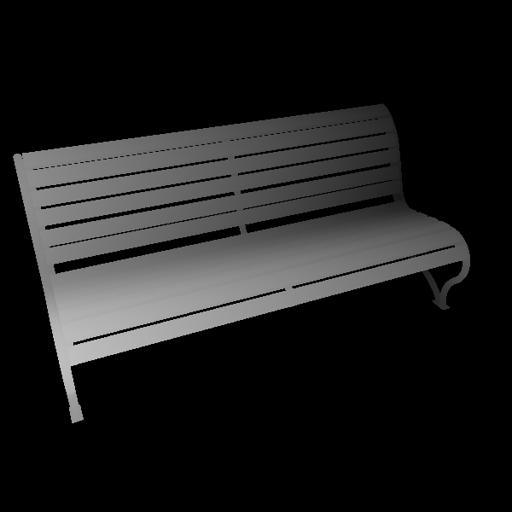} & % Shape 1 image
\includegraphics[width=\suppcompimagewidth]{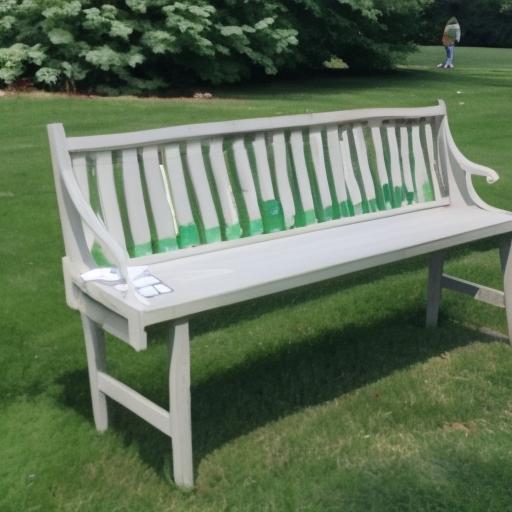} & % Shape 2 image
\includegraphics[width=\suppcompimagewidth]{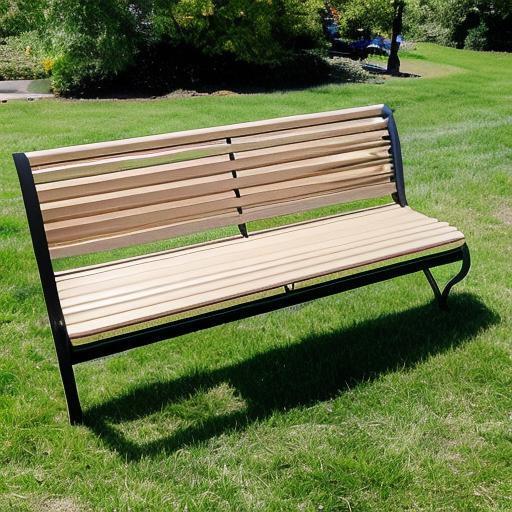} & % Shape 3 image
\includegraphics[width=\suppcompimagewidth]{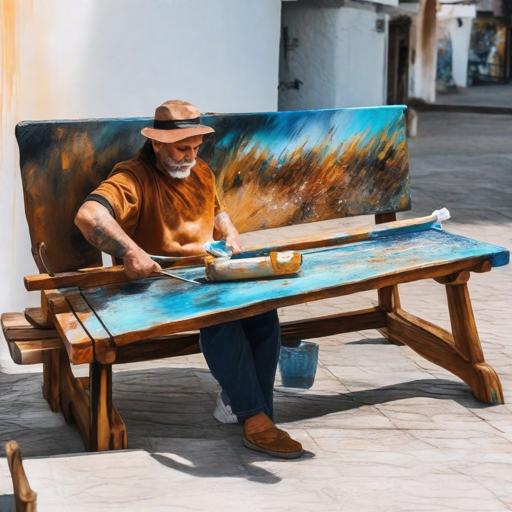} & 
\includegraphics[width=\suppcompimagewidth]{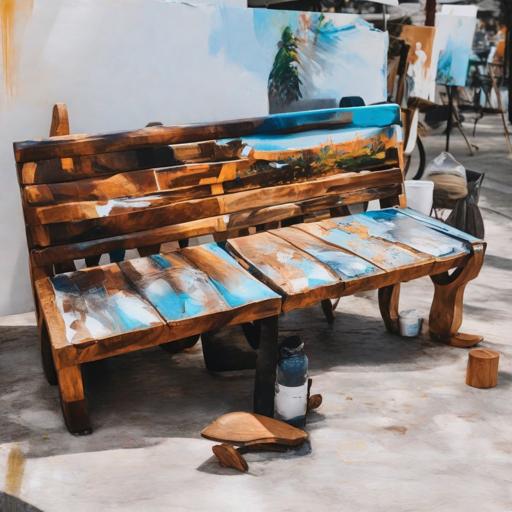} & 
% Shape 4 image
\includegraphics[width=\suppcompimagewidth]{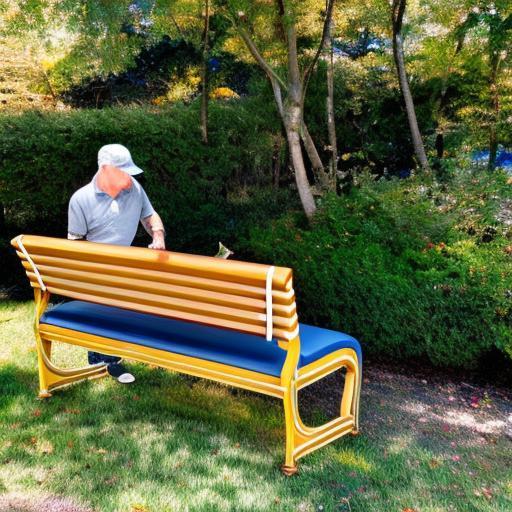} & % Shape 5 image
\includegraphics[width=\suppcompimagewidth]{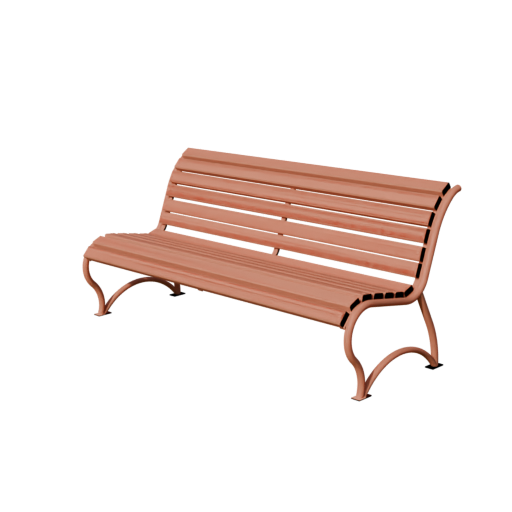}
\\ % Shape 6 image

% `A \textbf{car} in a snow globe'  &
% \includegraphics[width=\suppcompimagewidth]{figures/img/quality_supp_final/02958343_1ae184691a39e3d3e0e8bce75d28b114/depth.jpg} & % Shape 1 image
% \includegraphics[width=\suppcompimagewidth]{figures/img/quality_supp_final/02958343_1ae184691a39e3d3e0e8bce75d28b114/controlnet30.jpg} & % Shape 2 image
% \includegraphics[width=\suppcompimagewidth]{figures/img/quality_supp_final/02958343_1ae184691a39e3d3e0e8bce75d28b114/controlnet1.0.jpg} & % Shape 3 image
% \includegraphics[width=\suppcompimagewidth]{figures/img/quality_supp_final/02958343_1ae184691a39e3d3e0e8bce75d28b114/ctrl3.jpg} & 
% \includegraphics[width=\suppcompimagewidth]{figures/img/quality_supp_final/02958343_1ae184691a39e3d3e0e8bce75d28b114/ctrl6.jpg} & 
% % Shape 4 image
% \includegraphics[width=\suppcompimagewidth]{figures/img/quality_supp_final/02958343_1ae184691a39e3d3e0e8bce75d28b114/ours.jpg} & % Shape 5 image
% \includegraphics[width=\suppcompimagewidth]{figures/img/quality_supp_final/02958343_1ae184691a39e3d3e0e8bce75d28b114/render.jpg}
% \\ % Shape 6 image

\small{`A \textbf{car} under a tree'}  &
\includegraphics[width=\suppcompimagewidth]{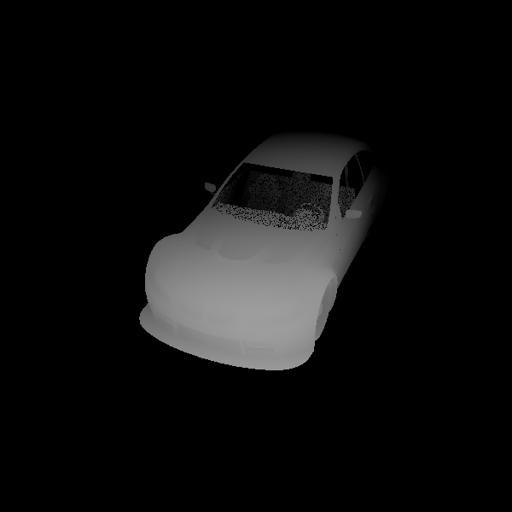} & % Shape 1 image
\includegraphics[width=\suppcompimagewidth]{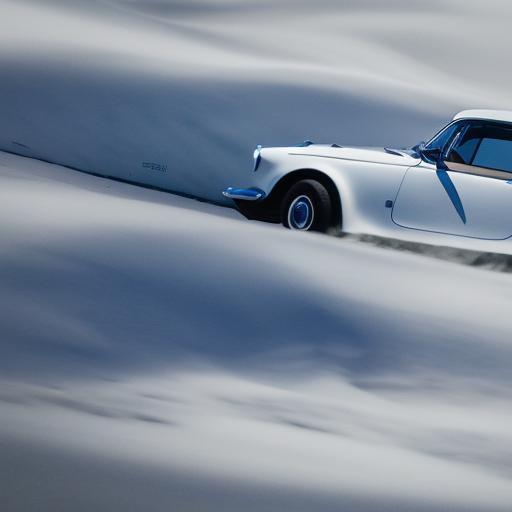} & % Shape 2 image
\includegraphics[width=\suppcompimagewidth]{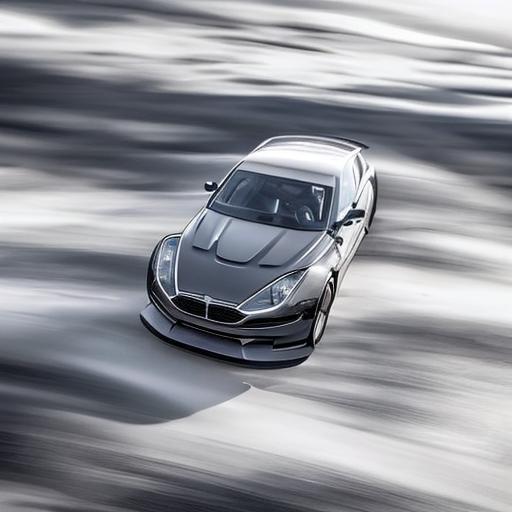} & % Shape 3 image
\includegraphics[width=\suppcompimagewidth]{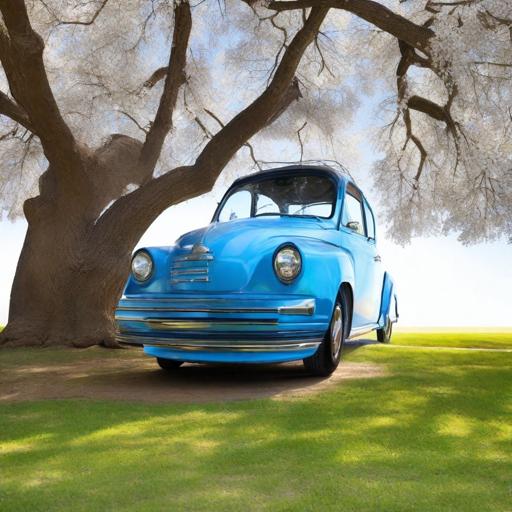} & 
\includegraphics[width=\suppcompimagewidth]{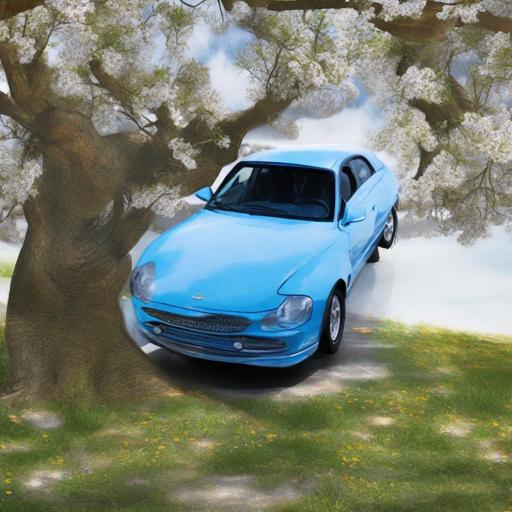} & 
% Shape 4 image
\includegraphics[width=\suppcompimagewidth]{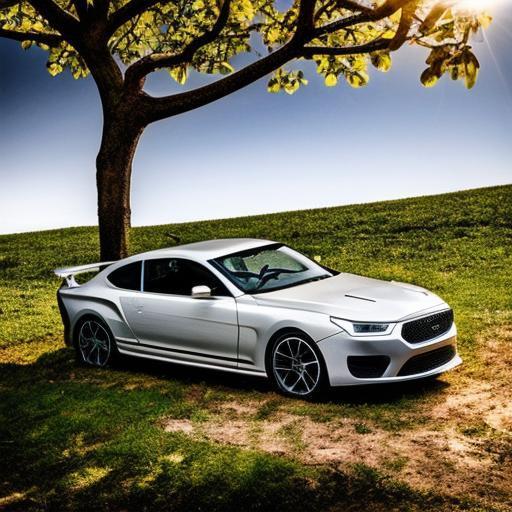} & % Shape 5 image
\includegraphics[width=\suppcompimagewidth]{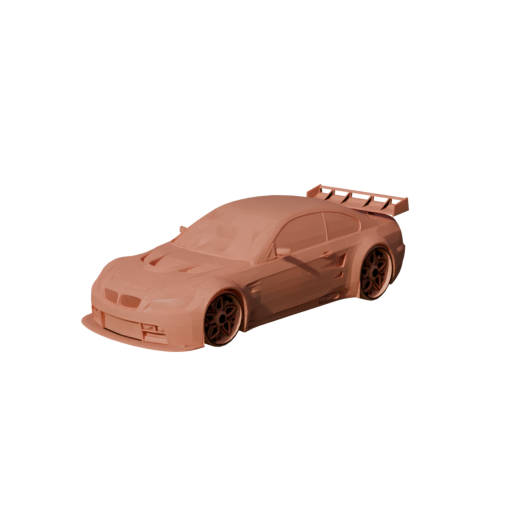}
\\ % Shape 6 image

\small{`A toy \textbf{car} in a box'}  &
\includegraphics[width=\suppcompimagewidth]{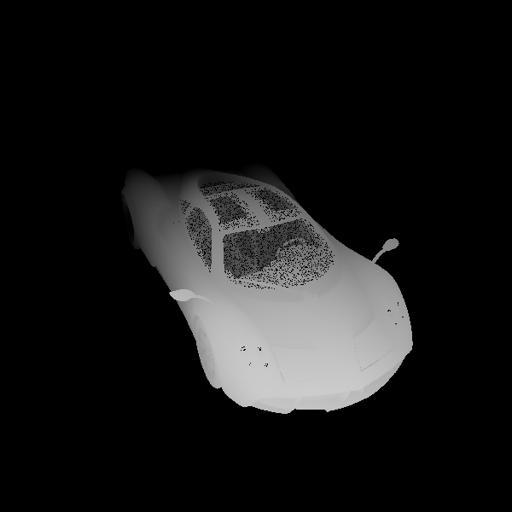} & % Shape 1 image
\includegraphics[width=\suppcompimagewidth]{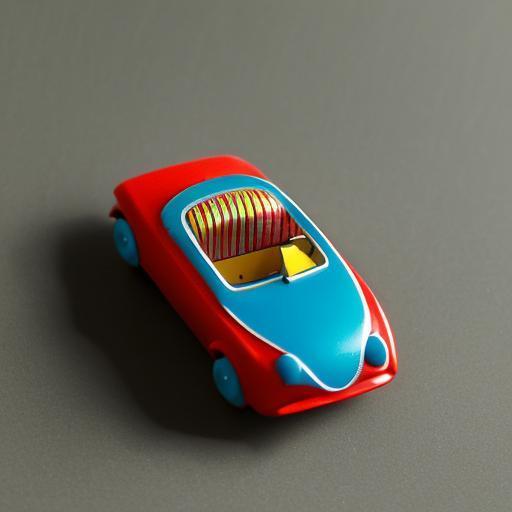} & % Shape 2 image
\includegraphics[width=\suppcompimagewidth]{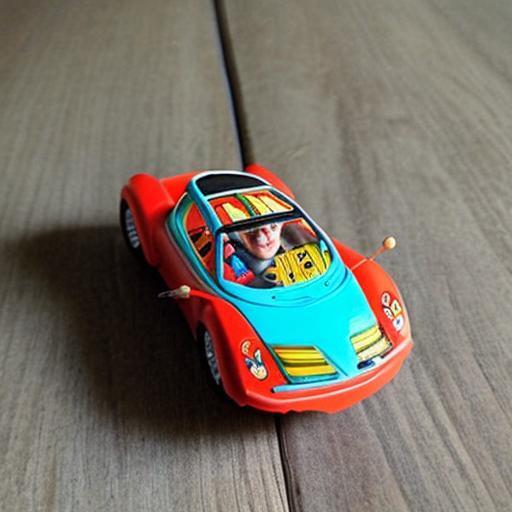} & % Shape 3 image
\includegraphics[width=\suppcompimagewidth]{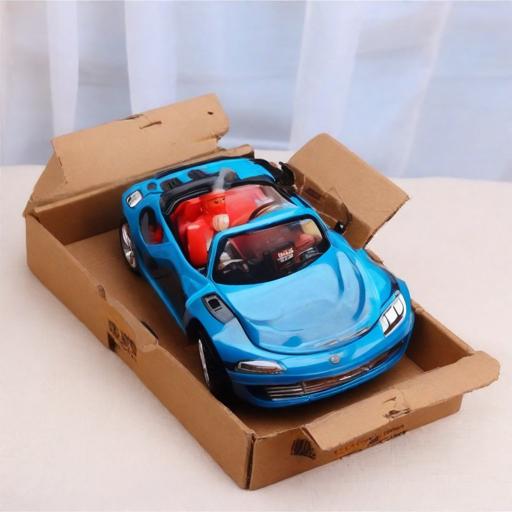} & 
\includegraphics[width=\suppcompimagewidth]{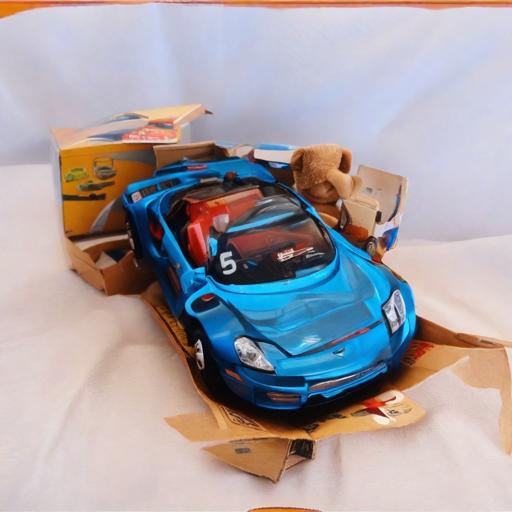} & 
% Shape 4 image
\includegraphics[width=\suppcompimagewidth]{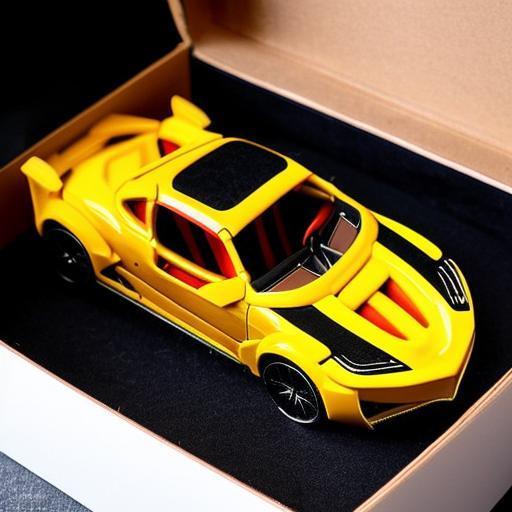} & % Shape 5 image
\includegraphics[width=\suppcompimagewidth]{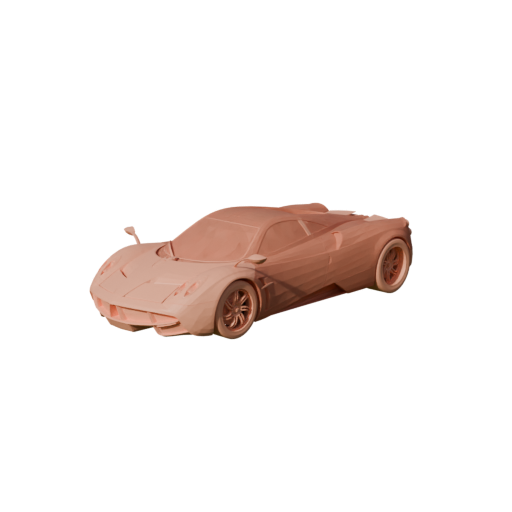}
\\ % Shape 6 image

%\small{`A \textbf{car} in a snow globe'}  &
%\includegraphics[width=\suppcompimagewidth]{figures/img/quality_supp_final/02958343_316086c3587f9c07d4db32f454eb6c4e/depth.jpg} & % Shape 1 image
%\includegraphics[width=\suppcompimagewidth]{figures/img/quality_supp_final/02958343_316086c3587f9c07d4db32f454eb6c4e/controlnet30.jpg} & % Shape 2 image
%\includegraphics[width=\suppcompimagewidth]{figures/img/quality_supp_final/02958343_316086c3587f9c07d4db32f454eb6c4e/controlnet1.0.jpg} & % Shape 3 image
%\includegraphics[width=\suppcompimagewidth]{figures/img/quality_supp_final/02958343_316086c3587f9c07d4db32f454eb6c4e/ctrl3.jpg} & 
%\includegraphics[width=\suppcompimagewidth]{figures/img/quality_supp_final/02958343_316086c3587f9c07d4db32f454eb6c4e/ctrl6.jpg} & 
% Shape 4 image
%\includegraphics[width=\suppcompimagewidth]{figures/img/quality_supp_final/02958343_316086c3587f9c07d4db32f454eb6c4e/ours.jpg} & % Shape 5 image
%\includegraphics[width=\suppcompimagewidth]{figures/img/quality_supp_final/02958343_316086c3587f9c07d4db32f454eb6c4e/render.jpg}
%\\ % Shape 6 image

\small{`A \textbf{chair} under a tree'}  &
\includegraphics[width=\suppcompimagewidth]{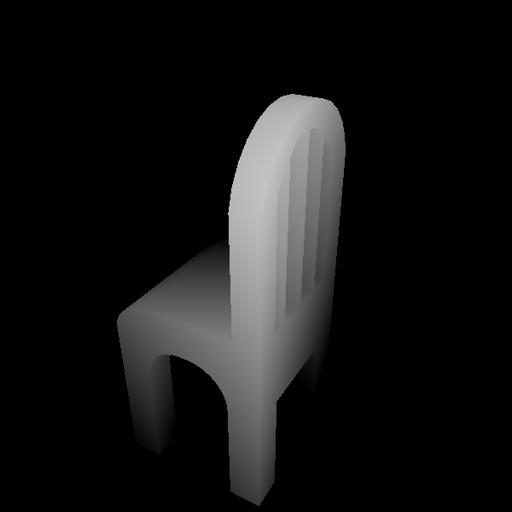} & % Shape 1 image
\includegraphics[width=\suppcompimagewidth]{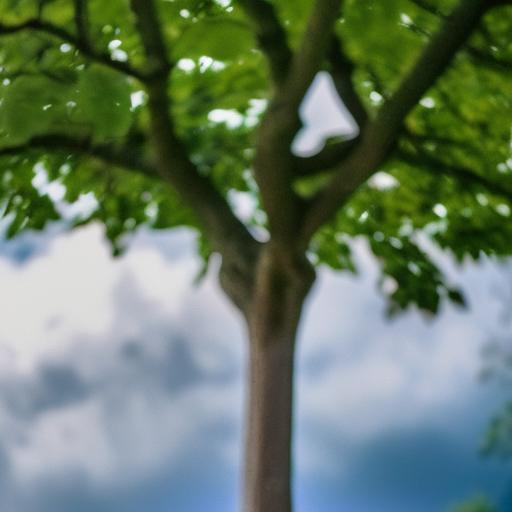} & % Shape 2 image
\includegraphics[width=\suppcompimagewidth]{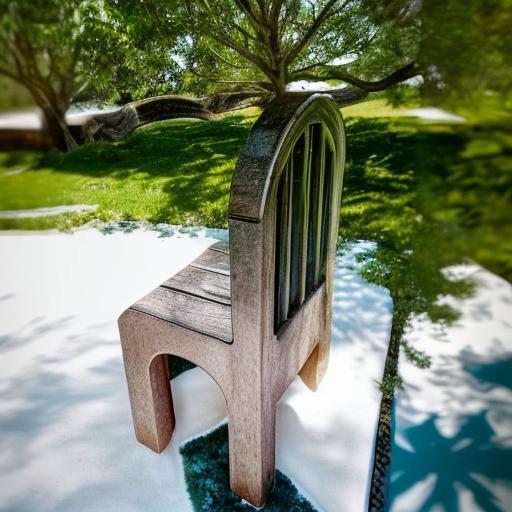} & % Shape 3 image
\includegraphics[width=\suppcompimagewidth]{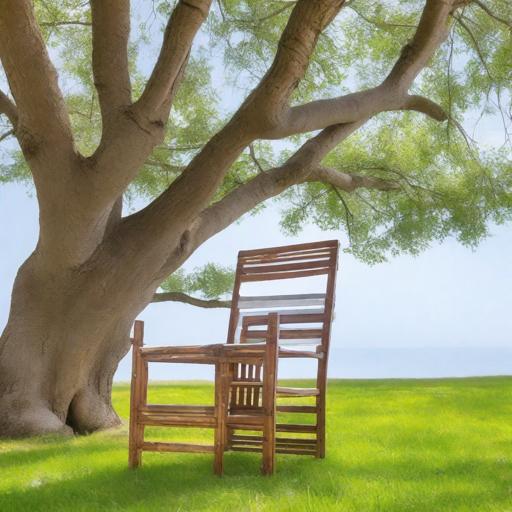} & 
\includegraphics[width=\suppcompimagewidth]{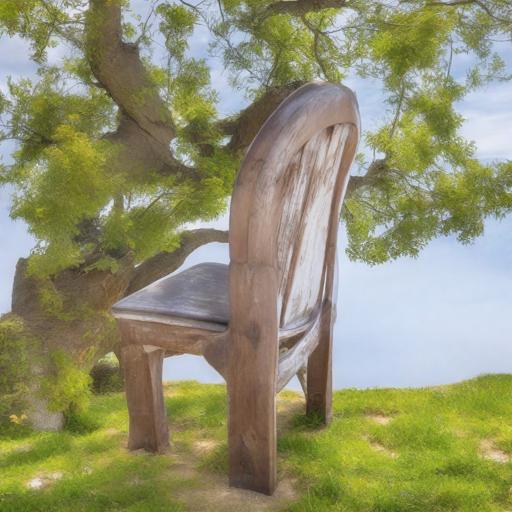} & 
% Shape 4 image
\includegraphics[width=\suppcompimagewidth]{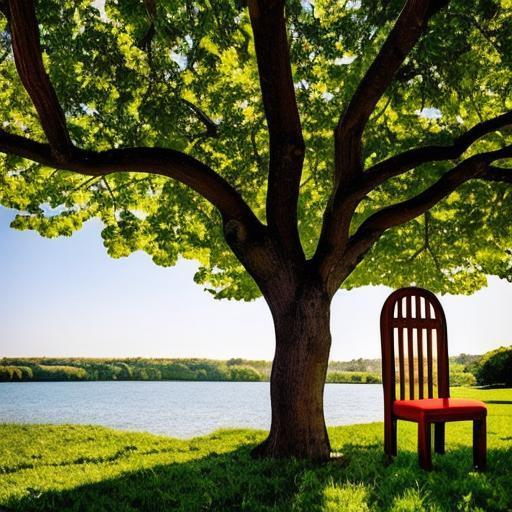} & % Shape 5 image
\includegraphics[width=\suppcompimagewidth]{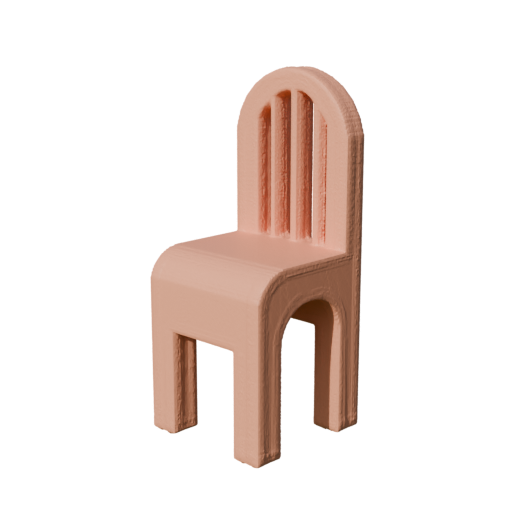}
\\ % Shape 6 image

\small{`A toy \textbf{guitar} in a box'}  &
\includegraphics[width=\suppcompimagewidth]{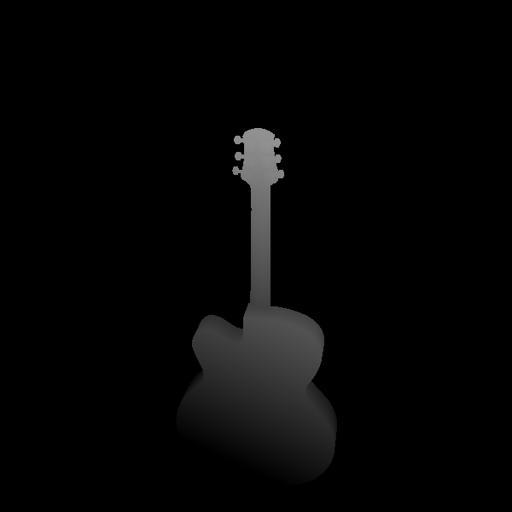} & % Shape 1 image
\includegraphics[width=\suppcompimagewidth]{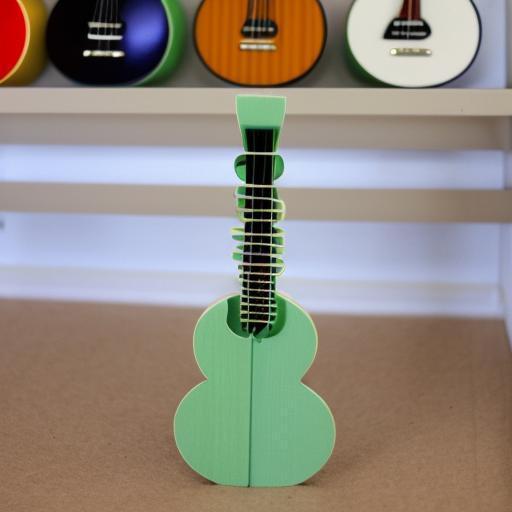} & % Shape 2 image
\includegraphics[width=\suppcompimagewidth]{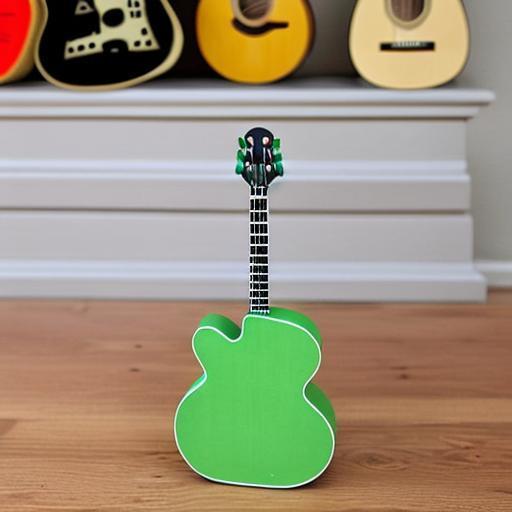} & % Shape 3 image
\includegraphics[width=\suppcompimagewidth]{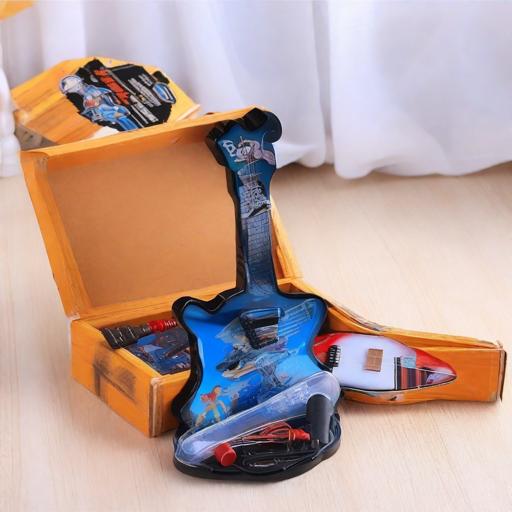} & 
\includegraphics[width=\suppcompimagewidth]{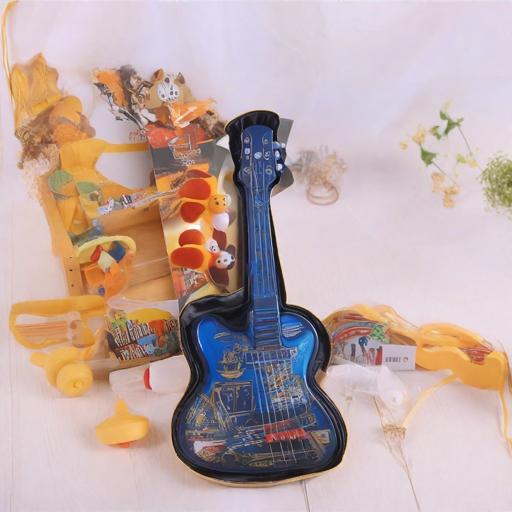} & 
% Shape 4 image
\includegraphics[width=\suppcompimagewidth]{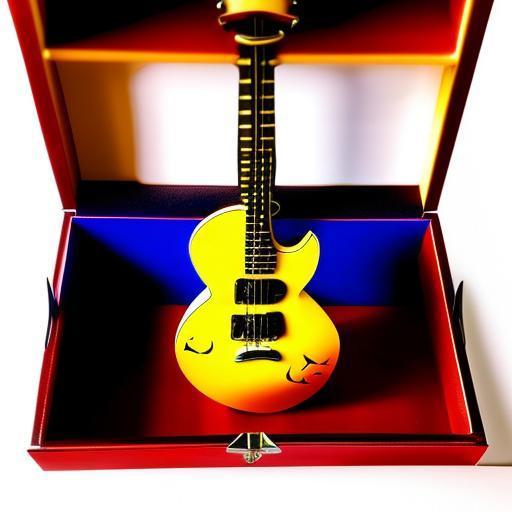} & % Shape 5 image
\includegraphics[width=\suppcompimagewidth]{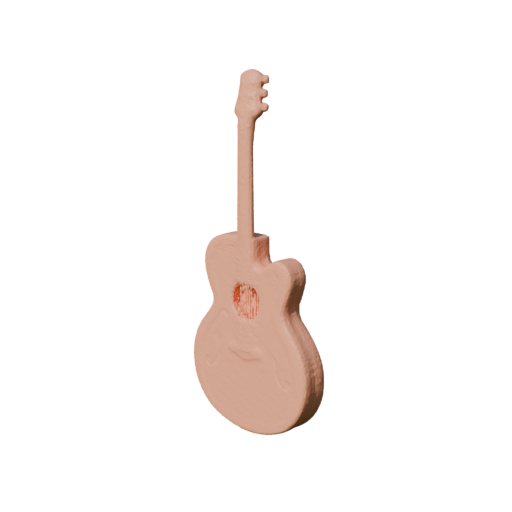}
\\ % Shape 6 image

\small{`An artist painting a \textbf{lamp}'}  &
\includegraphics[width=\suppcompimagewidth]{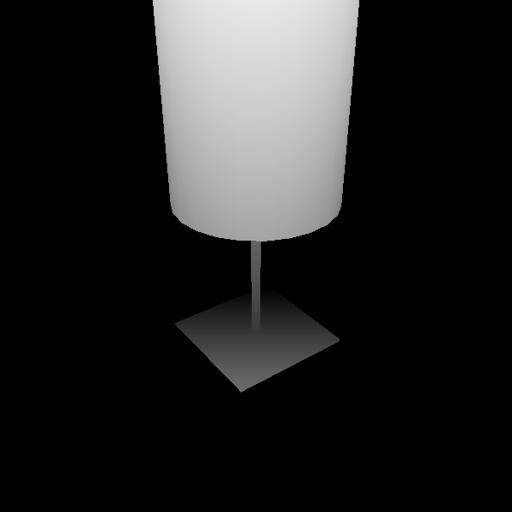} & % Shape 1 image
\includegraphics[width=\suppcompimagewidth]{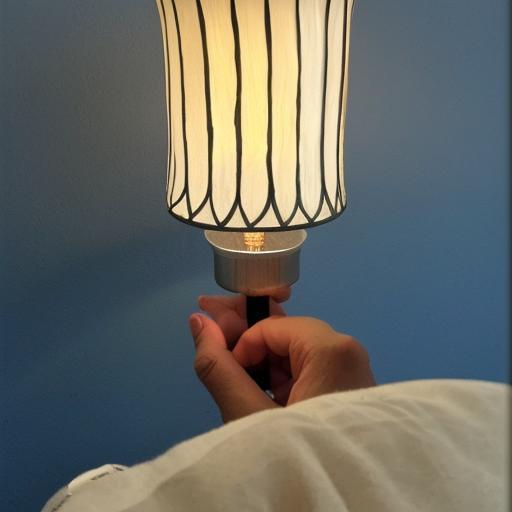} & % Shape 2 image
\includegraphics[width=\suppcompimagewidth]{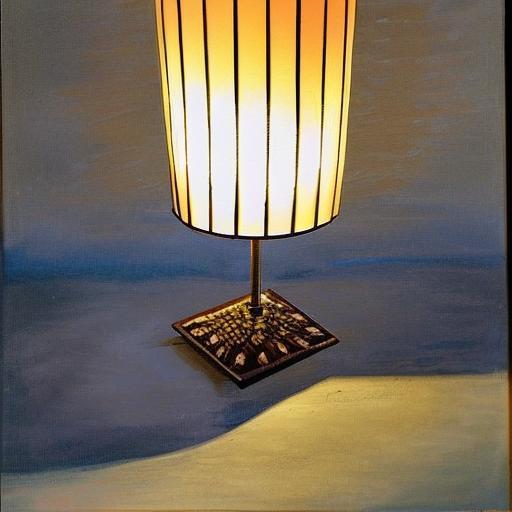} & % Shape 3 image
\includegraphics[width=\suppcompimagewidth]{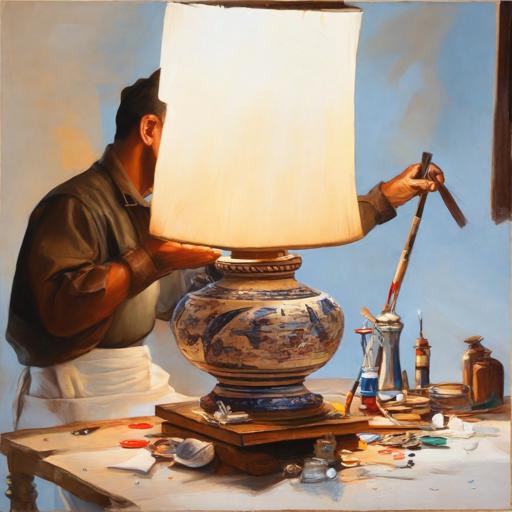} & 
\includegraphics[width=\suppcompimagewidth]{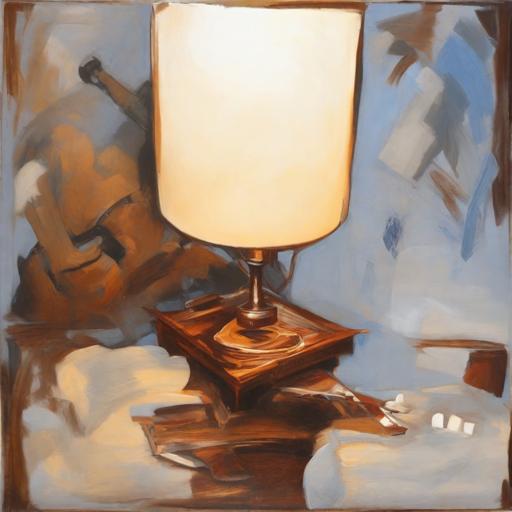} & 
% Shape 4 image
\includegraphics[width=\suppcompimagewidth]{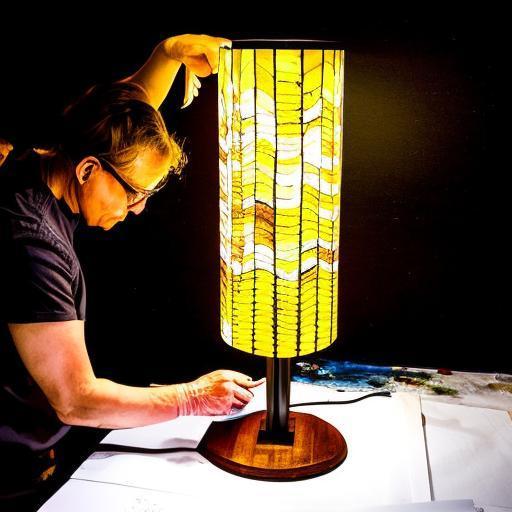} & % Shape 5 image
\includegraphics[width=\suppcompimagewidth]{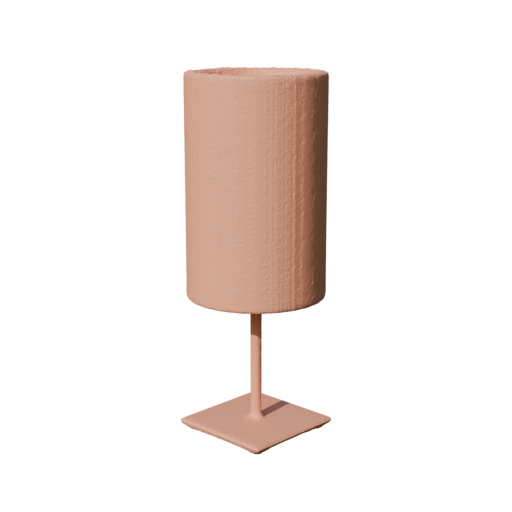}
\\ % Shape 6 image

\small{`A \textbf{table} under a tree'}  &
\includegraphics[width=\suppcompimagewidth]{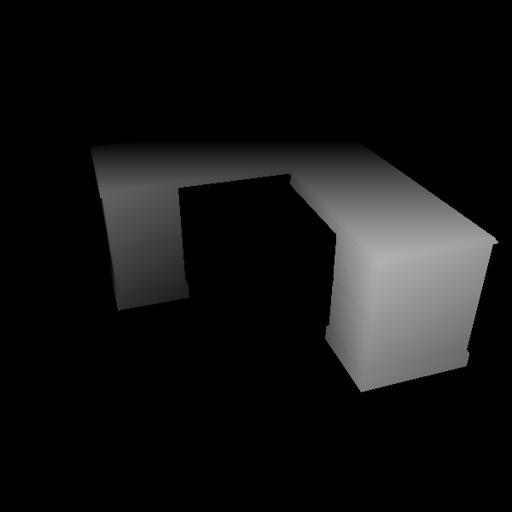} & % Shape 1 image
\includegraphics[width=\suppcompimagewidth]{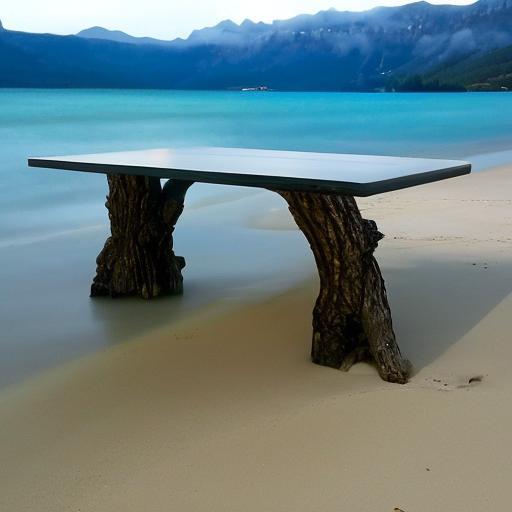} & % Shape 2 image
\includegraphics[width=\suppcompimagewidth]{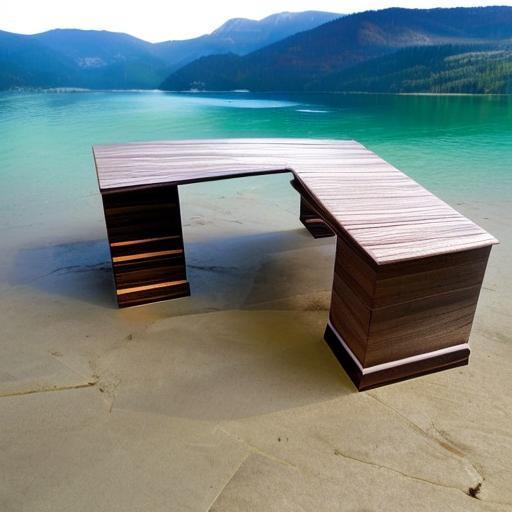} & % Shape 3 image
\includegraphics[width=\suppcompimagewidth]{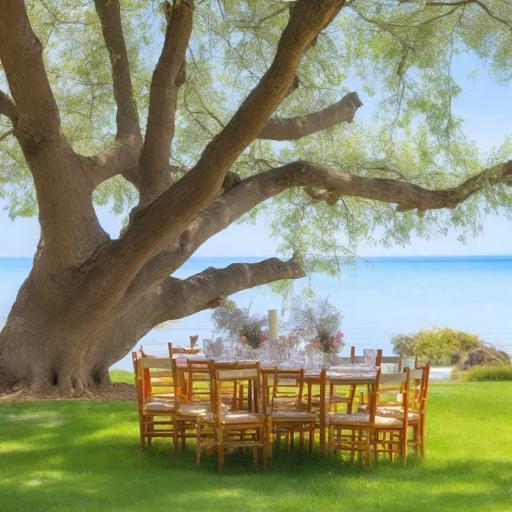} & 
\includegraphics[width=\suppcompimagewidth]{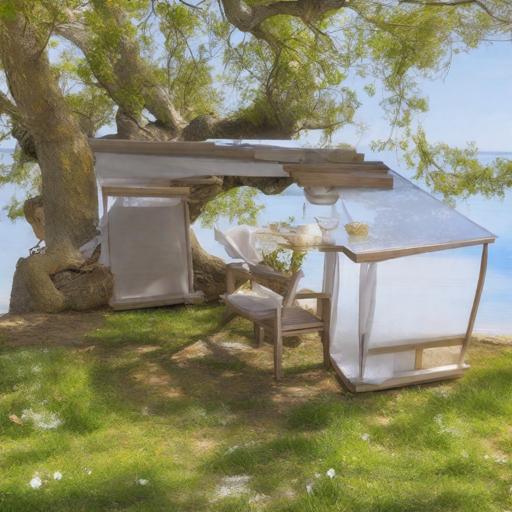} & 
% Shape 4 image
\includegraphics[width=\suppcompimagewidth]{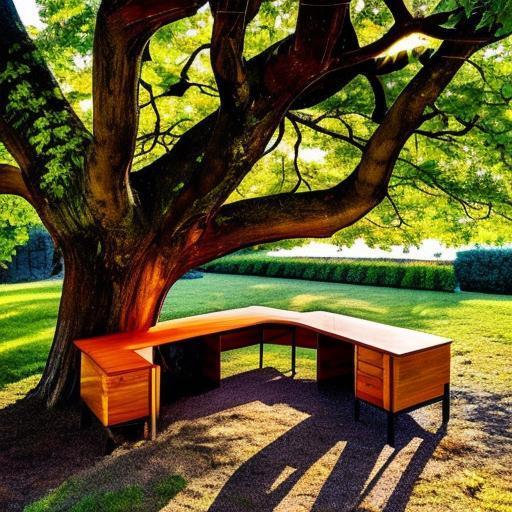} & % Shape 5 image
\includegraphics[width=\suppcompimagewidth]{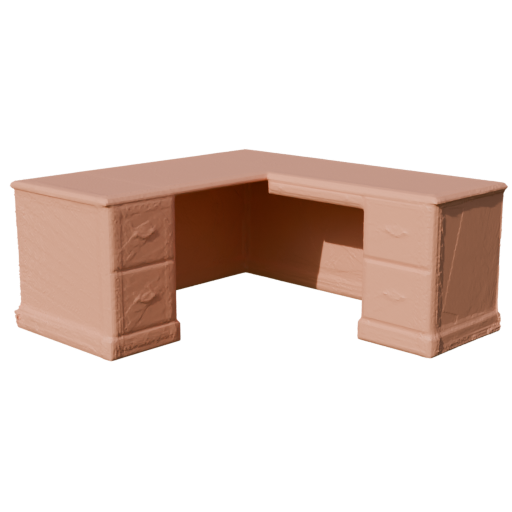}
\\ % Shape 6 image

\small{`A \textbf{train} under a tree'}  &
\includegraphics[width=\suppcompimagewidth]{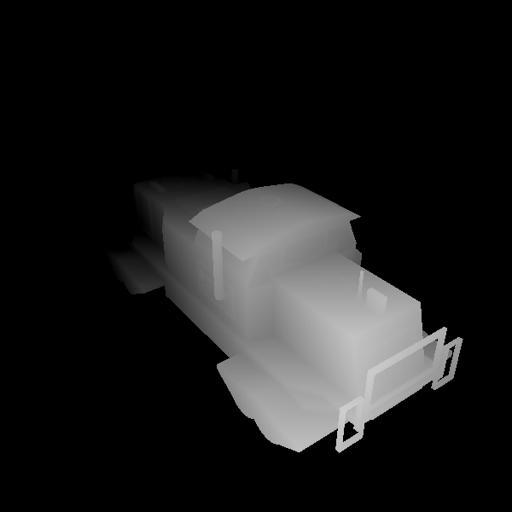} & % Shape 1 image
\includegraphics[width=\suppcompimagewidth]{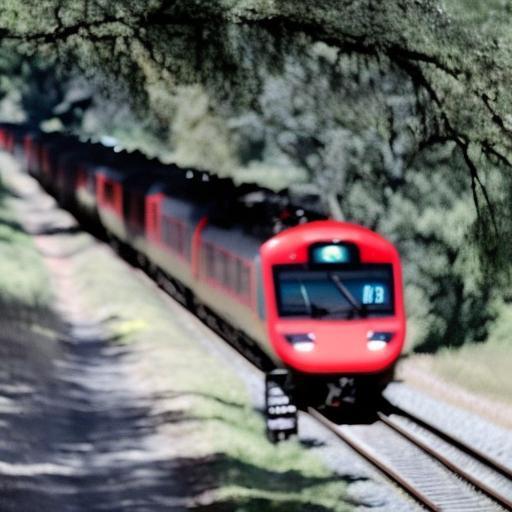} & % Shape 2 image
\includegraphics[width=\suppcompimagewidth]{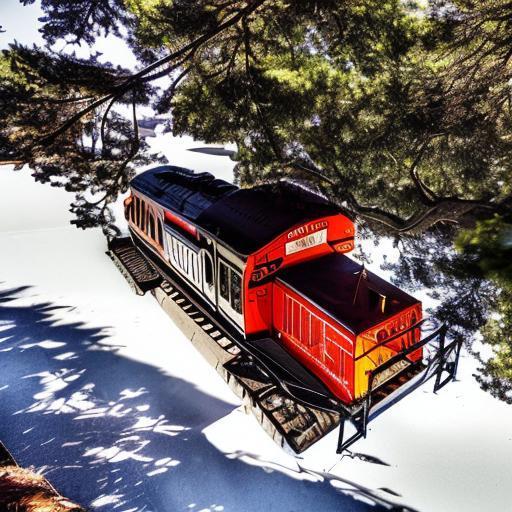} & % Shape 3 image
\includegraphics[width=\suppcompimagewidth]{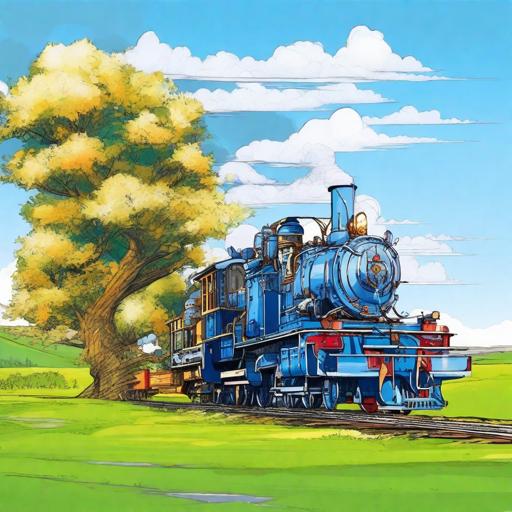} & 
\includegraphics[width=\suppcompimagewidth]{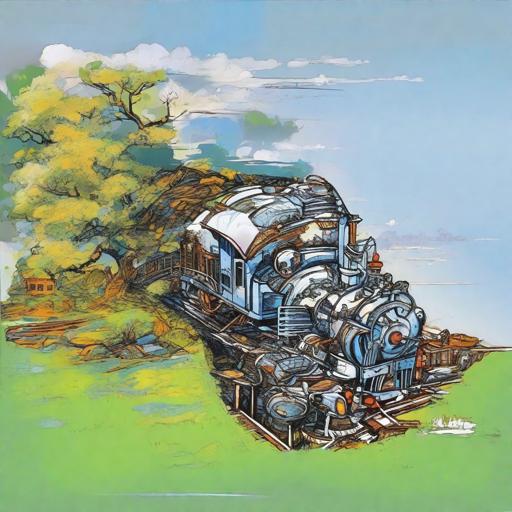} & 
% Shape 4 image
\includegraphics[width=\suppcompimagewidth]{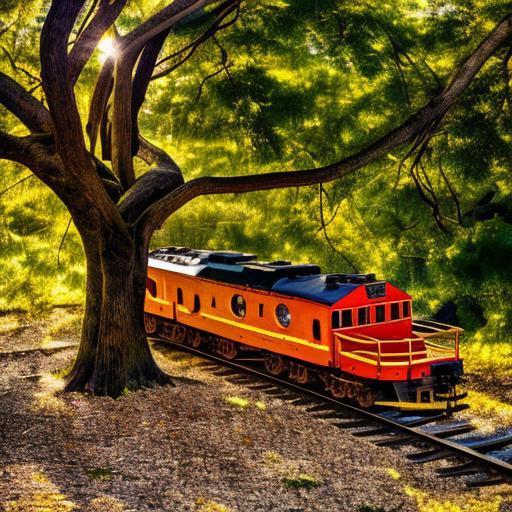} & % Shape 5 image
\includegraphics[width=\suppcompimagewidth]{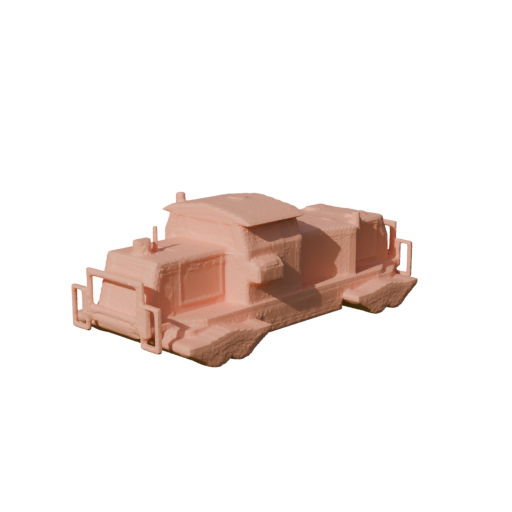}
\\ % Shape 6 image

\small{`A toy \textbf{train} in a box'}  &
\includegraphics[width=\suppcompimagewidth]{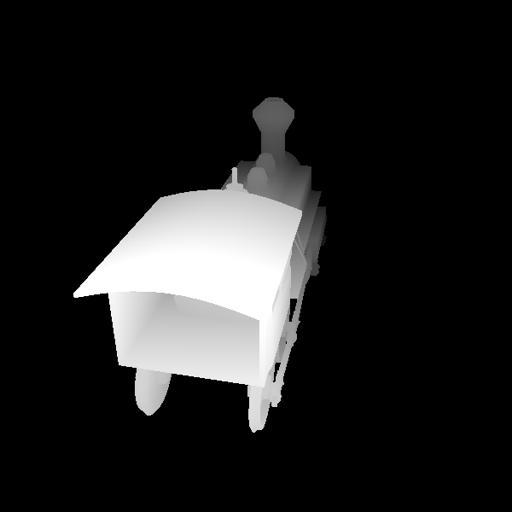} & % Shape 1 image
\includegraphics[width=\suppcompimagewidth]{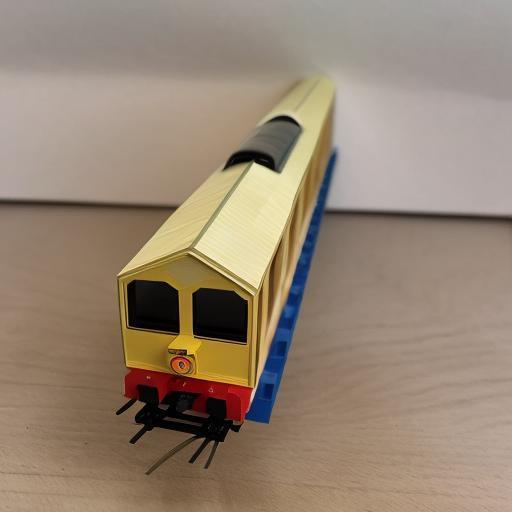} & % Shape 2 image
\includegraphics[width=\suppcompimagewidth]{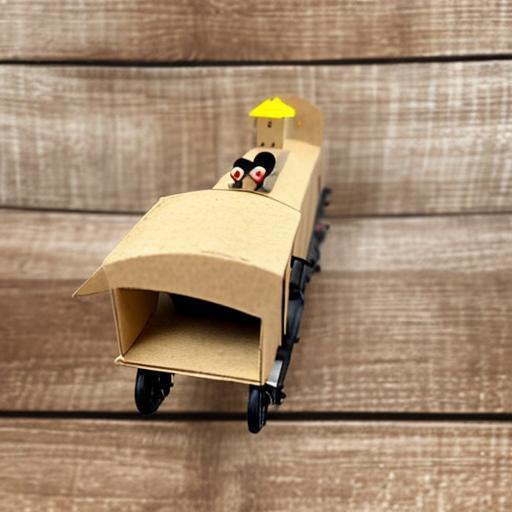} & % Shape 3 image
\includegraphics[width=\suppcompimagewidth]{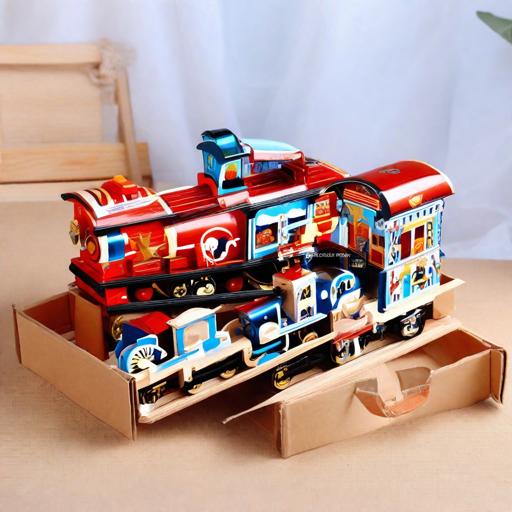} & 
\includegraphics[width=\suppcompimagewidth]{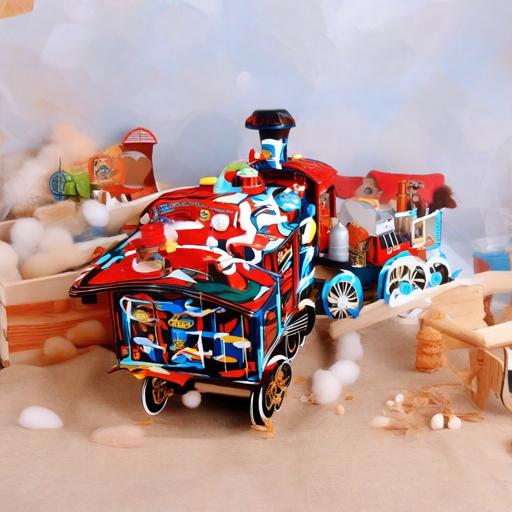} & 
% Shape 4 image
\includegraphics[width=\suppcompimagewidth]{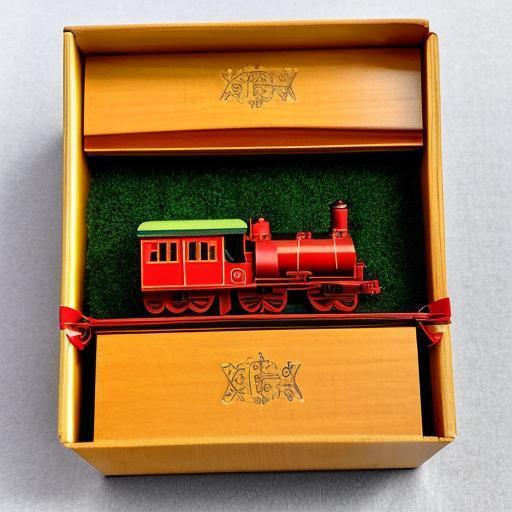} & % Shape 5 image
\includegraphics[width=\suppcompimagewidth]{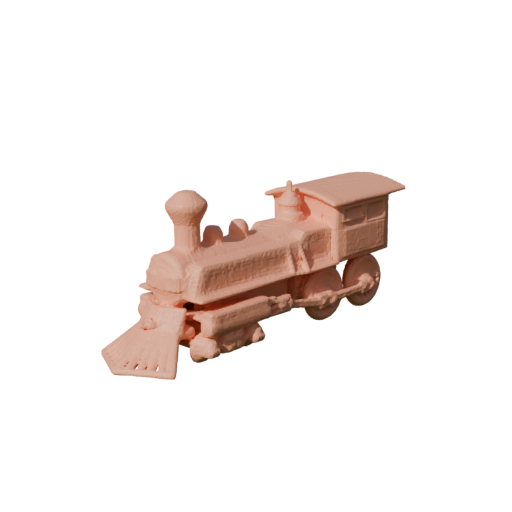}
\\ % Shape 6 image

\small{`A craftsman working on a \textbf{watercraft}'}  &
\includegraphics[width=\suppcompimagewidth]{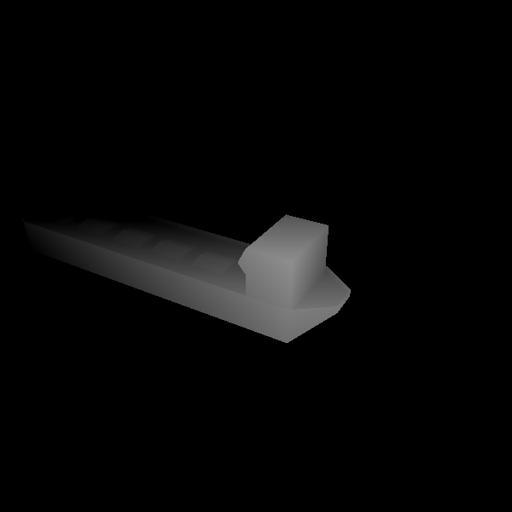} & % Shape 1 image
\includegraphics[width=\suppcompimagewidth]{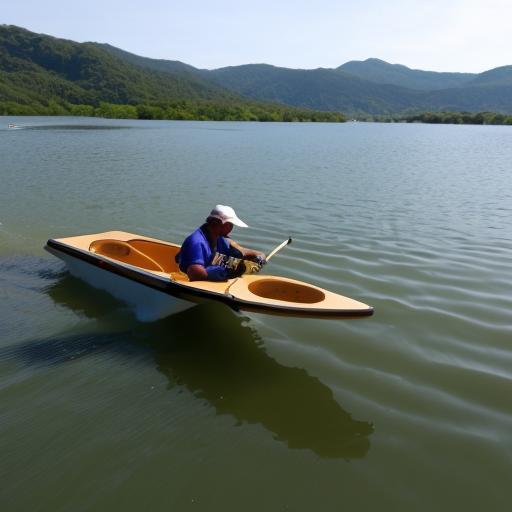} & % Shape 2 image
\includegraphics[width=\suppcompimagewidth]{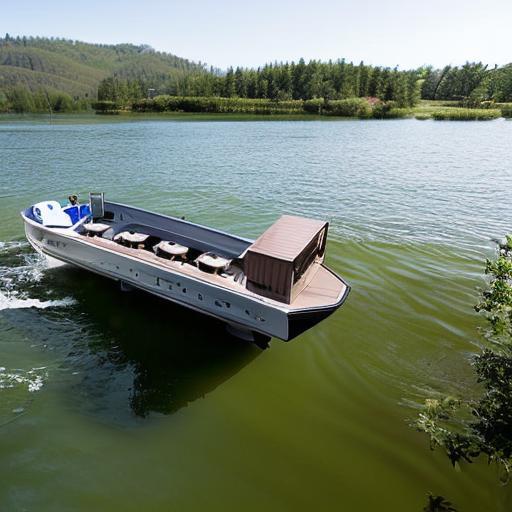} & % Shape 3 image
\includegraphics[width=\suppcompimagewidth]{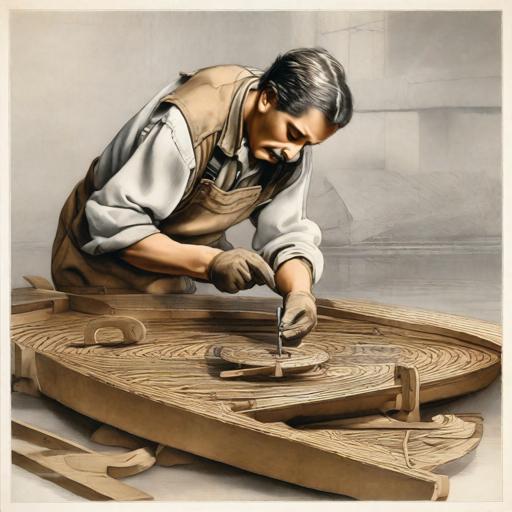} & 
\includegraphics[width=\suppcompimagewidth]{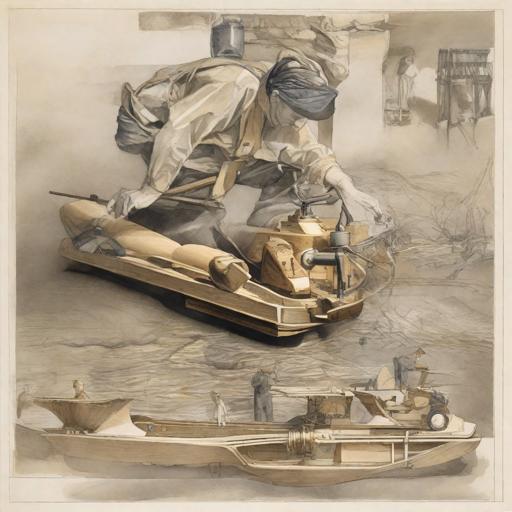} & 
% Shape 4 image
\includegraphics[width=\suppcompimagewidth]{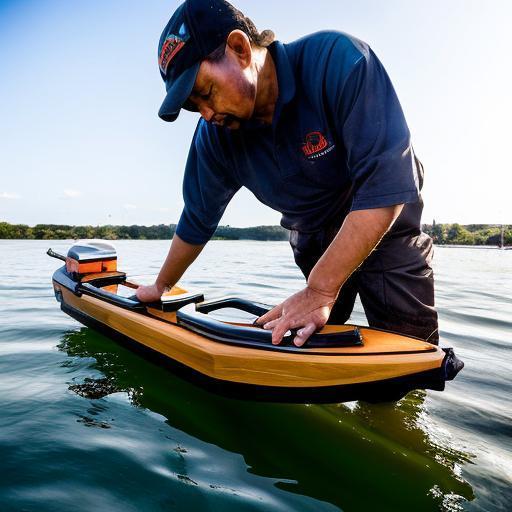} & % Shape 5 image
\includegraphics[width=\suppcompimagewidth]{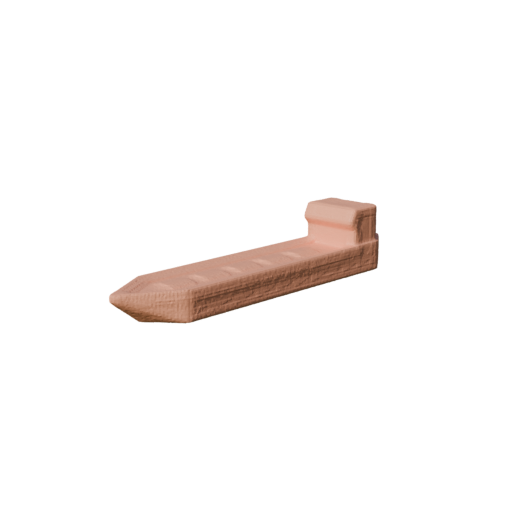}
\\ % Shape 6 image

%\bottomrule
\end{tabular}
\vspace{-3mm}
\caption{\textbf{Generalization to compositional prompts -- additional examples.} Baselines that heavily rely on input depth maps (e.g. ControlNet and Ctrl-X@60) appear to be over-constrained and ignore prompt composition. In contrast,  baselines that are under-constrained by the input depth
(e.g. CNet-Stop@30 or Ctrl-X@30) stray too much from target shape geometry. ShapeWords
achieves much better generalization to compositional prompts, while still demonstrating  strong adherence to the target shape. }
\label{fig:supp_prompt_qual_eval}
\end{figure*}

\begin{comment}
\begin{table*}[htbp]
\centering
\setlength{\tabcolsep}{1pt}
\begin{tabularx}{\textwidth}{>{\centering\arraybackslash}m{0.14\textwidth}*{6}{>{\centering\arraybackslash}X}}
\toprule
\textbf{Prompt} & \textbf{Input Depth} & \textbf{ControlNet} & \textbf{ControlNet@30} & \textbf{Ctrl-X} & \textbf{Ours} & \textbf{GT Shape} \\
\midrule
`A photo of an \textbf{object} on a beach' &
\includegraphics[width=\linewidth]{example-image-a} &
\includegraphics[width=\linewidth]{example-image-a} &
\includegraphics[width=\linewidth]{example-image-a} &
\includegraphics[width=\linewidth]{example-image-a} &
\includegraphics[width=\linewidth]{example-image-a} &
\includegraphics[width=\linewidth]{example-image-a} \\

`A photo of an \textbf{object} on a beach' &
\includegraphics[width=\linewidth]{example-image-a} &
\includegraphics[width=\linewidth]{example-image-a} &
\includegraphics[width=\linewidth]{example-image-a} &
\includegraphics[width=\linewidth]{example-image-a} &
\includegraphics[width=\linewidth]{example-image-a} &
\includegraphics[width=\linewidth]{example-image-a} \\
\bottomrule
\end{tabularx}
\caption{Caption of the table.}
\label{tab:example}
\end{table*}
\end{comment}